\documentclass{article}
\usepackage[utf8]{inputenc}
\usepackage{amssymb}

\PassOptionsToPackage{numbers, compress}{natbib}

\usepackage[main,final]{neurips_2026}

\usepackage[utf8]{inputenc} 
\usepackage[T1]{fontenc}    
\usepackage{hyperref}       
\usepackage{url}            
\usepackage{booktabs}       
\usepackage{amsfonts}       
\usepackage{nicefrac}       
\usepackage{microtype}      
\usepackage{xcolor}         
\usepackage{subcaption}
\usepackage{graphicx}
\usepackage{multirow}
\usepackage{amsmath}
\usepackage{wrapfig}

\def\revision{}

\title{Barycentric Guidance: Turning Foundation Image Editors into Continuous Affective Controllers} 

\author{%
  Harvey Mannering\\
  University of Southampton\\
  \texttt{hgm1g14@soton.ac.uk} \\
  \And
  Zhiwu Huang \\
  University of Southampton\\
  \texttt{Zhiwu.Huang@soton.ac.uk} \\
    \AND
   Adam Prügel-Bennett \\
  University of Southampton\\
  \texttt{apb@ecs.soton.ac.uk} \\
}

\begin{document}

\maketitle

\begin{abstract}

  {Emotions are expressed on a continuum and are naturally described in the valence-arousal (VA) space. Existing VA-based image editors are typically training-based and domain-specific, limiting them to VA regions seen during training and making them brittle across datasets and domains. Meanwhile, despite producing high-quality edits, state-of-the-art foundation image editors are not continuous affective controllers: non-linear emotion representations in text encoders makes prompt interpolation poorly calibrated, so accurate and continuous VA control remains unreliable. To fill this gap, we introduce Barycentric Guidance, a training-free framework that turns pretrained foundation image editors into continuous affective controllers over VA, requiring no finetuning, extra data, or model changes.
  Our method maps any target VA point to simplex-based guidance weights, {enabling VA controllability in three aspects}: (i) accurate targeting of VA states, (ii) broad coverage across the VA plane, and (iii) continuous control.
  Across three face datasets, we improve VA target accuracy by 25–29\%, expand VA-plane coverage by 28–50\%, and increase expression diversity by 29–49\%, over the strongest baseline. 
  Unlike prior domain-specific methods, our training-free approach exploits {the foundation model’s} inherent versatility, enabling generalization beyond human faces to animals, artworks, and complex scenes within a single framework. \revision{Code is available at \url{https://github.com/harveymannering/BarycentricGuidance}.}}
  
\end{abstract}

\section{Introduction}

\begin{figure}
  \centering
  \begin{subfigure}{0.325\linewidth}
  \centering
\includegraphics[height=4.55cm]{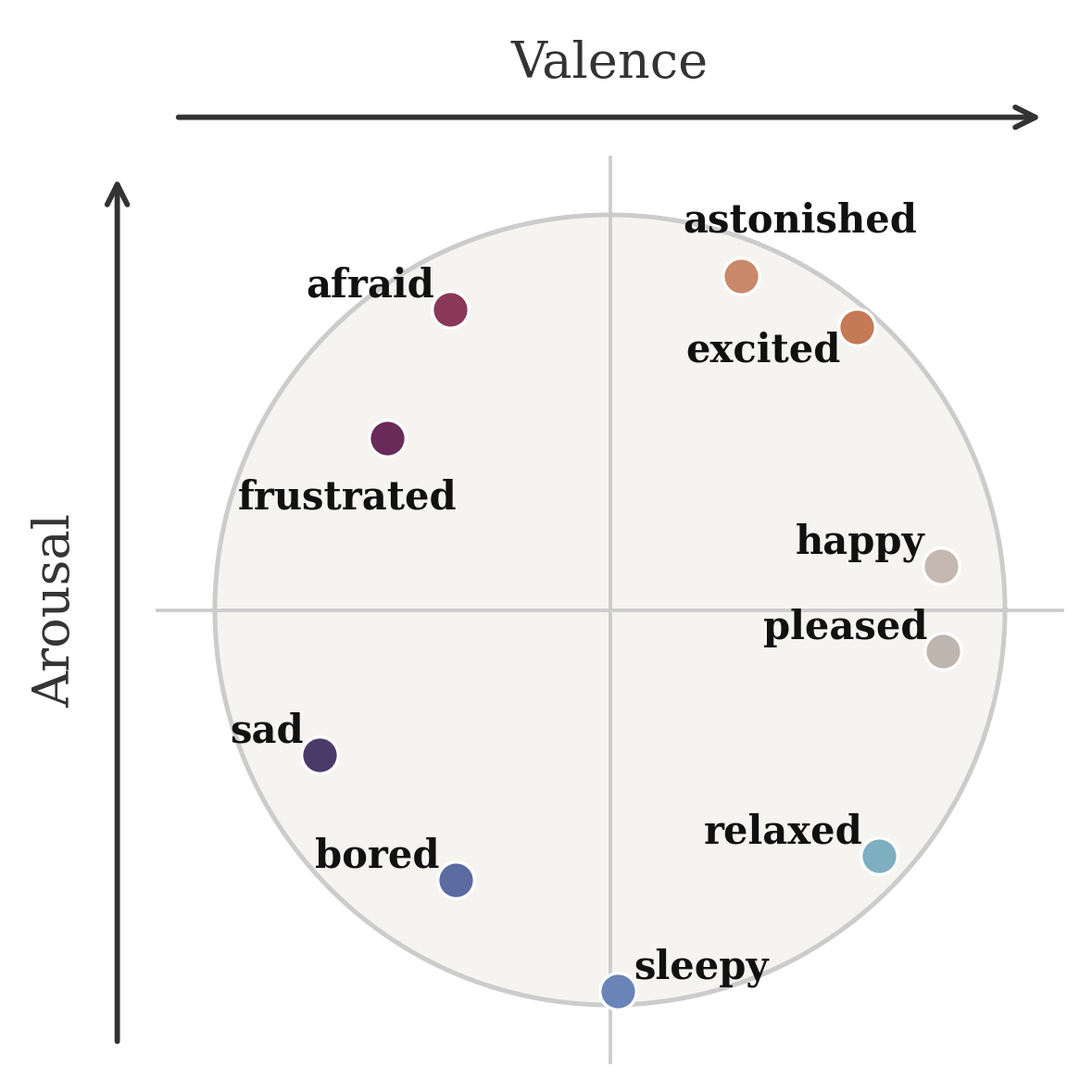}
    \caption{ Valence Arousal Model}
    \label{fig:va-a}
  \end{subfigure}
  \hfill
  \begin{subfigure}{0.325\linewidth}
  \centering
  \includegraphics[height=4.55cm]{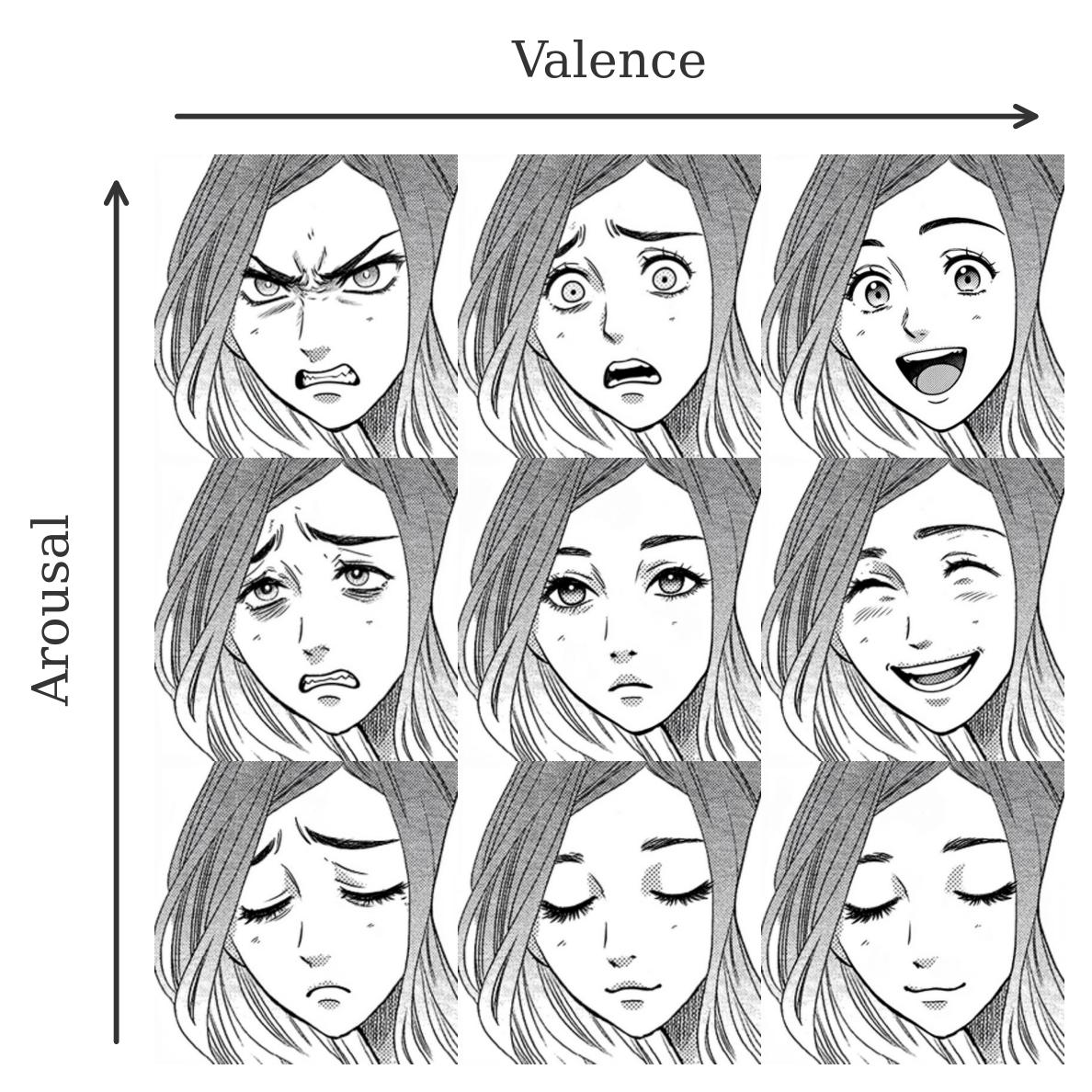}
    \caption{Manga Faces}
    \label{fig:preview-b}
  \end{subfigure}
  \hfill
  \begin{subfigure}{0.325\linewidth}
  \centering
  \includegraphics[height=4.55cm]{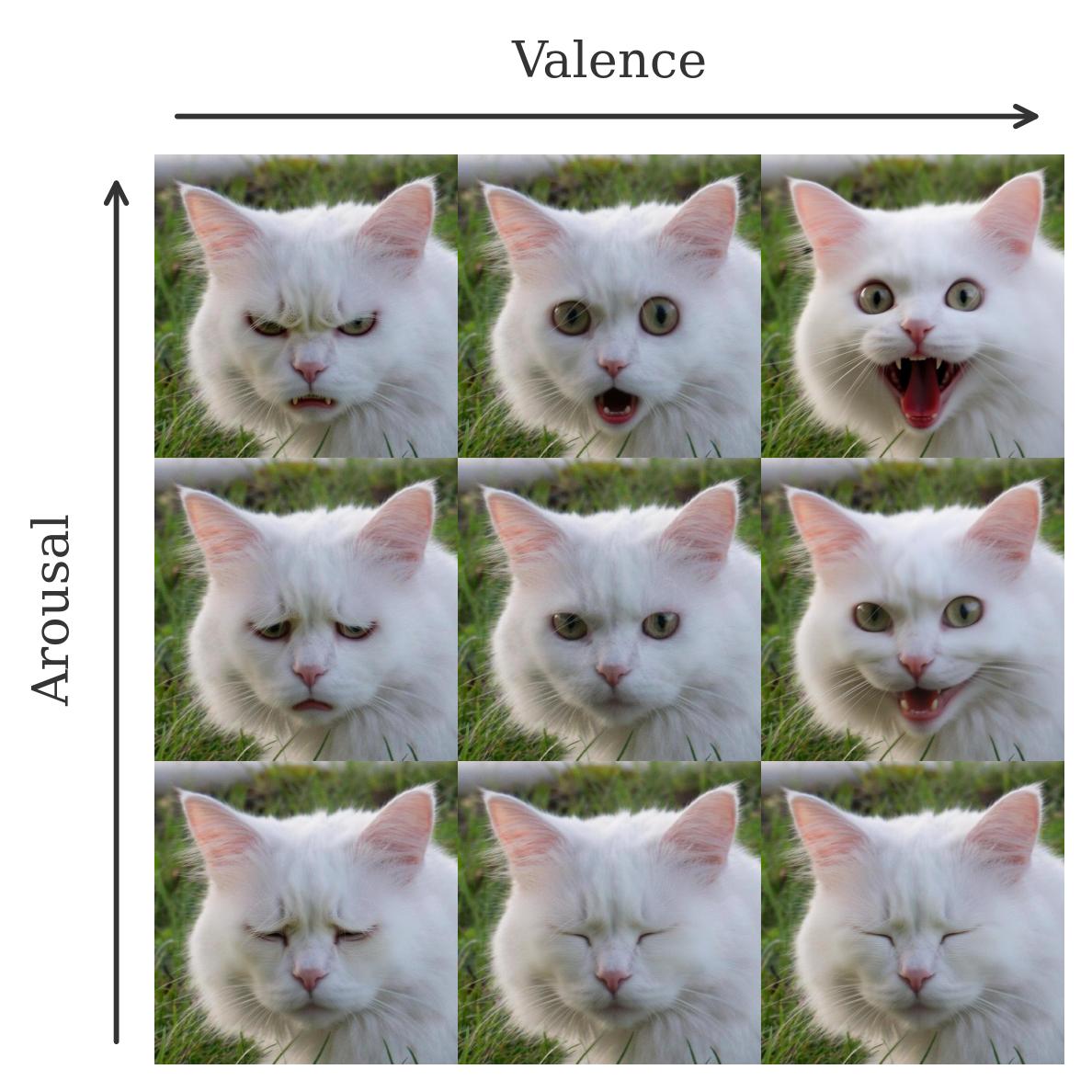}
    \caption{Animal Faces}
    \label{fig:preview-c}
  \end{subfigure} \\

  \begin{subfigure}{0.495\linewidth}
  \centering
  \includegraphics[height=6.75cm]{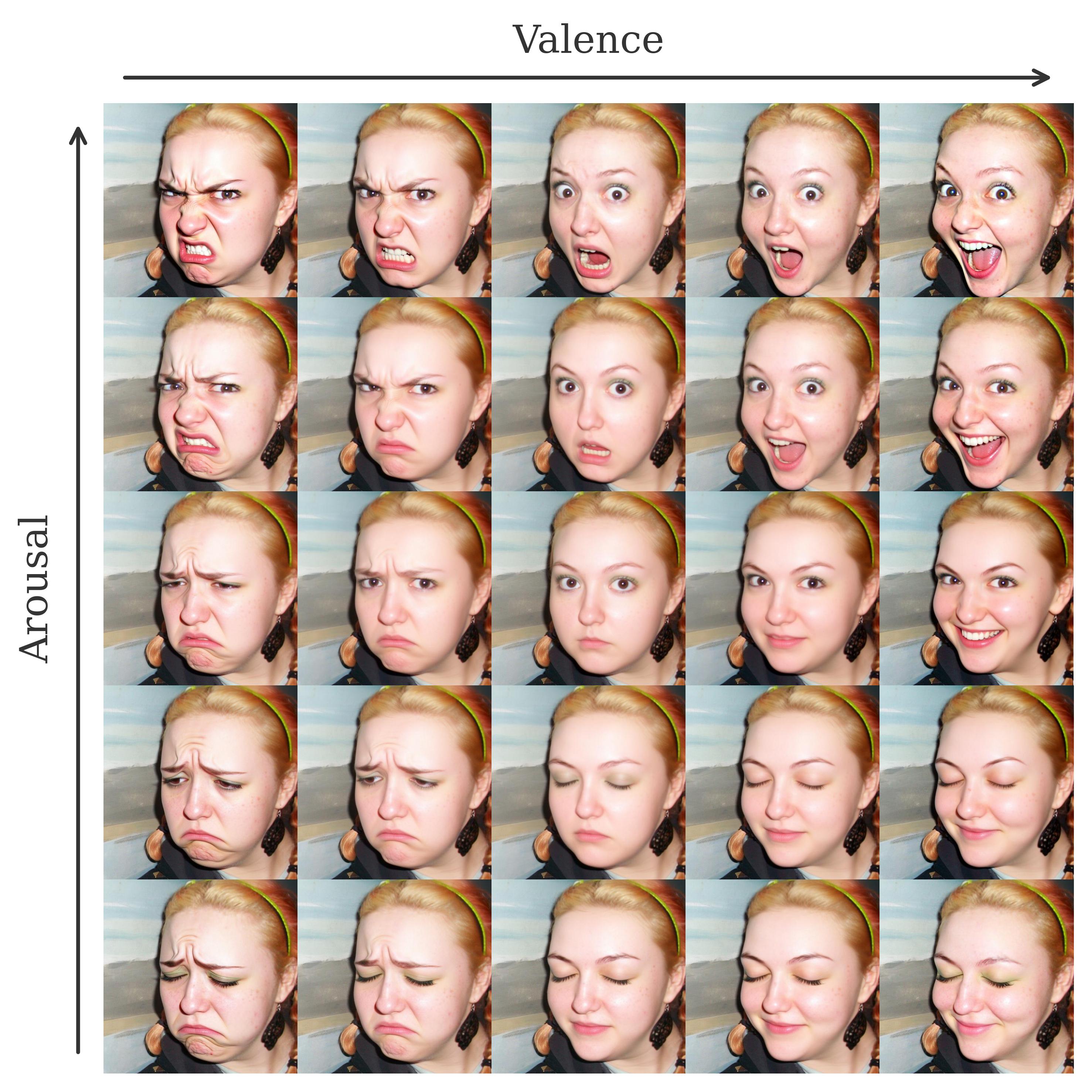}
    \caption{Human Faces}
    \label{fig:preview-e}
  \end{subfigure}
  \hfill
  \begin{subfigure}{0.495\linewidth}
  \centering
  \includegraphics[height=6.75cm]{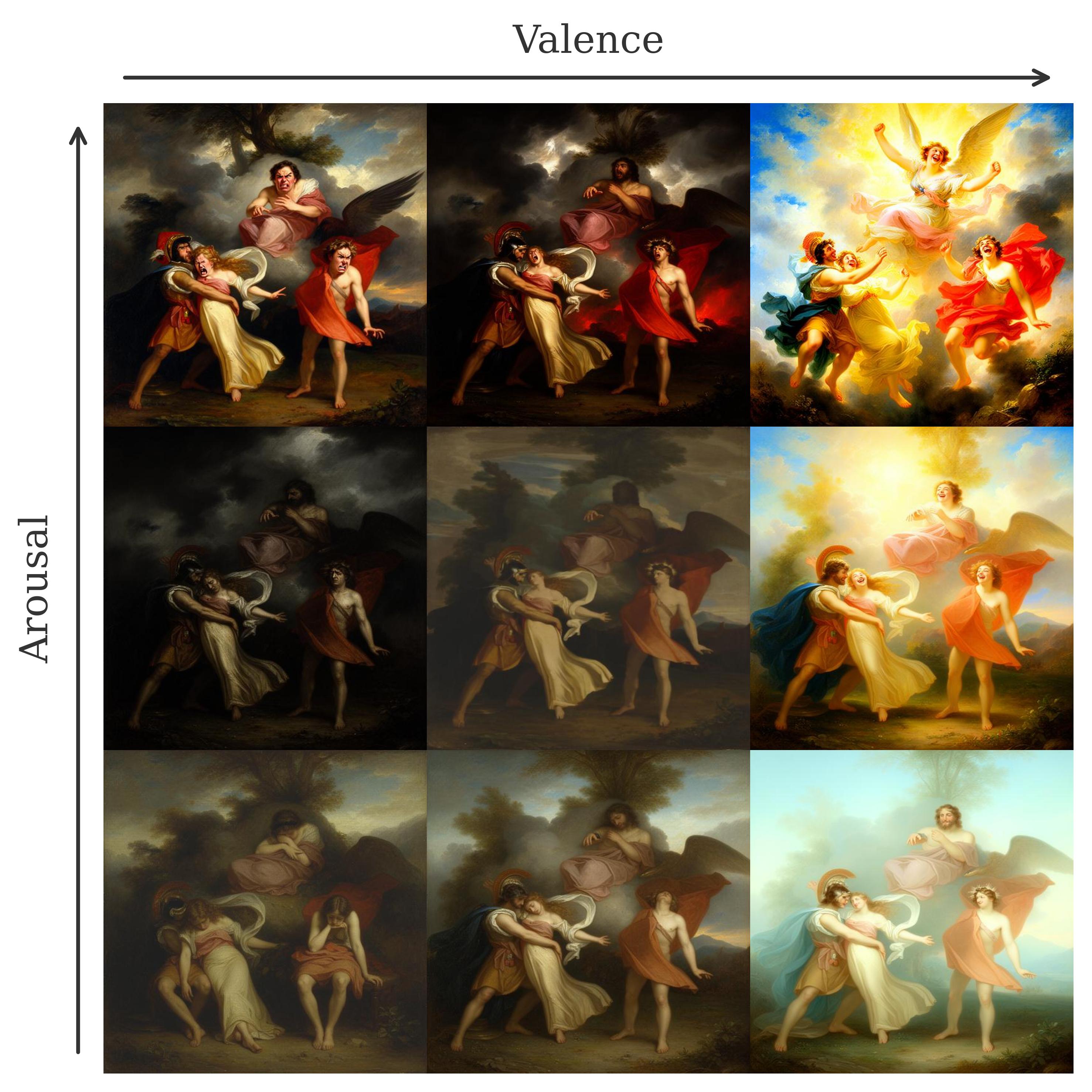}
    \caption{Artworks}
    \label{fig:preview-f}
  \end{subfigure}
  \caption{Valence-Arousal-based image editing results from our approach. {\bf(a)} The VA model of emotion from  \citet{russell1980circumplex} {\bf(b)} - {\bf(e)} VA-based editing across a range of datasets including  Manga Facial Expression \citep{koklu2021manga}, AFHQ \citep{choi2020stargan}, CelebA-HQ \citep{CelebAMask-HQ}, and EmoArt \citep{zhang2025emoart}. VA ranges [-1,1] with uniform steps.}
  \label{fig:preview}
\vspace{-4pt}
\end{figure}

Emotions are usually expressed in degrees, not only as discrete categories such as happy or sad. The valence–arousal (VA) model \citep{russell1980circumplex} describes this by representing affect as a continuous two-dimensional space: valence ranges from unpleasant to pleasant, and arousal from calm to excited (see Figure~\ref{fig:va-a}). This dimensional view provides an intuitive ``affective control’’ for visual emotion editing: changing VA should yield smooth, calibrated changes in facial expression (or visual mood) while keeping non-affective content, such as identity, hair, and background, unchanged. In this paper, we cast VA-based editing as a control problem: given a VA target, and optionally a source image, an editor should track that target reliably, behaving as a continuous affective controller over the VA plane.

{Despite this appeal, existing VA-based editing methods are typically training-based and tailored to specific tasks or domains, most commonly human faces or scenes \cite{d2021ganmut,azari2024emostyle,dang2025emoticrafter,mishima2025facecrafter}. These approaches often struggle to cover the full VA space: due to dataset bias toward near-neutral and socially common expressions, rare regions (e.g., high-arousal negative or low-arousal states) are underrepresented, leading to weak generalisation to unseen VA combinations and reduced controllability at the extremes \citep{darne2025generating}. Moreover, models trained for one domain often generalise poorly across domains e.g., between face datasets, from human faces to animal/cartoon/3D faces, or between faces and non-face imagery.}

At the same time, state-of-the-art foundation image editing diffusion models \citep{labs2025flux,wu2025qwenimagetechnicalreport,flux-2-2025} can perform a wide range of high-quality edits, including discrete emotional manipulation. However, they are not continuous affective controllers by default: 
non-linear text embedding spaces make prompt interpolation poorly calibrated, so accurate and continuous traversal of VA states remains unreliable.

\revision{We focus on VA because our aim is continuous affective control rather than a comprehensive model of emotion. Its two interpretable dimensions are naturally expressible in language and widely used in prior continuous affective editing work \cite{d2021ganmut,azari2024emostyle,dang2025emoticrafter,mishima2025facecrafter}. This exposes a gap between VA-based methods, which provide continuous control but are typically task-specific, and foundation image editors, which offer broad semantic and identity-aware editing but lack reliable continuous affect control. Alternative representations such as valence-arousal-dominance (VAD) and appraisal models capture richer affective factors, but are less standardised as smooth geometric control spaces for image editing.}

To address this gap, we introduce Barycentric Guidance, a {training-free framework} that turns pretrained foundation editors into continuous affective controllers over VA, requiring no additional data, finetuning, or model modification. Our method maps any target VA point to simplex-based guidance weights, enabling (i) accurate targeting of VA states, (ii) broad coverage across the VA plane, and (iii) continuous control with principled intensity scaling from a neutral centre. When needed, we further improve perceptual continuity with lightweight refinement (e.g., partial re-diffusion and interpolation such as morphing or optical flow).

{We present our main quantitative evaluation on facial expression editing (see Figure~\ref{fig:preview-e}), where VA control is well studied and benchmark datasets are available. Compared to prior VA-based facial editing methods \cite{d2021ganmut,azari2024emostyle,dang2025emoticrafter,mishima2025facecrafter}, our approach produces more diverse expressions, more accurate VA alignment, and stronger performance across a wider range of face datasets. Importantly, because our method is entirely training-free, it also generalises beyond human faces: Figure~\ref{fig:preview-b}, \ref{fig:preview-c}, \& \ref{fig:preview-f} illustrate results on cartoon faces, animal faces, and artworks, respectively. In summary, our contributions are:
\begin{itemize}
\item {We propose a training-free framework that turns pretrained foundation image editors into continuous affective controllers in VA space, focusing on {three aspects of VA controllability: accuracy, coverage, and continuity}.}
\item We evaluate our approach on facial expression editing, finding that we achieve state of the art alignment with the valence-arousal model.
\item We demonstrate that our approach can be applied beyond human faces, enabling training-free emotion editing of non-human faces, scenes, and artworks.
\end{itemize}

\section{Related Works}

\paragraph{Deep Generative Models}
Generative models learn data distributions and sample from them to produce new data. Early deep learning approaches include VAEs \citep{kingma2013vae} and GANs \citep{goodfellow2014generative}, while recent work focuses on autoregressive \citep{radford2018improving,tian2024visual} and diffusion models \citep{song2020score,ho2020denoising}. Diffusion and its flow-based variants \citep{lipman2022flow,liu2022flow} have been particularly impactful due to stable training, high-quality, and diverse outputs, enabling text-to-image systems such as Stable Diffusion \citep{rombach2022high} and FLUX \citep{flux2024}. However, because text input alone makes fine-grained control difficult, a series of finetuning approaches such as ControlNet \citep{zhang2023adding} and IP-Adapter \citep{ye2023ip} have been developed. These finetuning methods, in turn, face their own challenges due to the large size of text-to-image models themselves. Recently, a spate of image editing diffusion models have been released. Popular open source examples include FLUX.1 Kontext \citep{labs2025flux}, Qwen Image Edit \citep{wu2025qwenimagetechnicalreport}, and FLUX.2 \citep{flux-2-2025}. These models excel at discrete editing, but struggle to edit the properties of images in a continuous fashion \citep{parihar2025kontinuous}, making their application to VA-based editing non-trivial. While finetuning-based approaches enable continuous editing \cite{parihar2025kontinuous,hua2026pixelsmile}, they incur high computational cost and limit generalization. In contrast, our method is training-free, enabling broad applicability and easy adaptation to new models.

\begin{wrapfigure}{r}{0.39\textwidth}     \begin{minipage}{0.39\textwidth}
\centering
 \includegraphics[height=4.9cm]{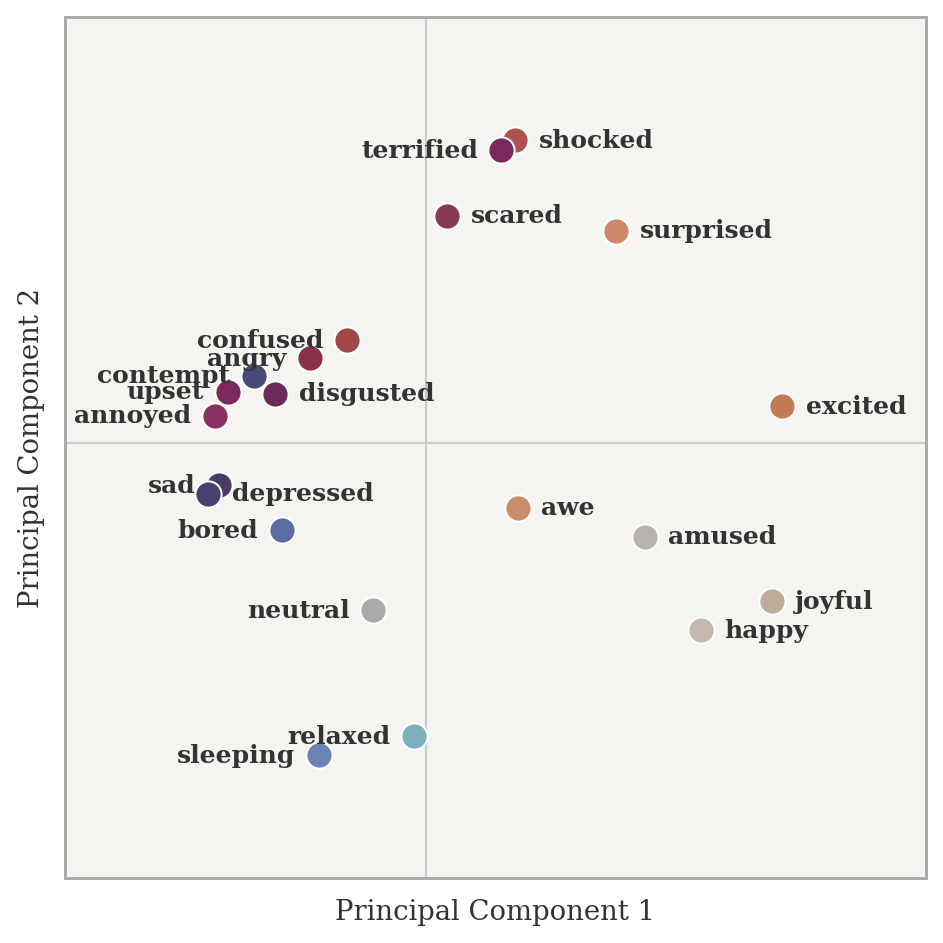}
    \caption{PCA Performed on CLIP Text Embedddings.}
    \label{fig:va-b}
  \end{minipage}
\end{wrapfigure}

\paragraph{Continuous Emotion Editing} 
Early work on facial expression editing used GANs \citep{choi2018stargan,pumarola2018ganimation,azari2024emostyle,d2021ganmut} with various control signals, including discrete emotions \citep{ekman1971universals}, Action Units \citep{ekman1978facial}, the VA model \citep{russell1980circumplex}, and learned continuous emotion spaces. However, these methods cannot leverage the generalization ability of text-to-image models, making them brittle when applied to out-of-distribution face images.
For diffusion models, a common way to control facial expressions is by conditioning on a 3D face model \citep{ding2023diffusionrig,wang2024facecomposer,han2024face}, Action Units \citep{varanka2024towards}, or landmarks \citep{zhao2023diffswap,xu2025hifi}. While these methods enable fine-grained control, they lack an intuitive and easily interpretable model of emotion, such as that provided by the valence–arousal model. 
\citet{paskaleva2024unified} unify Action Units and VA via an MLP that augments Stable Diffusion text conditioning. While limited by artifacts and training overhead, their emotion model is an interesting unification of existing approaches. FaceCrafter \citep{mishima2025facecrafter} finetunes Arc2Face \citep{papantoniou2024arc2face} for ID, expression, and pose conditioned face generation. Concurrent work, PixelSmile \citep{hua2026pixelsmile}, finetunes Qwen Image Edit with a LoRA \citep{hu2021lora} using a contrastive loss allowing for continuous control of several discrete emotions, we on the other hand focus on emotion editing based on the VA model.

Beyond faces, EmotiCrafter \citep{dang2025emoticrafter} 
trains a transformer to maps text and VA inputs to Stable Diffusion text embeddings for image generation of scenes. Emotional image editing \cite{yang2025emoedit,lin2025make,zhang2026affective}, generation \citep{yang2024emogen}, and stylization \citep{yang2025emostyle} methods have also been applied to control the emotion evoked by an image. While these approaches achieve impressive results, they typically require supervised finetuning or rely on discrete textual prompts. Together, the diffusion-based methods described achieve remarkable performance in emotion generation and editing across their respective domains. However, finetuning on specific datasets requires large compute resources and can restrict the capabilities of a foundation models. We instead focus on how best to leverage foundation models without any finetuning, thus inheriting the multi-domain strengths of the original text-to-image models.

\section{Proposed Approach}

\subsection{Motivation \& Background}

First, we explore the ability of text encoders to control valence and arousal in facial expressions. Specifically, we examine the CLIP \cite{radford2021learning} embeddings used in the Stable Diffusion \citep{rombach2022high} and FLUX.1 \citep{flux2024} models. We use CLIP to embed prompts with the template ``a portrait photo of a <emotion> face'', where ``<emotion>'' is replaced with an emotion word (e.g. ``happy'', ``sad'', etc). We then do dimensionality reduction on the CLIP embeddings using PCA. Interestingly, this largely recovers the valence–arousal model on the two principal components, as shown in Figure \ref{fig:va-b}, suggesting CLIP implicitly encodes VA structure.  This holds across different CLIP models and prompt templates, but not for encoders such as T5 \citep{raffel2020exploring} and QwenVL \citep{bai2025qwen3} (see Section \ref{sec:appendix_text_encoders} for details).

While this linear subspace enables VA control in CLIP-based models, newer systems rely less on CLIP. For example, FLUX.1 Kontext uses CLIP only as a secondary encoder, with T5 as primary, which lacks such structure. Since the text encoders for Qwen Image Edit and FLUX.2 also show no evidence of having learned a valence-arousal subspace, a different approach must be taken to control facial expressions based on valence and arousal. Moreover, prompt interpolation often yields abrupt edits (see Sections \ref{sec:Bary} and \ref{sec:fee}). To bypass these problems with the text encoder embedding spaces, we therefore work in the diffusion models latent space instead.

\subsection{Barycentric Guidance}
\label{sec:Bary}

\begin{wrapfigure}{r}{0.39\textwidth}     \begin{minipage}{0.39\textwidth}
\centering
  \includegraphics[height=4.9cm]{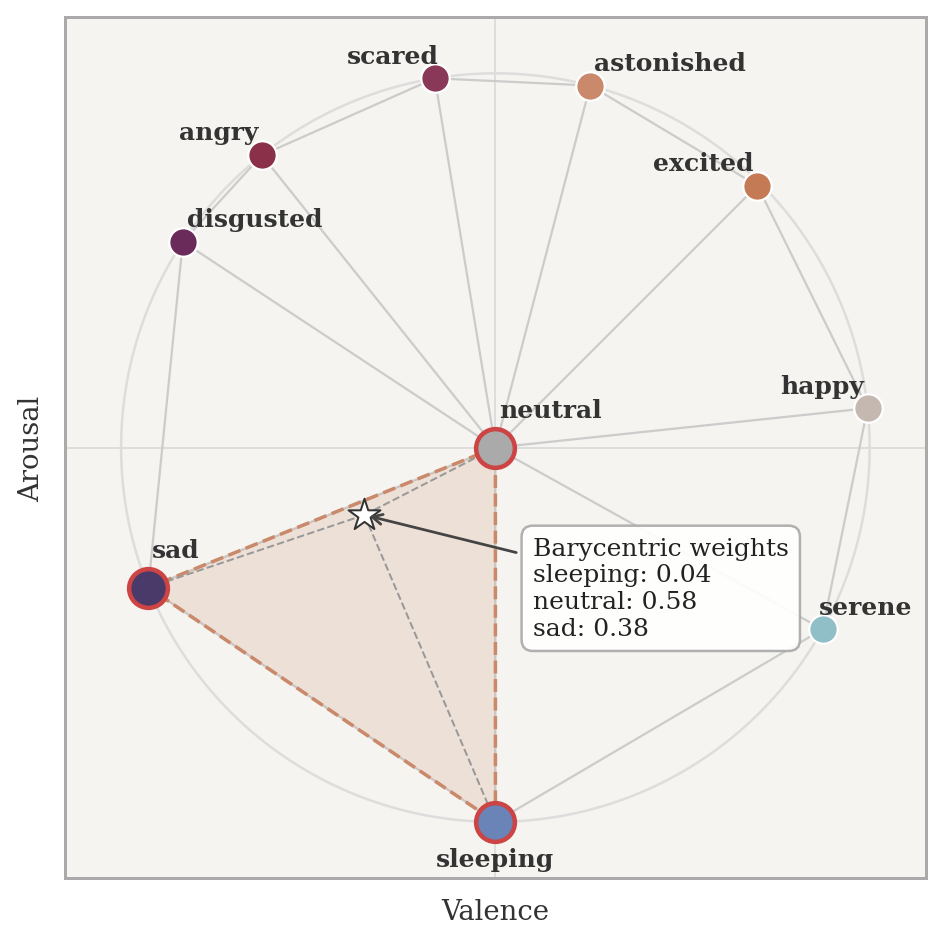}
    \caption{Triangle Mesh created from VA Points.}
    \label{fig:va-c}
  \end{minipage}
\end{wrapfigure}

Image editing models excel at discrete editing, but struggle with continuous control. We can leverage this advantage toward the goal of continuous editing by taking linear combinations of discrete outputs. To this end, as shown in Figure \ref{fig:va-c}, we approximate the VA model by constructing a triangle mesh in 2D of discrete emotions. The vertices of this mesh are defined by discrete emotion anchors relevant to the target domain; for instance, in facial expression editing, points are based on the average locations of the valence arousal points in the AffectNet dataset \citep{mollahosseini2017affectnet}, with additional expressions excited, sleepy, and serene used to fill out the remainder of the VA space. 

For a given valence arousal point $(v,a)$, we first need to identify the triangle within which the point lies. If the point is outside all triangles, we choose the triangle that is closest to $(v,a)$. From this we can get the three discrete emotions that are closest to our valence arousal point. Barycentric interpolation can be used to determine the weights $(w_1,w_2,w_3)$ to assign to each discrete emotion by solving the system of equations
\begin{equation}
    \centering
    \begin{bmatrix}
        A_x & B_x & C_x \\
        A_y & B_y & C_y \\
        1   & 1   & 1
    \end{bmatrix}
    \begin{bmatrix}
        w_1 \\
        w_2 \\
        w_3
    \end{bmatrix}
    =
    \begin{bmatrix}
        v \\
        a \\
        1
    \end{bmatrix},
    \label{eq:barycentric_interpolation}
\end{equation}
where $A$, $B$, and $C$ are the three corners of the triangle, and $\sum_{i=1}^3w_i=1$. These weights can then be used to interpolate between different prompt embeddings, however, we find that this prompt interpolation results in sudden and uneven changes when making edits (see Figure \ref{fig:color-b}). Instead of interpolating the text embedding, we therefore reformulate classifier-free guidance \citep{ho2022classifier} to be compatible with our weights $(w_1,w_2,w_3)$. Given a diffusion model $D_\theta$, we can reformulate the noise prediction as
\begin{equation}
\begin{aligned}
D^{edit}_{\text{bary}}(x_t)
= & w_1 D_{\theta}(x_t,y_1) + w_2 D_{\theta}(x_t,y_2) + w_3 D_{\theta}(x_t,y_3), \\
\approx & \nabla_{x_t} \log p_t(x_t)
  + \sum_{i=1}^3 w_i \nabla_{x_t} \log p_t(y_i \mid x_t),
\end{aligned}
\label{eq:barycentric_editing}
\end{equation}
where we have conditions $y_i$ for each discrete emotion. Conditions take the form of text prompts, for example, for facial expression editing we use a prompt template such as ``Change this persons facial expression to be <emotion> while maintaining the same facial features, composition, lighting, and camera angle''. Unlike prompt interpolation, this formulation lets us directly manipulate the directions that increase or decrease the likelihood of specific emotions in $y_i$ according to the implicit classifier $p(y_i|x_t)$, thus allowing for smoother interpolations. 

To show this visually, consider the example for continuous colour control in Figure \ref{fig:colour}. By defining the triangle mesh of different colours shown in Figure \ref{fig:color-a}, we can control the colour of our edit in two dimensions. However, using weights derived from our mesh to simply combine text embeddings, as shown in Figure \ref{fig:color-b}, results in large homogeneous patches of colour and sudden changes in colour. Barycentric guidance by comparison, shown in Figure \ref{fig:color-c}, gives smooth transition colours.
\begin{figure}
  \centering
  \begin{subfigure}{0.3\linewidth}
  \centering
  \includegraphics[height=3.85cm]{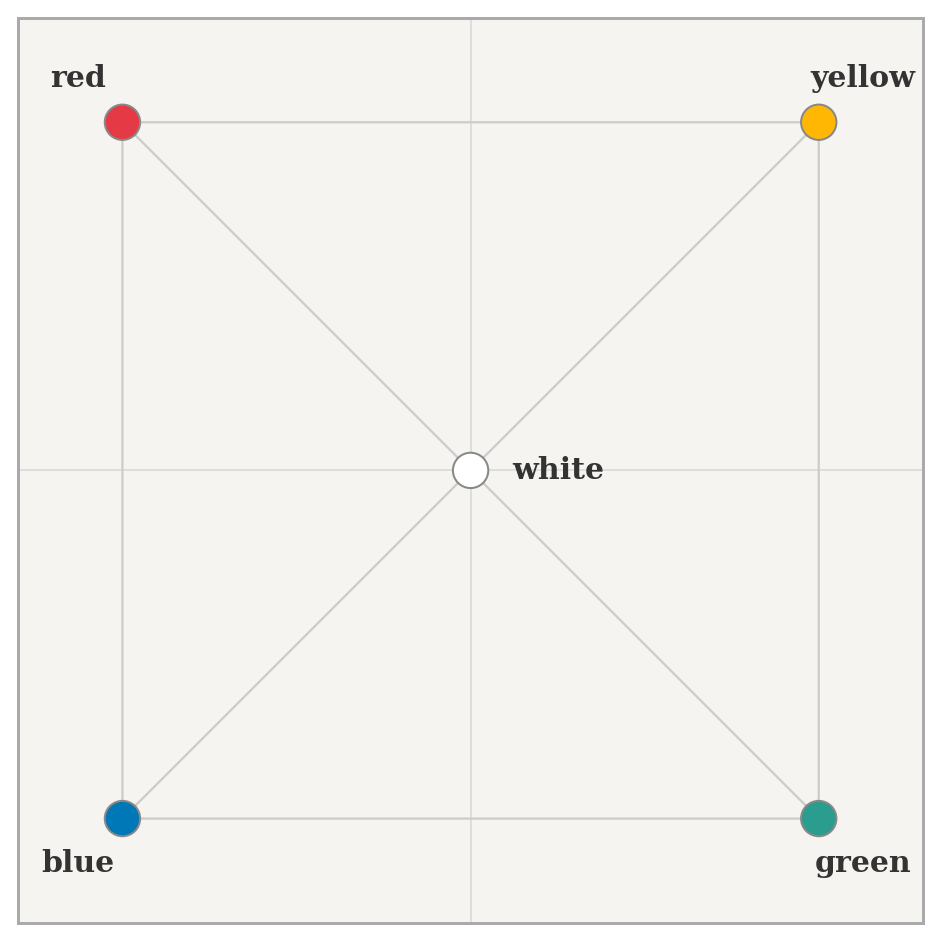}
    \caption{Simplex mesh of colours.}
    \label{fig:color-a}
  \end{subfigure}
  \hfill
  \begin{subfigure}{0.3\linewidth}
  \centering
  \includegraphics[height=3.85cm]{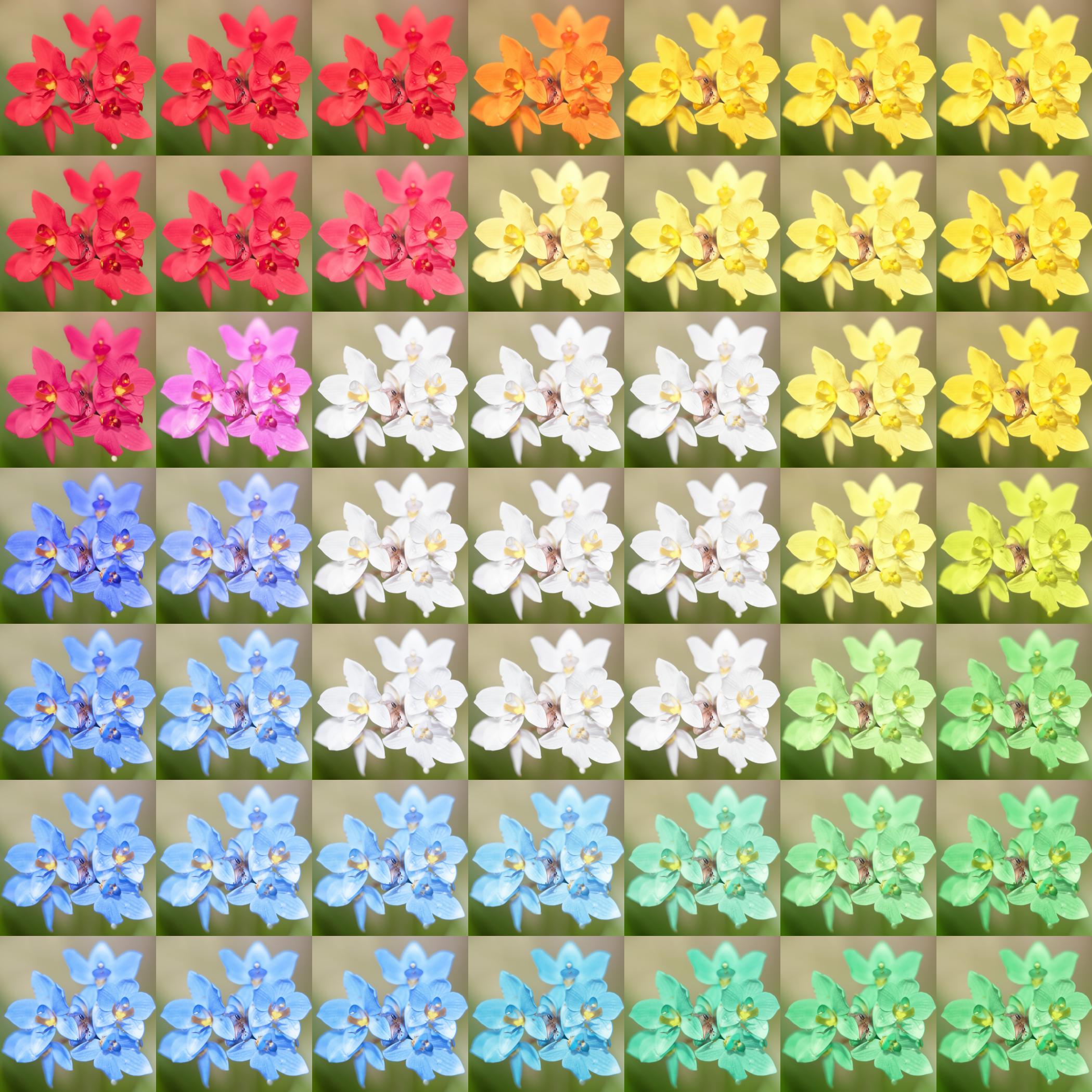}
    \caption{Prompt Interpolation.}
    \label{fig:color-b}
  \end{subfigure}
  \hfill
  \begin{subfigure}{0.3\linewidth}
  \centering
  \includegraphics[height=3.85cm]{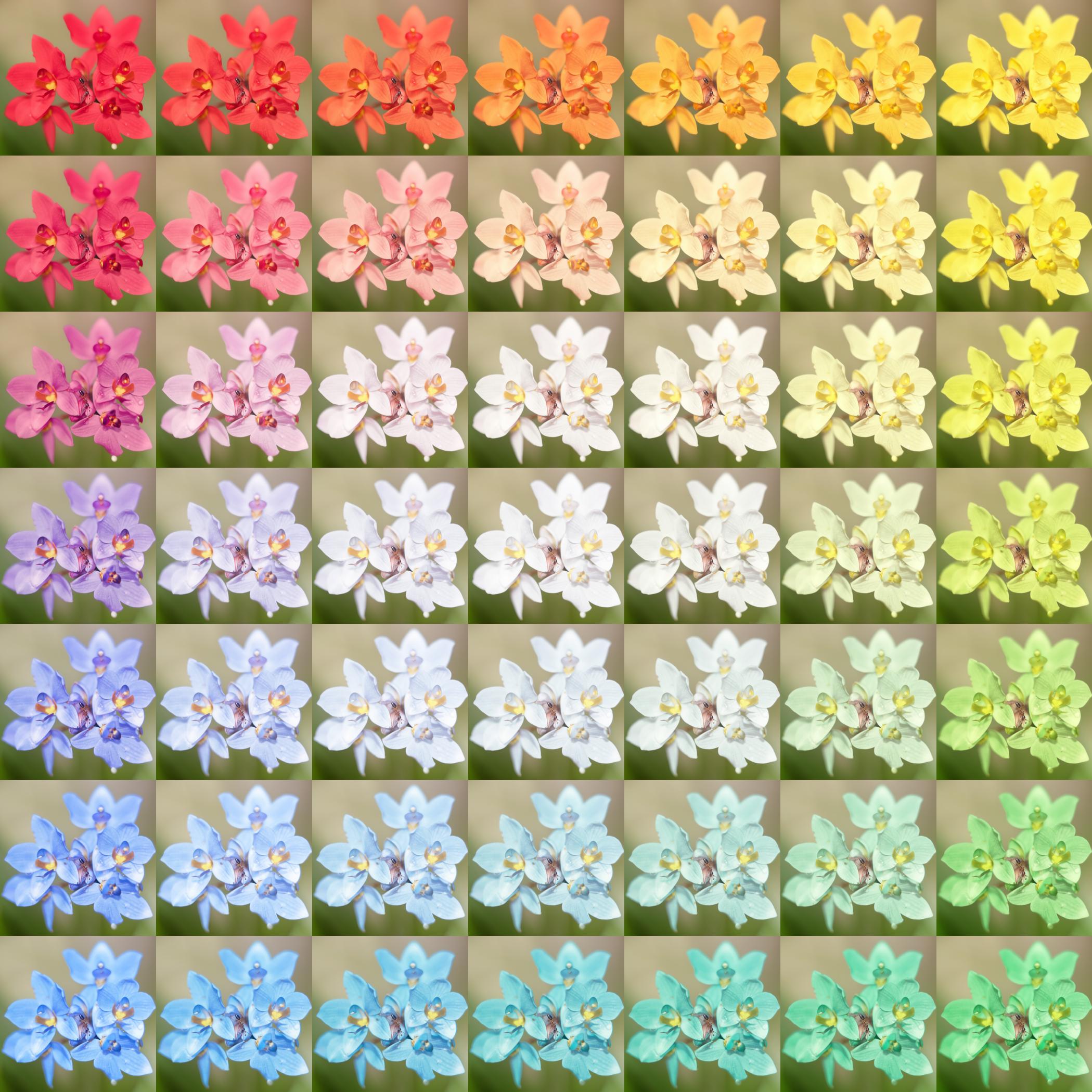}
    \caption{Barycentric Guidance.}
    \label{fig:color-c}
  \end{subfigure}
  \caption{Colour editing comparison between prompt interpolation and barycentric guidance. Prompts: ``Make this flower <colour>, keep everything else the same colour.''; model: FLUX.2 Klein 9B. Source image from EmoSet \citep{yang2023emoset}.}
  \label{fig:colour}
\end{figure}

If we want to increase the likelihoods of $y_i$ further we can combine Barycentric guidance with classifier-free guidance (CFG), using the guidance scale $w_{cfg}$, as
\begin{equation}
(1-w_{cfg})D_{\theta}(x_t) + w_{cfg} \sum_{i=1}^3 w_i D_{\theta}(x_t,y_i).
\end{equation}
We also find that the guidance scale can serve as a useful proxy for emotional intensity. To leverage this we propose adjusting the guidance scale like
\begin{equation} 
w^*_{cfg} = 1 + (w_{cfg} - 1)\sqrt{v^2+a^2},
\end{equation}
meaning the effective guidance strength is weak towards the center of the VA model and grows stronger towards more extreme values.

To apply Barycentric Guidance to image generation rather than editing, we can also introduce a base prompt $y_{base}$ (e.g. $y_{base}$=``a man is running fast'') that specifies the scene semantics. If we also apply CFG on top we then have
\begin{equation} 
\begin{aligned}
D^{gen}_{bary}(x_t) = (w_{cfg}-s)D_{\theta}(x_t,y_{base}) \revision{+} (1-w_{cfg})D_{\theta}(x_t) + s\left(\sum_{i=1}^3w_iD_{\theta}(x_t,y_i)\right), \\
\approx \nabla_{x_t} \log p_t(x_t) + (w_{cfg}-s) \nabla_{x_t} \log p_t(y_{base} \mid x_t) + s\sum_{i=1}^3 w_i \nabla_{x_t} \log p_t(y_i \mid x_t),
\end{aligned}
\label{eq:barycentric_generation}
\end{equation}
where $s<w_{cfg}$ is a parameter which trades-off the strength of the base prompt against the strength of the emotion prompts.

\subsection{Geometry-Aware Latent Interpolation}

While Barycentric Guidance provides a robust framework for semantic interpolation, guidance can occasionally struggle with fine grained geometric consistency in highly structured domains, such as human faces. To bridge this gap, we propose an optional structural refinement module. This module complements Barycentric Guidance by explicitly interpolating latent geometries before the final diffusion pass.

Specifically for human faces, we first generate edits for three discrete emotions corresponding to three emotions found in our triangle mesh. We then combine them using face morphing, weighting each image and set of face landmarks according to our barycentric weights. An inbetween image allows us to smoothly interpolate facial geometric, but also creates noticeable ghosting artifacts. To rectify this, we can simply restart the reverse diffusion process part way through using this morphed image in the style of SDEdit \citep{meng2021sdedit}, effectively creating a structural prior. We find that restarting the diffusion process one-third of the way through the total steps ensures that morphed geometry is preserved without ghosting, as shown in Section \ref{sec:morph_step}.  For non-human faces, landmarks cannot be used, however, we find that optical flow based image in betweening works well as a substitute.

\section{Experiments}
\label{sec:experiments}

In this section, we evaluate VA-based face and non-face image editing and generation. Unless otherwise stated, we use FLUX.2-dev, though some experiments employ FLUX.2-klein, FLUX.1 Kontext, or Qwen Image Edit 2511.  See Section \ref{sec:exp_details} for hyperparameters and further experimental details.


\begin{figure}[htbp]
  \centering
  

  \begin{minipage}{0.30\textwidth}
    \centering
    \captionof{table}{Impact of facial expression diversity (VA range) on FID. Images are taken from AffectNet.}
    \label{tab:FID_faces}
    \small 
      \begin{tabular}{lc}
        \toprule
        VA Subset & FID $\downarrow$ \\
        \midrule
        Neutral Only & 8.36 \\
        $|VA| \leq 0.25$ & 8.56 \\
        $|VA| \leq 0.5$ & 10.05 \\
        $|VA| \leq 0.75$ & 11.64 \\
        Full VA Range & 13.40 \\
        \bottomrule
      \end{tabular}
  \end{minipage}
  \hfill
  \begin{minipage}{0.68\textwidth}
  \centering
  \includegraphics[height=3.5cm]{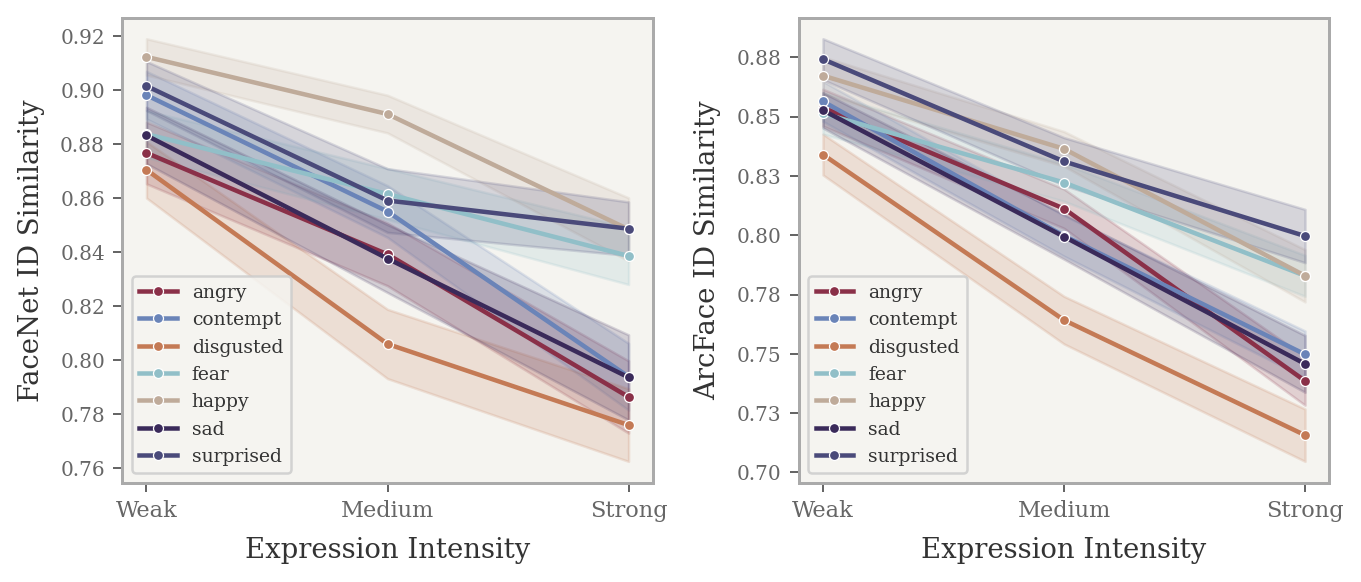}
  \caption{As facial expression diversity increases, FaceNet/ArcFace similarity decrease. Comparing neutral to target emotions across intensity levels, ID similarity drops as expression intensity increases. Images from MEAD dataset (examples shown in Figure \ref{fig:mead_examples}). }
  \label{fig:trade_off_mead}
  \end{minipage}
\end{figure}


\paragraph{Valence-Arousal Accuracy} We measure the accuracy of VA editing and generation using RMSE between predicted and target VA values. To cover the entire VA space we generate 100 images using a 10 by 10 grid of images with uniformly spaced target VA values in the range of $[-1,1]$. These 10 by 10 grids are also used when calculating smoothness and diversity metrics.

\paragraph{Valence-Arousal Diversity} Following EmoStyle \citep{azari2024emostyle}, we report the standard deviation of the predicted VA values as a measure of facial expression diversity. However, we note that this metric is imperfect, since a model producing only extreme facial expressions registers as more diverse than one producing a mix of extreme and mild expressions. We therefore propose measuring the entropy of predicted VA points using Kernel Density Estimation as a better diversity measure.

For face generation specifically, we note one important caveat when looking at diversity, which is that facial expression diversity trades-off against ID similarity as measured by facial recognition networks like FaceNet \citep{schroff2015facenet} or ArcFace \citep{deng2019arcface}.  To validate this on real data, we used the MEAD dataset \citep{kaisiyuan2020mead}, which contains videos of 60 actors displaying basic emotions at three intensity levels (weak, medium, strong), along with neutral expressions. We extract mid-frame images for 48 actors (split 0) and compute cosine similarity between neutral and emotional expressions across intensities. Although identity should ideally remain constant, we observed a consistent decrease in ID similarity with increasing expression strength (Figure \ref{fig:trade_off_mead}), confirming a strong identity-expression trade-off.

\paragraph{Valence-Arousal Smoothness} We evaluate smoothness over the full VA grid using a Laplacian-based curvature measure. Consider a function $z(x,y)$ which maps from a point on a square lattice $(x,y)$ to a valence arousal point $(v,a)$. We estimate local curvature via a discrete Laplacian, approximated using finite differences, and then measure smoothness as
\begin{align}
S^*_{\mathrm{2D}} &= \frac{1}{(W-2)(H-2)}\sum_{i=2}^{H-1}\sum_{j=2}^{W-1}\sqrt{\left(\frac{\Delta z^v(x,y)}{r_v}\right)^2 + \left(\frac{\Delta z^a(x,y)}{r_a}\right)^2},
\end{align}
\begin{align}
\Delta z(x,y) = z(x+1,y)+z(x-1,y)+z(x,y+1)+z(x,y-1)-4z(x,y),
\end{align}
where normalizing by the ranges of each dimension $r_v$ and $r_a$ ensures that this measurement is scale invariant. That is, we only consider the smoothness relative to the overall change in expression, which avoids methods that produce little to no change in the emotion being registered as smoother.

\paragraph{Image Quality}  Finally, we report Fréchet Inception Distance (FID) \citep{heusel2017gans} and Inception Score (IS) \citep{salimans2016improved} to measure image quality, calculated using 3.5k images. As with ID similarity, we find that facial expression diversity can impact FID negatively. To study this, we formed six 4k-image subsets of AffectNet (Neutral A/B and four quasi-uniform samples with VA magnitude bounds of $\leq0.25$, $\leq0.5$, $\leq0.75$, and $\leq1.0$) and computed FID against Neutral A (Table \ref{tab:FID_faces}). The rising FID scores ($8.36$ to $13.40$) suggest that the metric penalizes expression distribution shifts in addition to measuring realism.

\subsection{Facial Expression Editing}
\label{sec:fee}

\begin{table}
  \caption{Facial Expression Editing comparison with other VA editing models. {We treat Entropy, VA Std, RMSE, and $S^{*}_{2D}$ as the primary controllability metrics.} {$^{\dagger}$: The method does not control VA directly; instead, it approximates VA editing using polar coordinates to steer emotion.}}
  \label{tab:fee}
  \centering
  \scriptsize
  \begin{tabular}{llcccc|cccc}
    \toprule
    Dataset & Model & \multicolumn{4}{c|}{\textbf{VA Controllability}} & \multicolumn{4}{c}{Image Quality/ Preservation}  \\

\cmidrule(lr){3-6} \cmidrule(lr){7-10}

 &  & {RMSE}$\downarrow$ & {VA Entropy}$\uparrow$ & {VA Std}$\uparrow$ & {$S^{*}_{2D}$}$\downarrow$ & FID$\downarrow$ & IS$\uparrow$ & ID Sim.$\uparrow$ & LPIPS$\downarrow$
 \\
    \midrule
    \multirow{4}{*}{AffectNet} & GANmut$^{\dagger}$ & -	 & 0.616 & 0.47 / 0.38	& - & 36.89 & 3.90	& \bf 0.76 & 0.128 \\
     & EmoStyle & 0.610 & 0.423	& 0.46 / 0.31 & \bf0.147 & 36.72	&2.81 &	0.31 & 0.276 \\
      & {Ours (Qwen)} & \bf{0.431} & {0.813} & {0.55 / 0.49} & {0.278} & \bf{10.67} & \bf{4.71} & {0.72} & \bf{0.062} \\
      & Ours & 0.433 & \bf 0.916 & \bf 0.64 / 0.57 & 0.303 & 11.02 & 4.42 & 0.55 & 0.096 \\
      
     \midrule
     \multirow{3}{*}{FFHQ} & GANmut$^{\dagger}$ & - & 0.687 & 0.50 / 0.36 & - & 73.20 & 4.49 & \bf0.71 & 0.282  \\
     & EmoStyle & 0.568 & 0.457 & 0.48 / 0.32 & \bf 0.148 & 25.50 & 4.12 & 0.67 & 0.172 \\
     & Ours & \bf 0.418 & \bf0.890 & {\bf0.64} / {\bf0.51} & 0.307 & \bf13.85 & \bf5.61 & 0.55 & \bf0.097  \\
     \midrule
     \multirow{3}{*}{CelebA-HQ} & GANmut$^{\dagger}$ & - & 0.651	& 0.48 / 0.37 & - & 59.59 & 3.24 & \bf0.70	& 0.267\\
     & EmoStyle & 0.550 & 0.513 & 0.47 / 0.34 & \bf0.149 & 17.44 & 2.94 & 0.65 & 0.154 \\
     & Ours & \bf0.412 & \bf0.877 &	{\bf0.65} / {\bf0.54} & 0.293 & \bf11.80	& \bf3.53	&0.54 & \bf0.125\\
    \bottomrule
  \end{tabular}
\end{table}

\begin{figure}
  \centering
  \begin{subfigure}{0.99\linewidth}
  \centering
  \includegraphics[height=3.1cm]{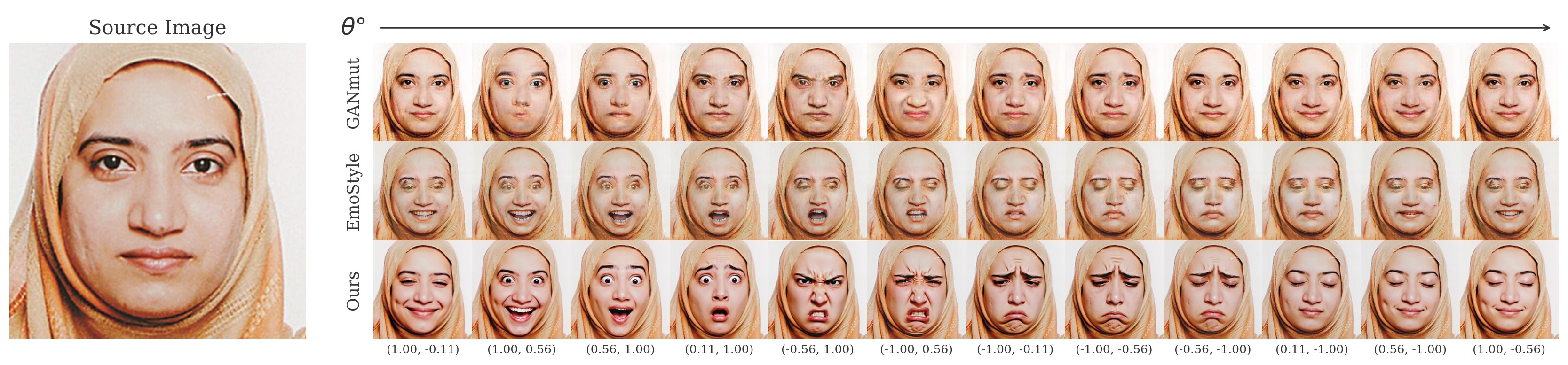}
    \label{fig:affectnet}
  \end{subfigure}
  \hfill
  \begin{subfigure}{0.99\linewidth}
  \centering
  \includegraphics[height=3.1cm]{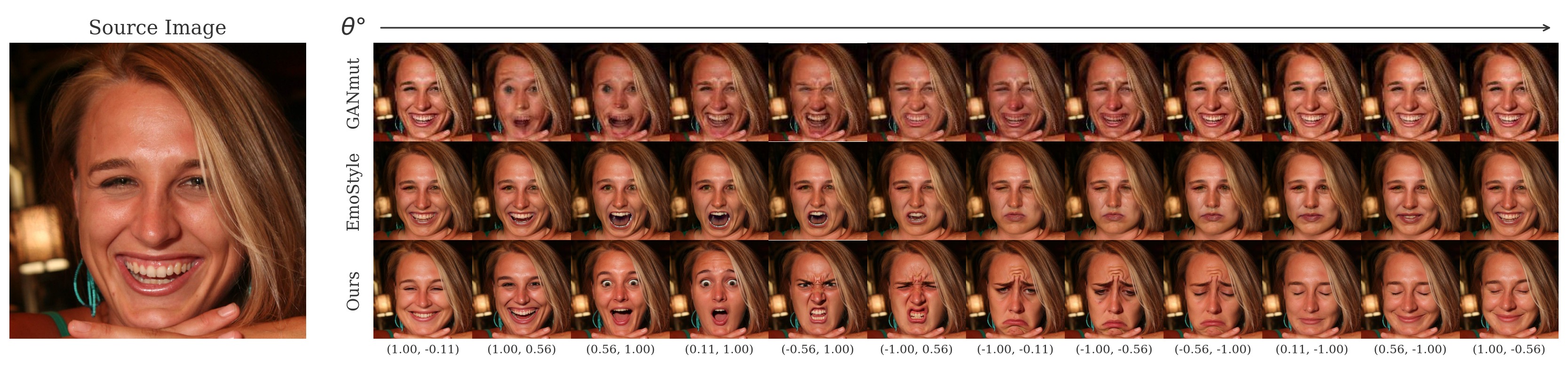}
    \label{fig:ffhq}
  \end{subfigure}
  \hfill
  \begin{subfigure}{0.99\linewidth}
  \centering
  \includegraphics[height=3.1cm]{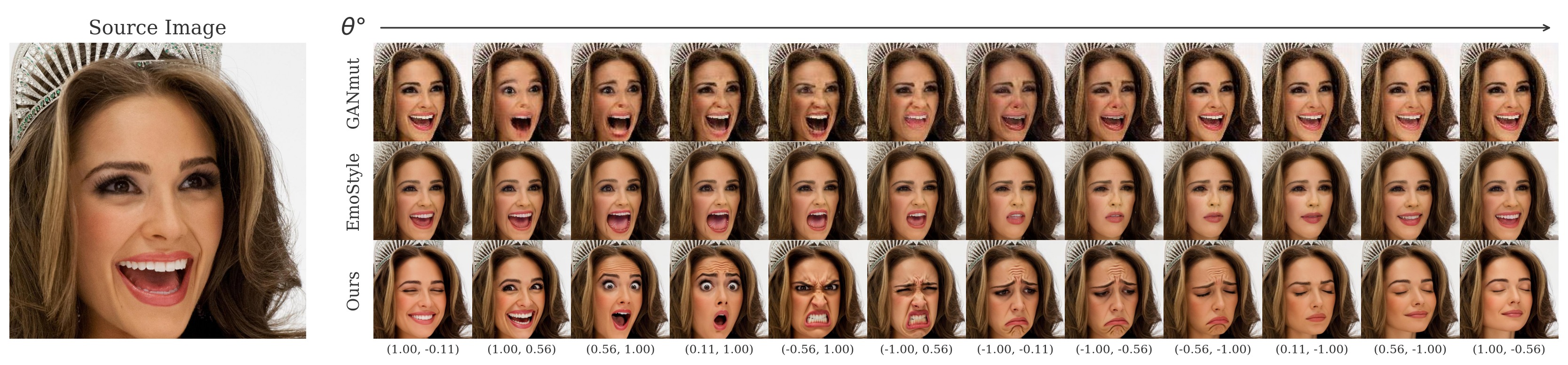}
    \label{fig:celeba}
  \end{subfigure}
  \caption{A sweep of VA values around the circumference of the unit circle, where $\theta$ is the angle. GANmut, EmoStyle, and Our edits are shown across three datasets AffectNet (top), FFHQ (middle), and CelebA (bottom). Target VA values (shown below columns) apply to EmoStyle and Ours only.} 
  \label{fig:qualitative_human_faces}
\end{figure}

\paragraph{Competing Methods} We compare identity-preserving facial expression editing against the two state-of-the-art methods GANmut \citep{d2021ganmut} and EmoStyle \citep{azari2024emostyle}. EmoStyle edits StyleGAN \citep{karras2019style} latents (trained on FFHQ) to control valence and arousal (VA), while GANmut learns its own two dimensional space from AffectNet. To better understand the generalization of each approach, we calculate metrics across three human face datasets: AffectNet \citep{mollahosseini2017affectnet}, FFHQ \cite{karras2019style}, and CelebA-HQ \citep{CelebAMask-HQ}. Following the recommendation from \citet{darne2025generating}, we use EmotiEffLib \citep{savchenko2023facial} to estimate VA values of each image. 
In Table \ref{tab:fee}, we see that our method outperforms competing methods on image quality, VA accuracy, expression diversity, and background preservation (LPIPS \cite{zhang2018unreasonable} on non-face regions). 
Building on the base models robust image quality and background preservation (indicated by FID, IS, and LPIPS), our approach maximizes VA accuracy and diversity while mitigating issues with edit smoothness. Our approach still lags behind in ID similarity, measured using ArcFace, which may in part be explained by our greater expression diversity. 
In line with lower LPIPS scores, Figure \ref{fig:qualitative_human_faces} demonstrates better background preservation (e.g., hair, accessories) alongside better coverage of extreme expressions, especially in the top-left VA quadrant (RMSE down 0.337 vs. EmoStyle) and for ``sleepy'' expressions, which, as noted by \citet{darne2025generating}, prior work struggles to capture. See Section~\ref{sec:gpt-image-2} for additional comparison to the closed source GPT-Image-2 model.

\revision{We also implement our approach on Qwen Image Edit 2511 and include results in Table~\ref{tab:fee} as an additional point of comparison. This implementation beats prior approaches in terms of VA accuracy and diversity, but also improves FID, IS, and ID similarity compared the FLUX.2-based implementation of our method. This demonstrates the potential of our approach to generalize across image editing models, enabling it to benefit from new model releases without retraining.}

\paragraph{Non-Human Faces} We show VA-based editing results in Figure \ref{fig:preview} and \ref{fig:face_preview} for a range of datasets \cite{CelebAMask-HQ,koklu2021manga,aneja2016modeling,choi2020stargan}. This shows that our method can be applied to animal faces, cartoon faces, and 3D models of faces as well as to regular human face images. In cases where face morphing cannot be applied, we use WAFT \citep{wang2025waft} to find the optical flow. 

\paragraph{Ablation Study} 
In Table~\ref{tab:ablation}, we perform an ablation study by removing or substituting components from our full approach. The full method achieves the best overall VA accuracy, diversity, and smoothness. Removing scaled CFG leads to a drop in VA accuracy and diversity. Substituting Barycentric Guidance with prompt interpolation further degrades VA control, significantly worsening RMSE and smoothness. Finally, removing face morphing results in the lowest smoothness and hence lower diversity. While this setting attains the highest image quality and identity similarity by staying closer to the input, it fails to produce precise affect changes. Overall, these results highlight the diversity–fidelity trade-off: removing components improves FID and ID similarity (by virtue of staying closer to the source), but our full approach is required to maximize VA coverage and control. Despite this, our method still outperforms prior VA-editing approaches on image quality metrics (see Table~\ref{tab:fee}), while achieving state-of-the-art VA accuracy, diversity, and smoothness.

\paragraph{User Study} 
We conduct a user study with \revision{66 participants} 
comparing EmoStyle and our method across 50 CelebA edits. See Section \ref{sec:additional_user_study} for full experimental setup. Results in Table \ref{tab:user_study} show a significant preference ($p<0.05$, Wilcoxon signed-rank test) for our approach on emotion accuracy (Q1) and non-affective content preservation (Q4). These results provide direct human evidence that our approach more accurately achieves the target affect while better preserving non-affective aspects of the image. While EmoStyle is preferred for realism (Q2) and identity (Q3), this is explained by edit magnitude: EmoStyle's average VA change of $|\Delta \mathrm{VA}| = 0.46$ is much lower than ours, $0.91$, against a target of $0.86$ (see Figure~\ref{fig:user_study_VA_changes}). This suggests EmoStyle remains closer to the input, which is consistent with higher realism and ID scores, whereas our method performs larger, more targeted affect changes, aligning with its advantage on VA accuracy and non-affective content preservation (see also examples in Figure~\ref{fig:user_study_example}). 
Due to cost, we do not conduct a full VA-plane user study. Instead, we evaluate a representative subset, with broader VA coverage and continuity assessed via grid-based quantitative metrics (see Tables \ref{tab:fee} \& \ref{tab:emoart}) and qualitative sweeps (see Figures \ref{fig:preview}, \ref{fig:qualitative_human_faces}, \ref{fig:face_preview}, \ref{fig:non_face_preview}, \ref{fig:grid_ffhq_ours} \& \ref{fig:grid_affectnet_ours}}).

\paragraph{Continuous Editing Comparison} Finally, we compare against other methods of continuous control for image editing models. \revision{Because existing approaches operate exclusively in the 1D setting, we restrict our evaluations to 1D interpolations for a direct comparison.} We compare against Kontinuous Kontext \citep{parihar2025kontinuous} (finetuned FLUX.1 Kontext), prompt interpolation, CFG-based control, and FluxSpace \citep{dalva2024fluxspace} (ported to FLUX.1 Kontext). For fair comparison, we reimplement our method on FLUX.1 Kontext. To measure smoothness of the edit we adapt the smoothness metric from \citep{d2021ganmut}  which involves interpolating between neutral and a target emotion over 10 steps. We measure LPIPS between consecutive images, reporting the max-to-mean LPIPS ratio, and additionally report RMSE using predicted VA values from EmotiEffLib. As shown in Table \ref{tab:continuous_control} and Figure \ref{fig:smoothness_plot}, our method yields smoother edits and lower RMSE, outperforming text-based, attention-based, and finetuning-based approaches.

\begin{table}
  \caption{Ablation Study on AffectNet. {$\leftrightarrows$: replaces Bary. Guidance with Prompt Interp.}}
  \label{tab:ablation}
  \centering
  \scriptsize
  \begin{tabular}{lcccc|ccc}
    \toprule
    Setting & \multicolumn{4}{c|}{\textbf{VA Controllability}} & \multicolumn{3}{c}{Image Quality / Preservation} \\
    \cmidrule(lr){2-5} \cmidrule(lr){6-8}
    & RMSE $\downarrow$  & VA Entropy $\uparrow$ & VA Std $\uparrow$ & $S^*_{2D}$ $\downarrow$ & FID $\downarrow$ & IS $\uparrow$ & ID Sim. $\uparrow$ \\
    \midrule
    Full Approach & \bf0.433 & \bf0.916 & {\bf0.64} / {\bf0.57} & \bf0.303 & 11.02 & 4.42 & {0.55} \\
    ~ $-$ ~Scaled CFG & \underline{0.439} & \underline{0.906} & \underline{0.59} / \underline{0.51} & \underline{0.305} & 11.10 & \underline{4.56} & \underline{0.61} \\
    ~ $\leftrightarrows$ Bary. Guidance $\rightarrow$ Prompt Interp. & 0.461 & 0.869 & 0.57 / 0.49 & 0.342 & \underline{10.75} & {4.50} & {0.55} \\
    ~ $-$ ~Face Morphing & 0.451 & 0.756 & 0.56 / 0.50 & 0.377 & \bf8.94 & {\bf4.62} & \bf0.66 \\
    \bottomrule
  \end{tabular}
\end{table}						

\begin{figure}[htbp]
  \centering
  
  \begin{minipage}{0.56\textwidth}
    \centering
    \captionof{table}{User study win rates between our approach and EmoStyle on CelebA facial expression editing. See Section \ref{sec:additional_user_study} for exact question wording.}
    \label{tab:user_study}
    \scriptsize 
    \begin{tabular}{lcccc}
      \toprule
      Question & EmoStyle & Ours & 
      Equal & p-value \\
      \midrule
      (1) VA Accuracy & \revision{9} & \revision{40} & \revision{1} & \revision{8.1e-6} \\ 
      (2) Realism & \revision{33} & \revision{15} & \revision{2} & \revision{2.9e-4} \\ 
      (3) ID Similarity & \revision{41} & \revision{8} & \revision{1} & \revision{8.6e-7} \\ 
      (4) Non-Face Pres. & \revision{11} & \revision{36} & \revision{3} & \revision{6.4e-5} \\ 
      \bottomrule
    \end{tabular}
    \hfill
    \captionof{table}{Continuous editing comparison on FLUX.1-Kontext for one dimensional editing.}
    \label{tab:continuous_control}
    \scriptsize 
    \begin{tabular}{lccc}
      \toprule
      Method & RMSE & Smooth. & C-LPIPS \\
      \midrule
      Prompt Interp. & 0.310 & 0.707 & 3.679 \\
      CFG & 0.316 & 0.720 & 3.987 \\
      FluxSpace & 0.476 & 0.686 & 3.490 \\
      Kontinuous Kontext & 0.321 & 0.665 & 3.552 \\
      Ours & \bf 0.292& \bf 0.643 & \bf 1.922  \\
      \bottomrule
    \end{tabular}
  \end{minipage}
  \hfill 
  \begin{minipage}{0.42\textwidth}
    \centering
    \includegraphics[width=\textwidth]{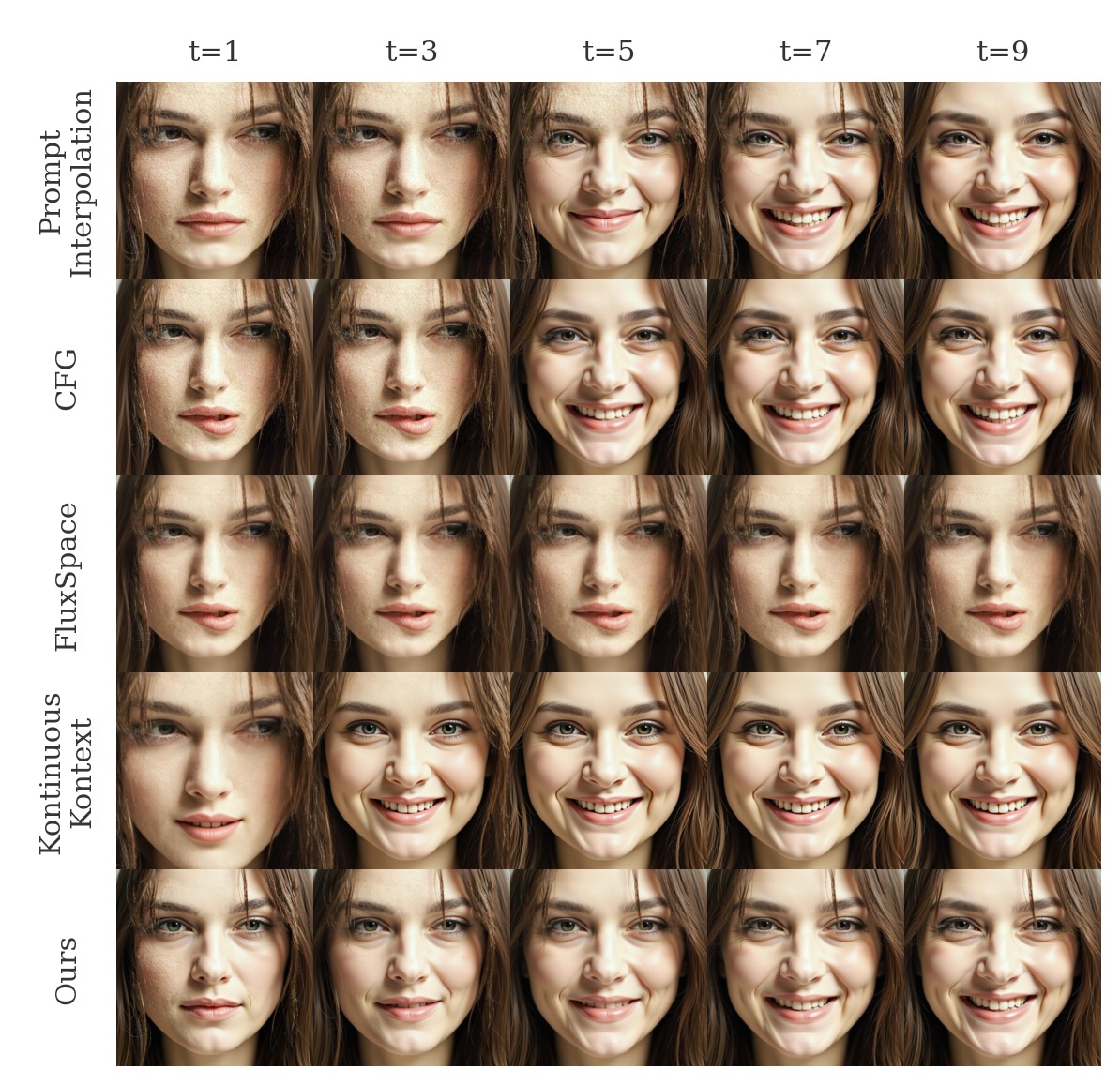}
    \caption{One dimensional continuous editing for FLUX.1 Kontext.}
    \label{fig:smoothness_plot}
  \end{minipage}
\end{figure}

\subsection{Non-Face Emotional Image Editing \& Generation}
\label{sec:non_fee}

In this section, we extend our approach to two non-face datasets: EmoArt \citep{zhang2025emoart} and FindingEmo \citep{mertens2024findingemo}.

\paragraph{EmoArt} EmoArt contains 130k images of artworks labeled with emotions designed to evenly cover the VA spectrum. Application of our method to EmoArt, shown in Figure \ref{fig:preview-f}, demonstrates that our VA-based image editing approach can effectively alter scenes, expressions, body language, and overall semantics.  Examples include adding angel wings for high valence/arousal, darkening scenes for sad emotions, portraying slumping postures for low arousal, and rendering angry expressions. Quantitative results in Table \ref{tab:emoart} show that, compared to prompt interpolation on the base model, our approach achieves superior alignment with the VA-model.

\begin{table}
  \caption{Non-face results. This includes VA-based editing of artworks on EmoArt \citep{zhang2025emoart} and VA-based generation of scenes using FindingEmo \citep{mertens2024findingemo}.}
  \label{tab:emoart}
  \centering
  \scriptsize
  \begin{tabular}{llcccc|cc}
    \toprule
    Dataset & Method & \multicolumn{4}{c|}{\textbf{VA Controllability}} & \multicolumn{2}{c}{Image Quality / Preservation} \\

    \cmidrule(lr){3-6} \cmidrule(lr){7-8}
    
     & & RMSE $\downarrow$ & VA Entropy $\uparrow$ & VA Std $\uparrow$ & $S^*_{2D}$ $\downarrow$ & FID $\downarrow$ & IS $\uparrow$ \\
    \midrule
    \multirow{2}{*}{EmoArt} & Prompt Interp. & 0.851 & 0.129 & 0.35 / 0.39 & 0.502 & 74.85 & \bf 9.64 \\
    & Ours & \bf 0.735 & \bf 0.327 & {\bf 0.36} / {\bf 0.42} & \bf 0.483 & \bf 67.96 & 9.08 \\
    \midrule
    \multirow{2}{*}{FindingEmo} & EmotiCrafter & 0.796 & \bf 0.298 & 0.38 / {\bf 0.25} & 0.540 & 233.30 & 8.85 \\
    &  Ours (klein) & \bf 0.783 & 0.163 & {\bf 0.40} / 0.21  & \bf 0.497 & \bf 221.51 & \bf 10.68  \\
    \bottomrule
  \end{tabular}
\end{table}



\paragraph{FindingEmo} FindingEmo contains 25k images of scenes labeled with a single VA value for the whole scene. We compare our approach against EmotiCrafter \cite{dang2025emoticrafter}, which generates images of scenes based on a text prompt and a VA label. We use FLUX.2 Klein 9B as our base model. In Table \ref{tab:emoart} we get competitive results, outperforming EmotiCrafter on both smoothness and image quality, while also requiring no training. Qualitative results are shown in Figure \ref{fig:non_face_preview-a}. We leave additional analysis of scene factors like lighting, composition, and pose as a compelling direction for future work.


\section{Conclusion}
\label{sec:conclusion}

In this work, we present an approach that enables {the foundation image editors} to perform valence-arousal-based editing and generation across a range of domains. To this end, we approximate the VA model using a triangle mesh. From this we derive weights, which can then be used for both guidance-based and latent-space-based editing. Results suggest this approach can achieve competitive performance across a wide variety of domains, without requiring retraining {the foundation models}.

\paragraph{Limitations \& Future Work} \revision{Although the proposed VA mesh can represent complex or mixed states to the extent that they are locally interpolable in VA space, the VA model does not capture the full spectrum of human emotion. For example, compound emotions such as bittersweet involve the simultaneous experience of happiness and sadness, which cannot be adequately represented by a single point in the two-dimensional VA space. In the supplementary materials, we present a simple extension to Barycentric Guidance that utilizes multiple VA points, thereby enabling the representation of compound emotions such as bittersweetness. These initial exploratory results suggest a promising direction for future work. Another avenue would be to extend the Barycentric Guidance framework to alternative models of emotion, such as the VAD model or appraisal-based models.}

While our approach does not require training, it may require some adaptation for new domains. Specifically, we found that different domains require different editing prompts. For domains that are more discrete, such as images of text, our continuous editing approach would also not be appropriate.

Finally, while our approach avoids expensive finetuning on individual images, our guidance and latent space editing algorithms add compute time in the form of increased diffusion steps. Future work may look to mitigate this via guidance-based distillation methods or application to few step image editing models.

\acksection

This research was mainly supported by Studentship from the Faculty of Engineering and Physical Sciences - Electronic and Computer Science at the University of Southampton. The
authors acknowledge the use of the IRIDIS X High Performance Computing Facility and the Southampton-Wolfson AI Research Machine (SWARM) GPU cluster, funded by the Wolfson Foundation, together with the associated support services at the University of Southampton in the completion of this work.

\medskip

{
    \small
    \bibliographystyle{unsrtnat}
    \bibliography{main}
}


\appendix

\newpage

\section{Appendix}

Here we provide more details about our method, including additional derivations, experiments, results, and experimental details. In Section~\ref{sec:appendix_qualitative} we present qualitative results with additional datasets and models. Section~\ref{sec:appendix_metrics} provides more derivations and experimental details that informed our metrics. Section~\ref{sec:morph_step} presents an experiment exploring the optimal step at which to restart diffusion. Section~\ref{sec:appendix_compound} extends the Barycentric Guidance method to capture compound emotions. Section~\ref{sec:appendix_trade_off} explores the controllability-fidelity trade-off for facial expression editing. Section~\ref{sec:gpt-image-2} compares our approach to a leading closed-source model. Section~\ref{sec:appendix_relative} details the relative improvements that were discussed in the abstract. Section~\ref{sec:derivations} goes over the derivation of Equations~\ref{eq:barycentric_editing} and \ref{eq:barycentric_generation}. Section~\ref{sec:exp_details} contains any experimental details we were not able to fit into the main body of the paper.

\begin{figure}[h!]
  \centering
  \begin{subfigure}{0.485\linewidth}
  \centering
  \includegraphics[height=7.0cm]{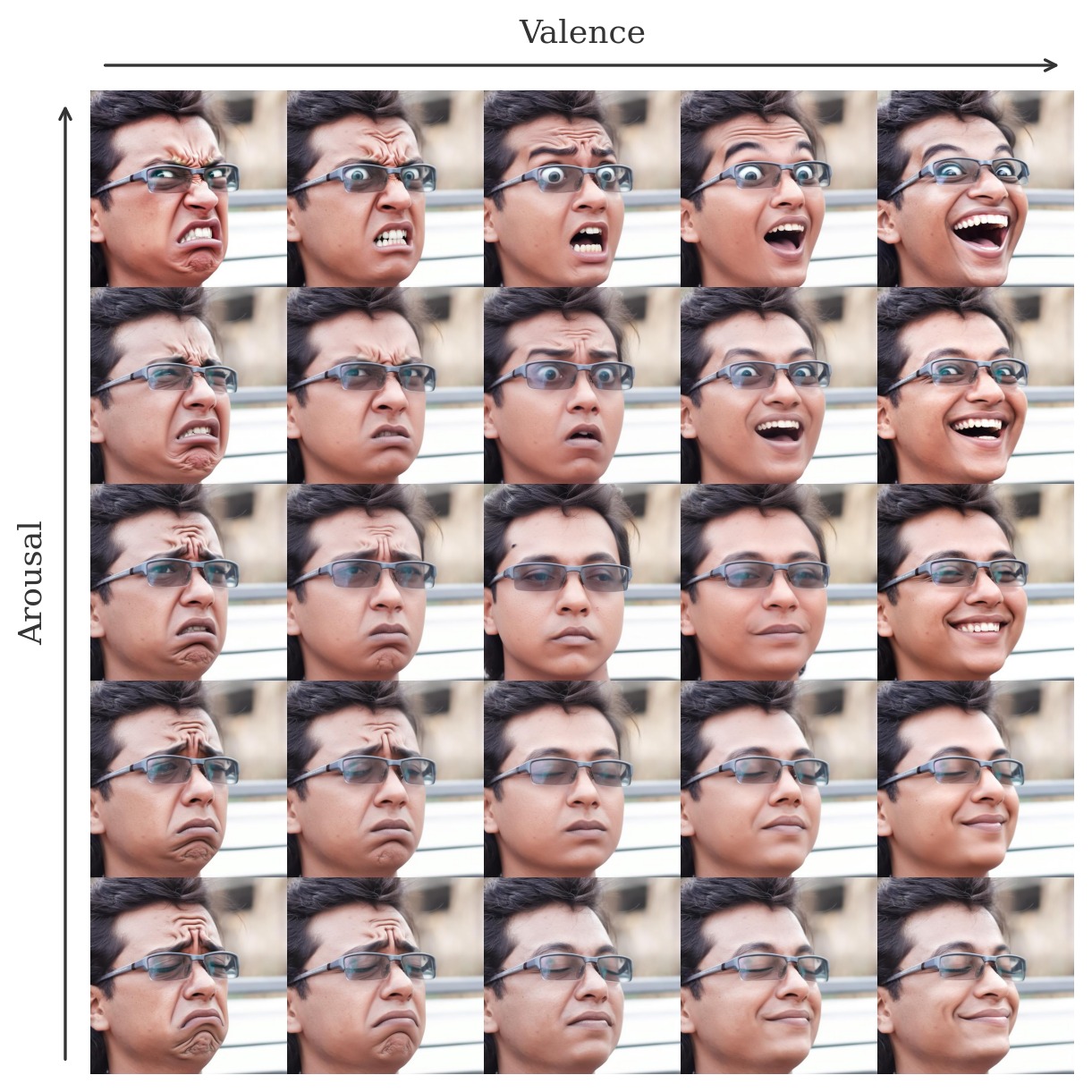}
    \caption{Human Faces (CelebA-HQ)}
    \label{fig:face_preview-b}
  \end{subfigure} \hfill
  \begin{subfigure}{0.485\linewidth}
  \centering
  \includegraphics[height=7.0cm]{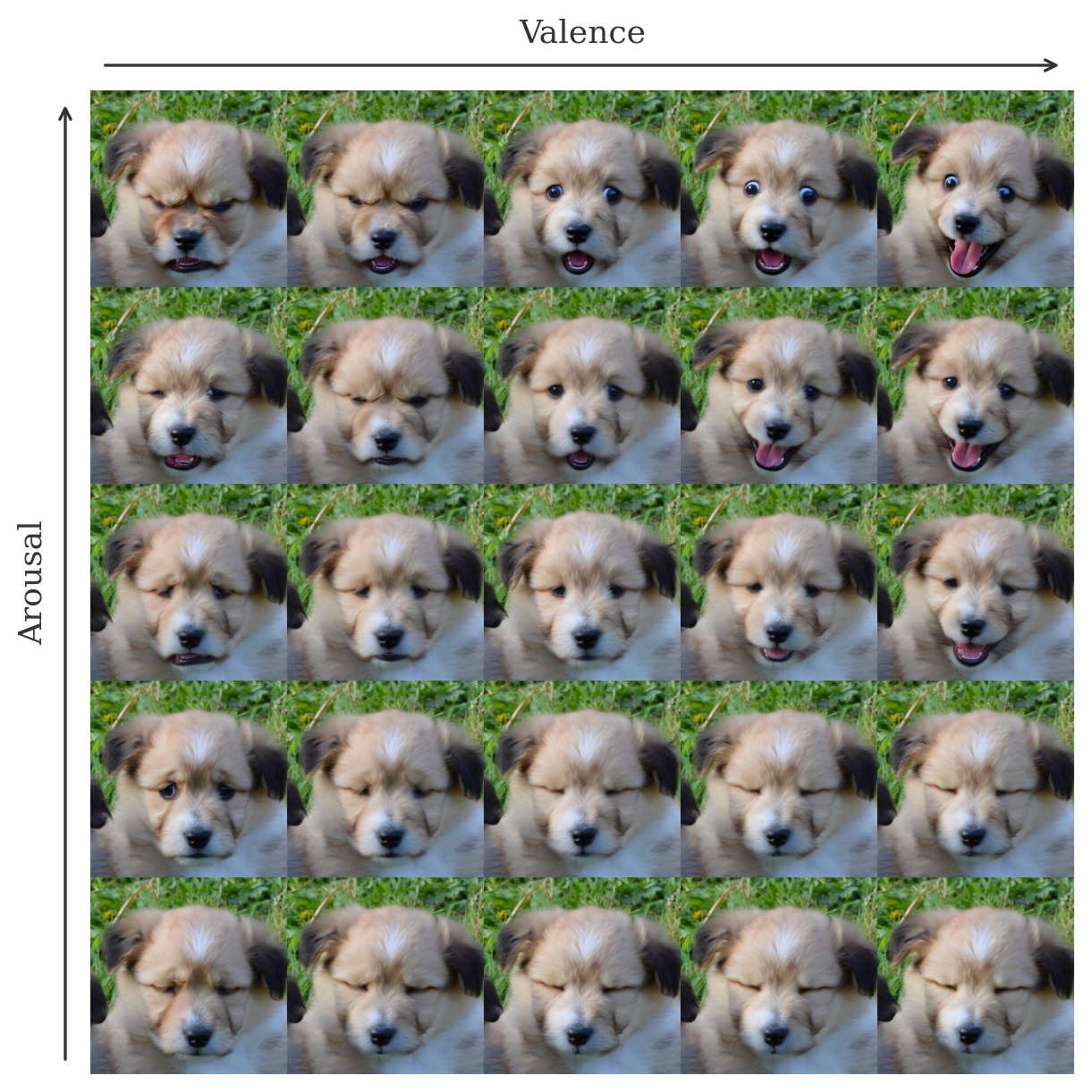}
    \caption{Animal Faces (AFHQ)}
    \label{fig:face_preview-c}
  \end{subfigure} 
   \\
  \begin{subfigure}{0.485\linewidth}
  \centering
  \includegraphics[height=7.0cm]{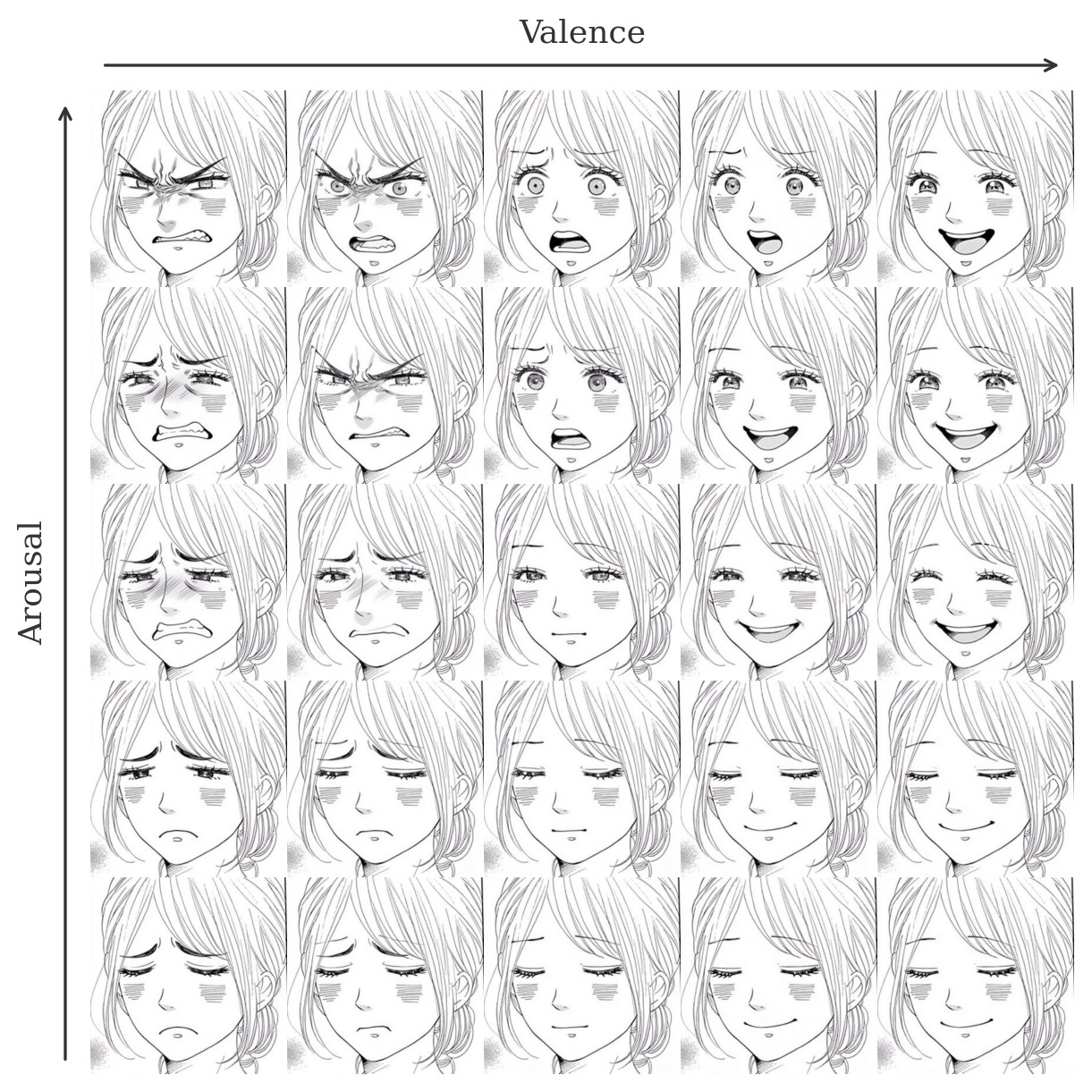}
    \caption{Cartoons (Manga Facial Expressions)}
    \label{fig:face_preview-d}
  \end{subfigure} \hfill
  \begin{subfigure}{0.485\linewidth}
  \centering
  \includegraphics[height=7.0cm]{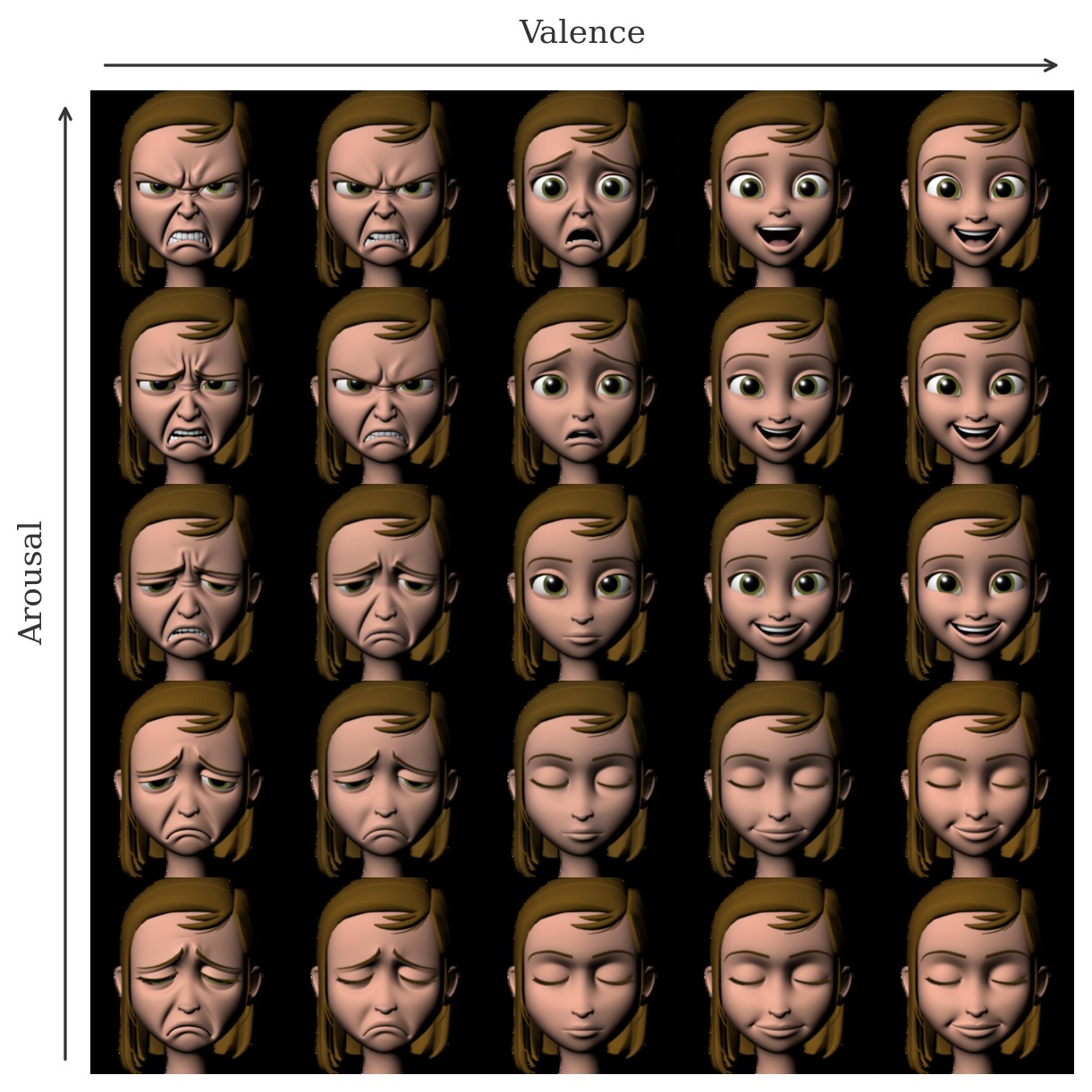}
    \caption{3D Models (FERG-DB)}
    \label{fig:face_preview-e}
  \end{subfigure}
  \caption{Our approach can perform valence arousal based emotional image editing across a wide variety of domains. VA range [-1,1] with uniform step size for rows and columns.} 
  \label{fig:face_preview}
\end{figure}

\subsection{Qualitative Results}\label{sec:appendix_qualitative} 

\begin{figure}[t]
  \centering
  \begin{subfigure}{0.485\linewidth}
  \centering
  \includegraphics[height=7.0cm]{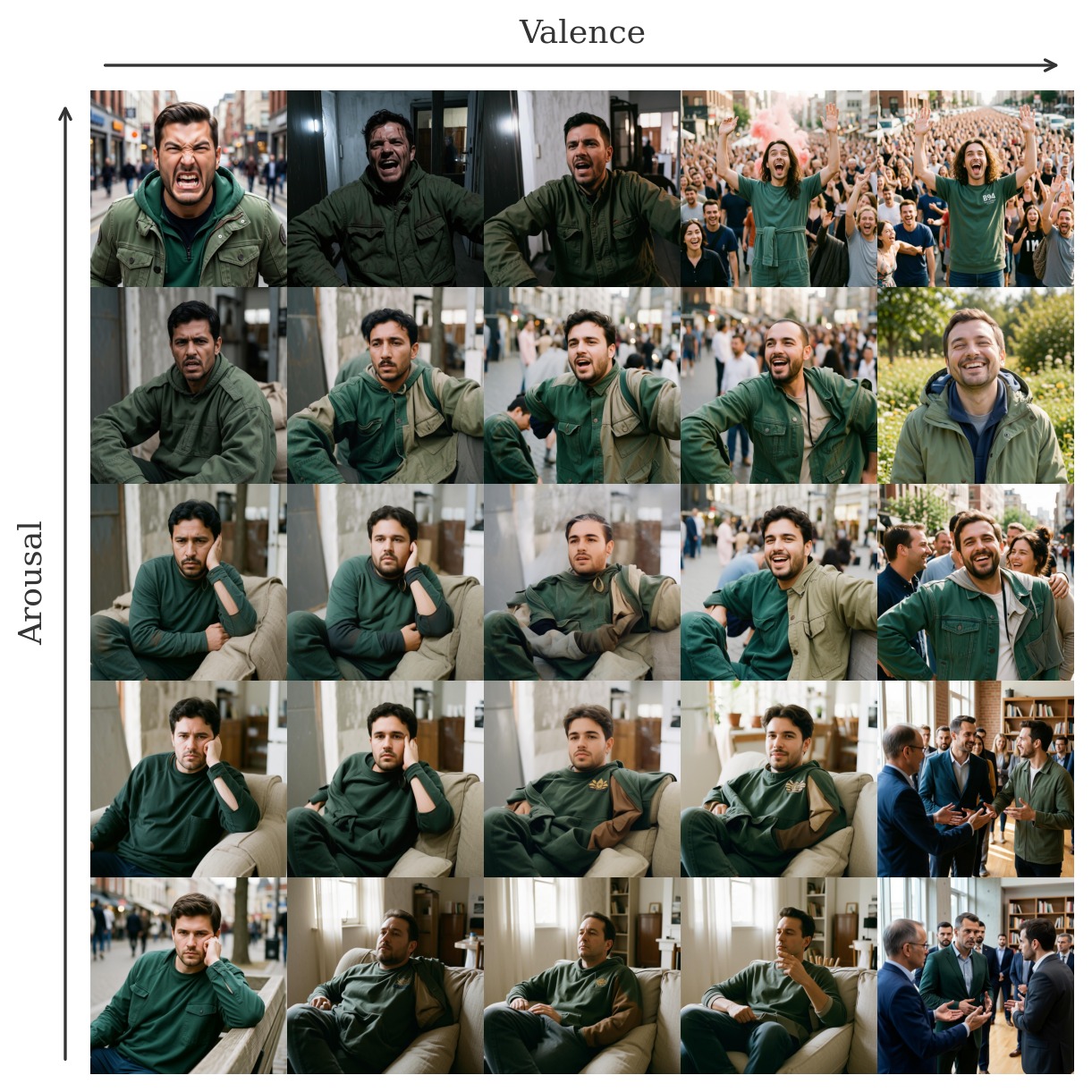}
    \caption{FindingEmo}
    \label{fig:non_face_preview-a}
  \end{subfigure} \hfill
  \begin{subfigure}{0.485\linewidth}
  \centering
  \includegraphics[height=7.0cm]{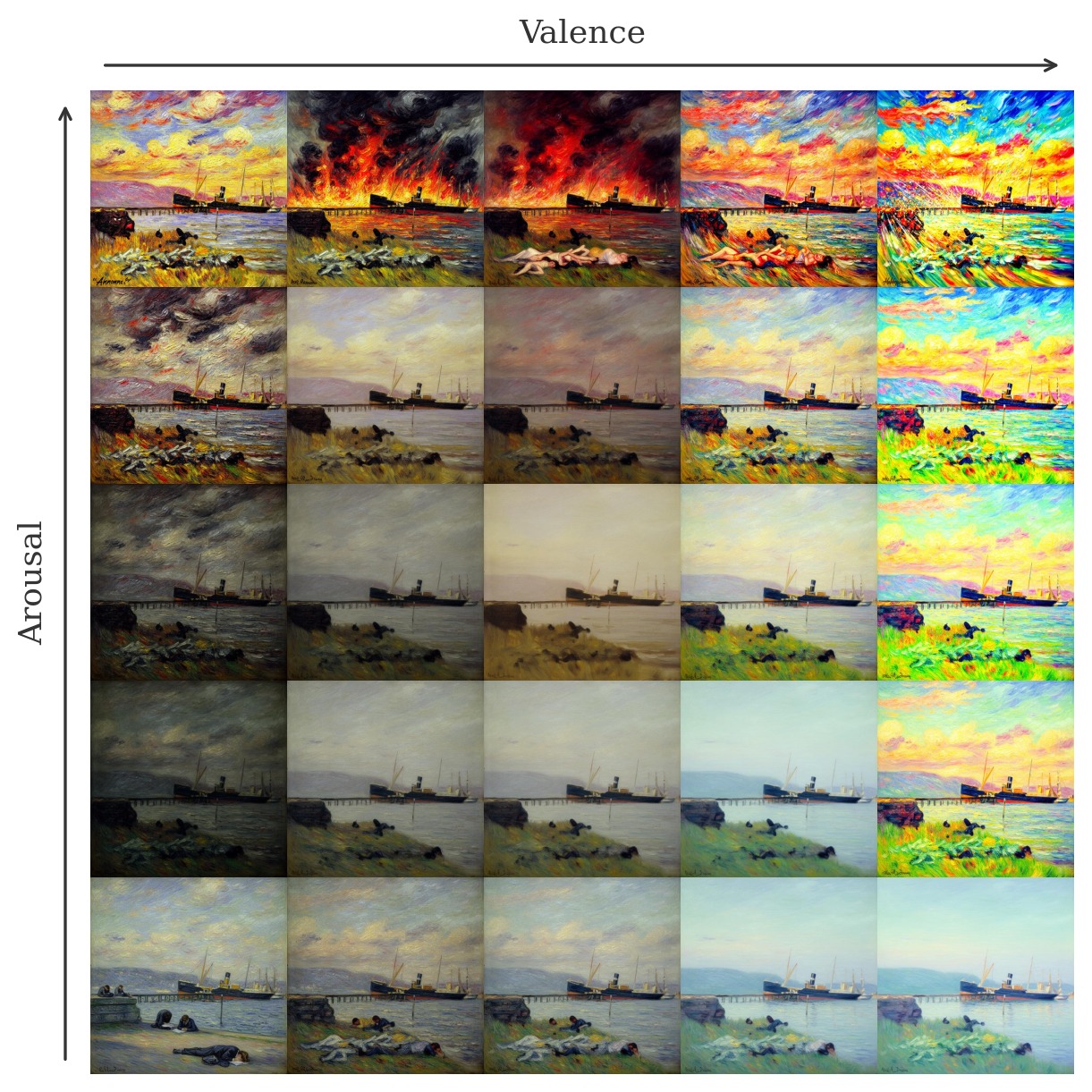}
    \caption{EmoArt}
    \label{fig:non_face_preview-b}
  \end{subfigure} 
   \\
  \caption{Qualitative results for non-face emotion image editing and generation. The base prompt used for {\bf(a)} is ``a man in a green jacket''. VA range [-1,1] with uniform step size for rows and columns}
  \label{fig:non_face_preview}
\end{figure}

In this section, we provide additional qualitative results. In Figure \ref{fig:face_preview}, we provide 5 by 5 grids showing a range of valence arousal edits for human face datasets and non-human face datasets. These include CelebA \citep{CelebAMask-HQ}, AFHQ \citep{choi2020stargan}, FERG-DB \citep{aneja2016modeling} and Manga Facial Expressions \citep{koklu2021manga}. In Figure \ref{fig:non_face_preview}, we show qualitative results for non-face datasets FindingEmo \citep{mertens2024findingemo} and EmoArt \citep{zhang2025emoart}.

\begin{figure}[t]
  \centering
  \begin{subfigure}{0.485\linewidth}
  \centering
  \includegraphics[height=7.0cm]{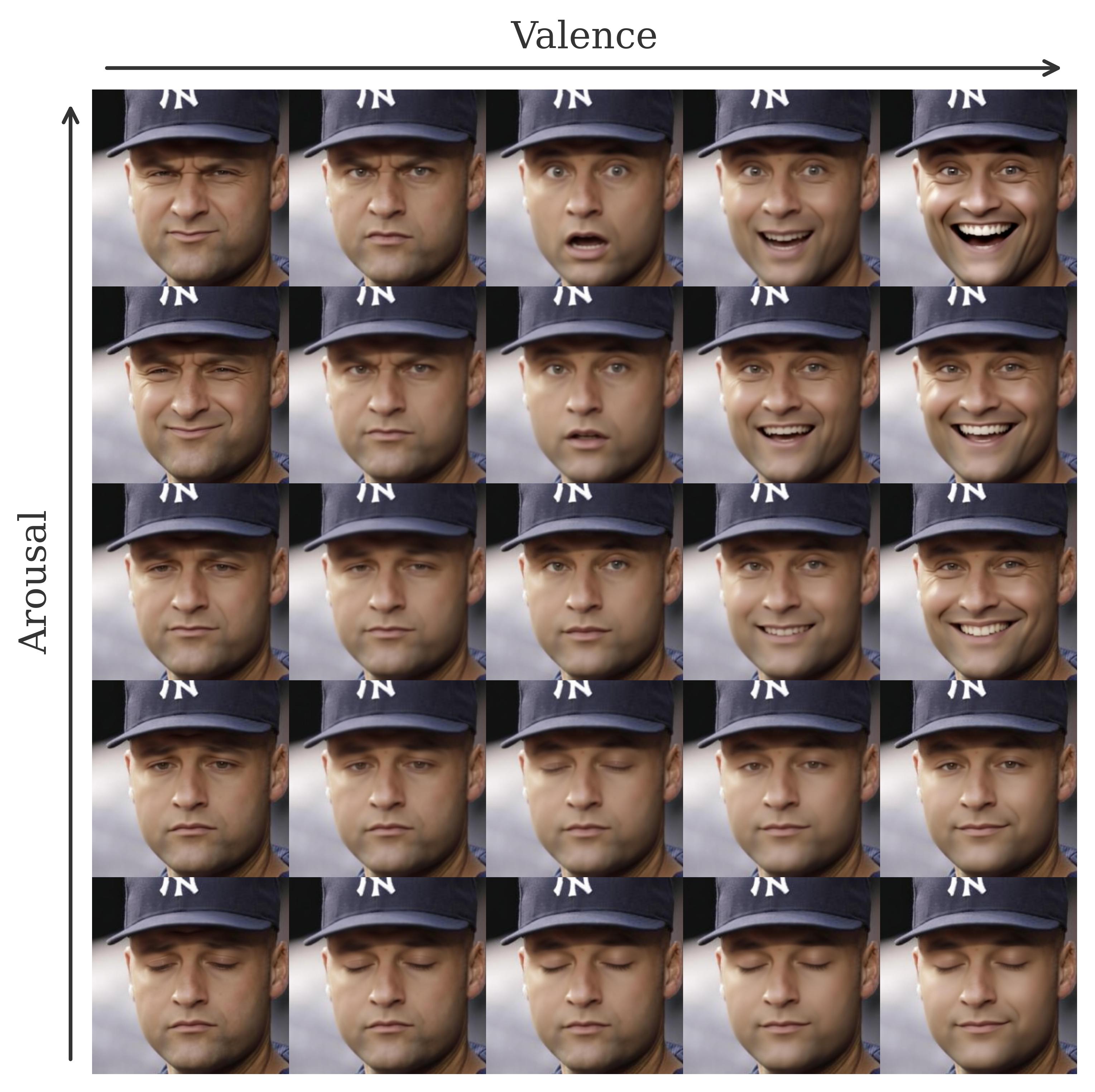}
  \end{subfigure} \hfill
  \begin{subfigure}{0.485\linewidth}
  \centering
  \includegraphics[height=7.0cm]{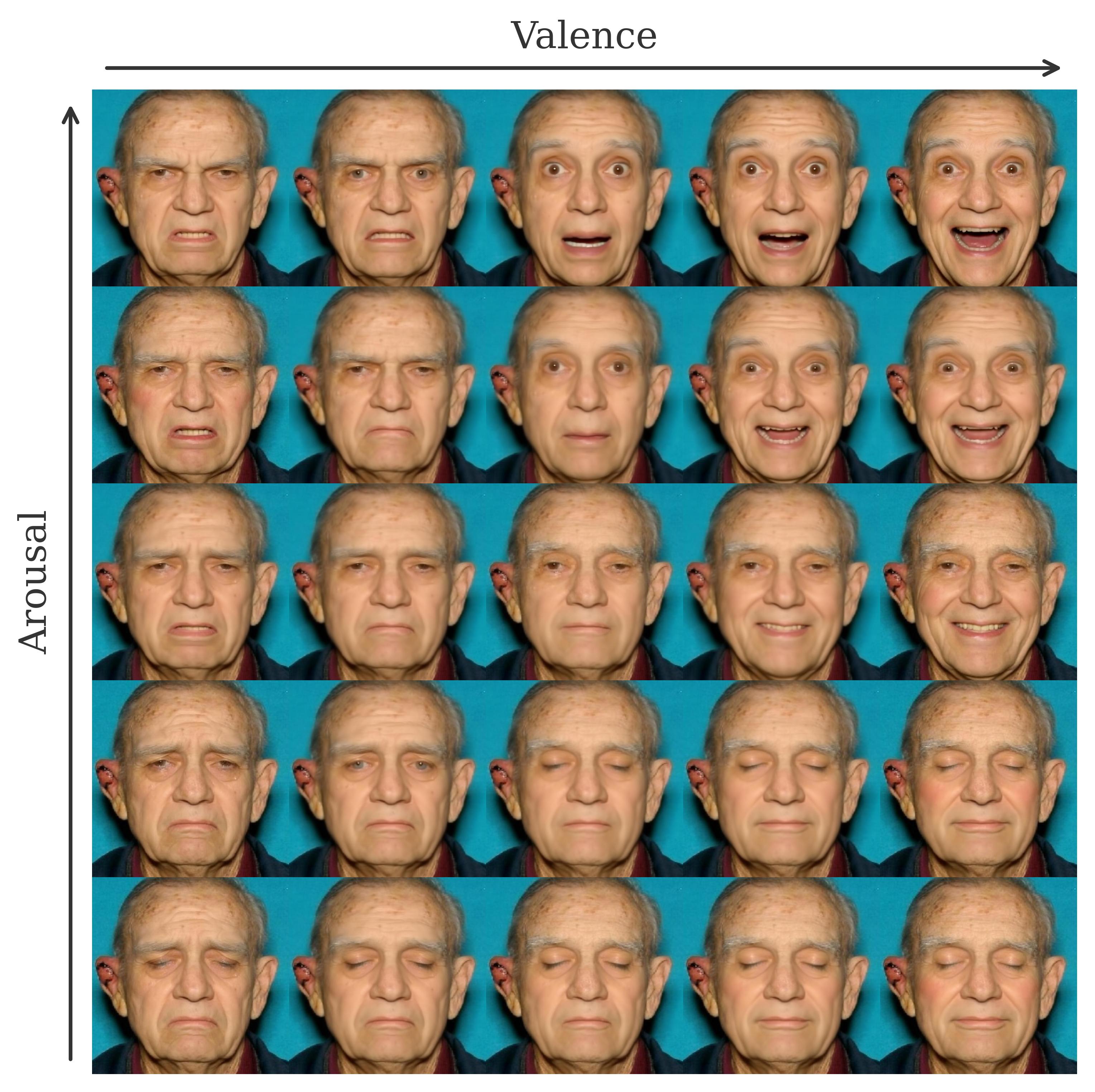}
  \end{subfigure} 
   \\
  \caption{Qualitative results on AffectNet using Qwen Image Edit 2511 as our base model. VA range [-1,1] with uniform step size for rows and columns}
  \label{fig:qwen_image_edit}
\end{figure}

In Figure \ref{fig:qwen_image_edit}, we apply our full approach to Qwen Image Edit 2511 \citep{wu2025qwenimagetechnicalreport}. These qualitative results show that our method can be applied beyond the FLUX family of models. Throughout the paper, we apply our approach to FLUX.2 dev, FLUX.2 klein, FLUX.1 Kontext, and Qwen Image Edit 2511, suggesting that our approach can be applied broadly across diffusion-based foundation image editing models.

\begin{figure}[t]
  \centering
  \includegraphics[width=\linewidth]{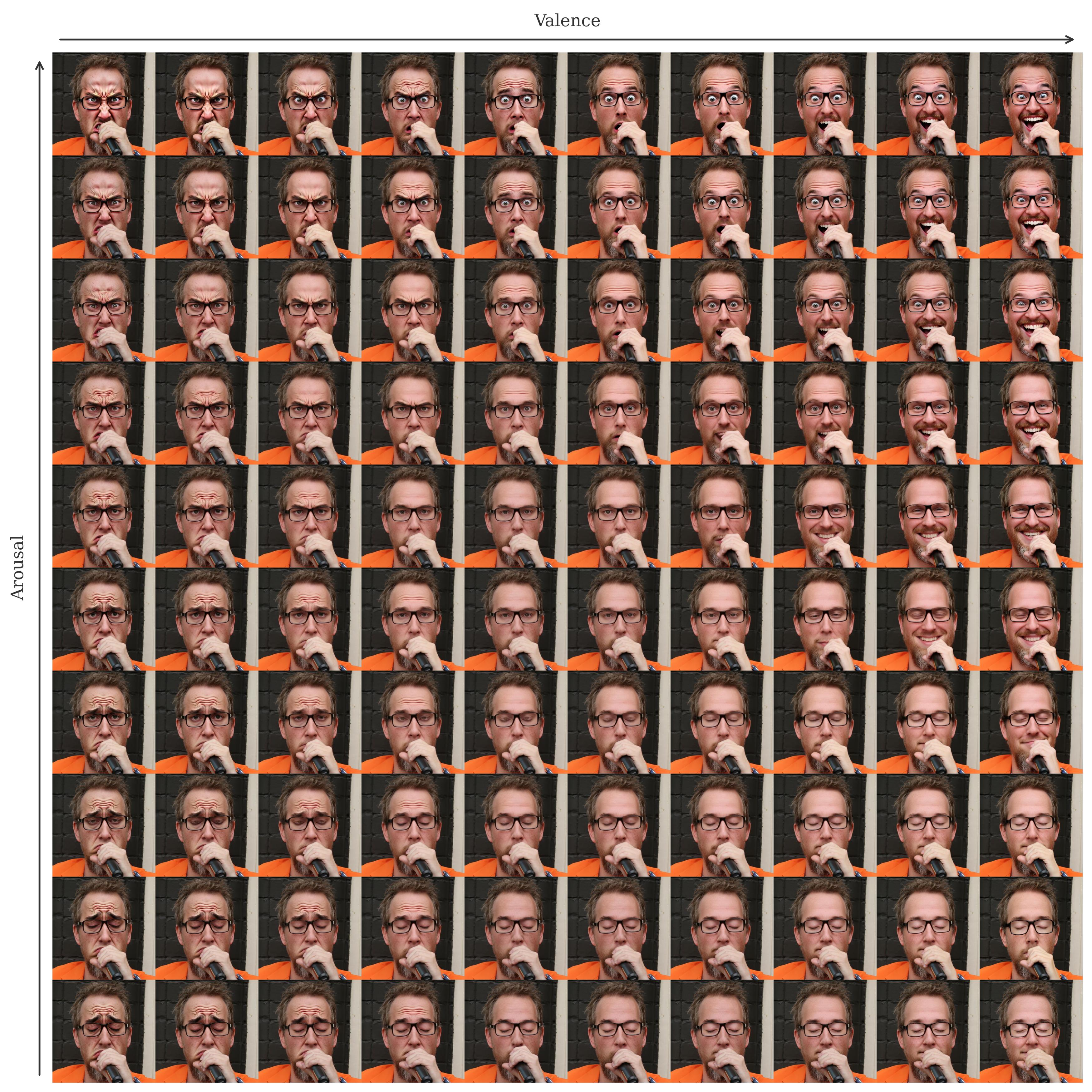}
  \caption{Our approach on FFHQ. VA range [-1,1] with uniform step size for rows and columns.}
  \label{fig:grid_ffhq_ours}
\end{figure}

\begin{figure}[t]
  \centering
  \includegraphics[width=\linewidth]{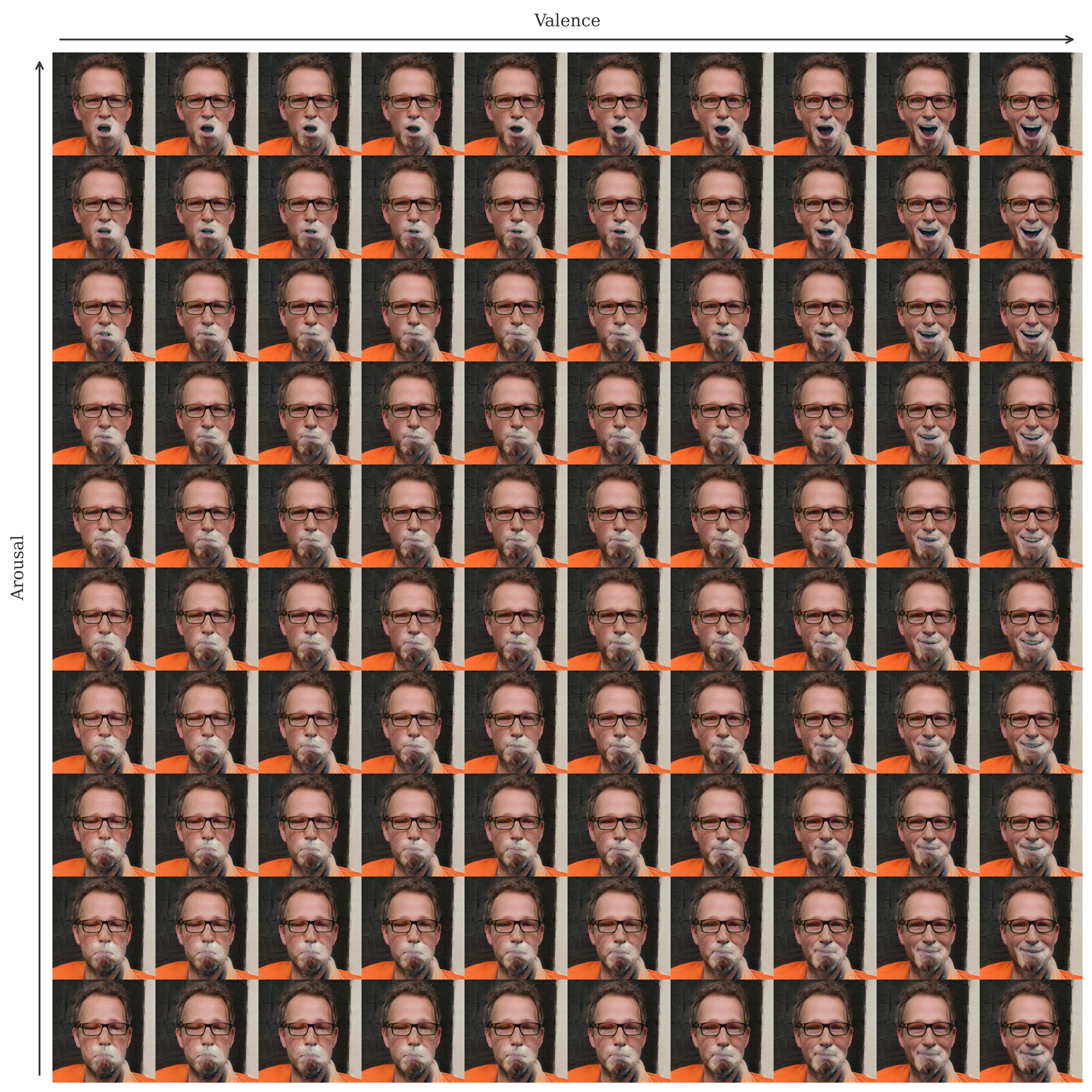}
  \caption{EmoStyle on FFHQ. VA range [-1,1] with uniform step size for rows and columns.}
  \label{fig:grid_ffhq_emostyle}
\end{figure}

\begin{figure}[t]
  \centering
  \includegraphics[width=\linewidth]{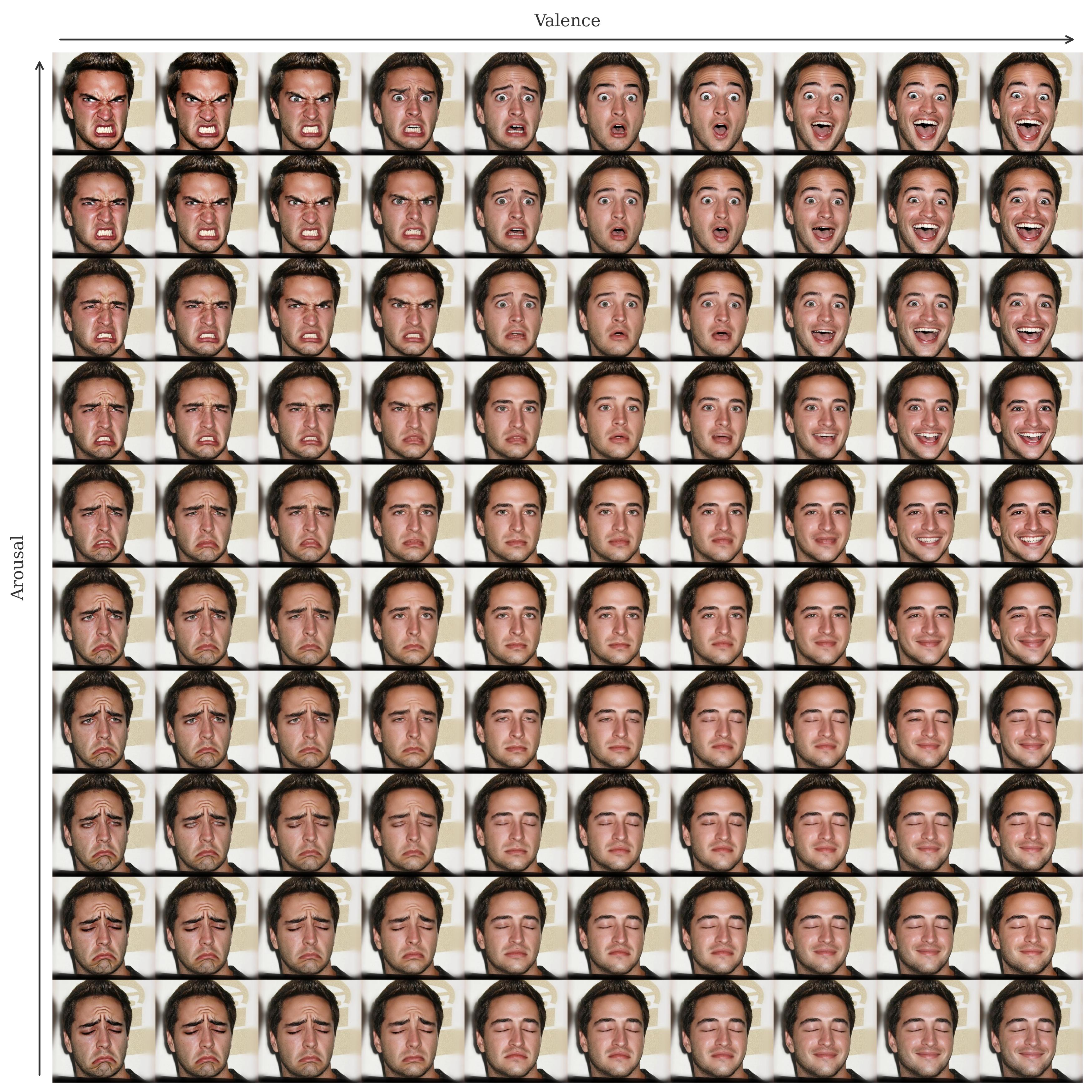}
  \caption{Our approach on CelebA. VA range [-1,1] with uniform step size for rows and columns.}
  \label{fig:grid_celeba_ours}
\end{figure}

\begin{figure}[t]
  \centering
  \includegraphics[width=\linewidth]{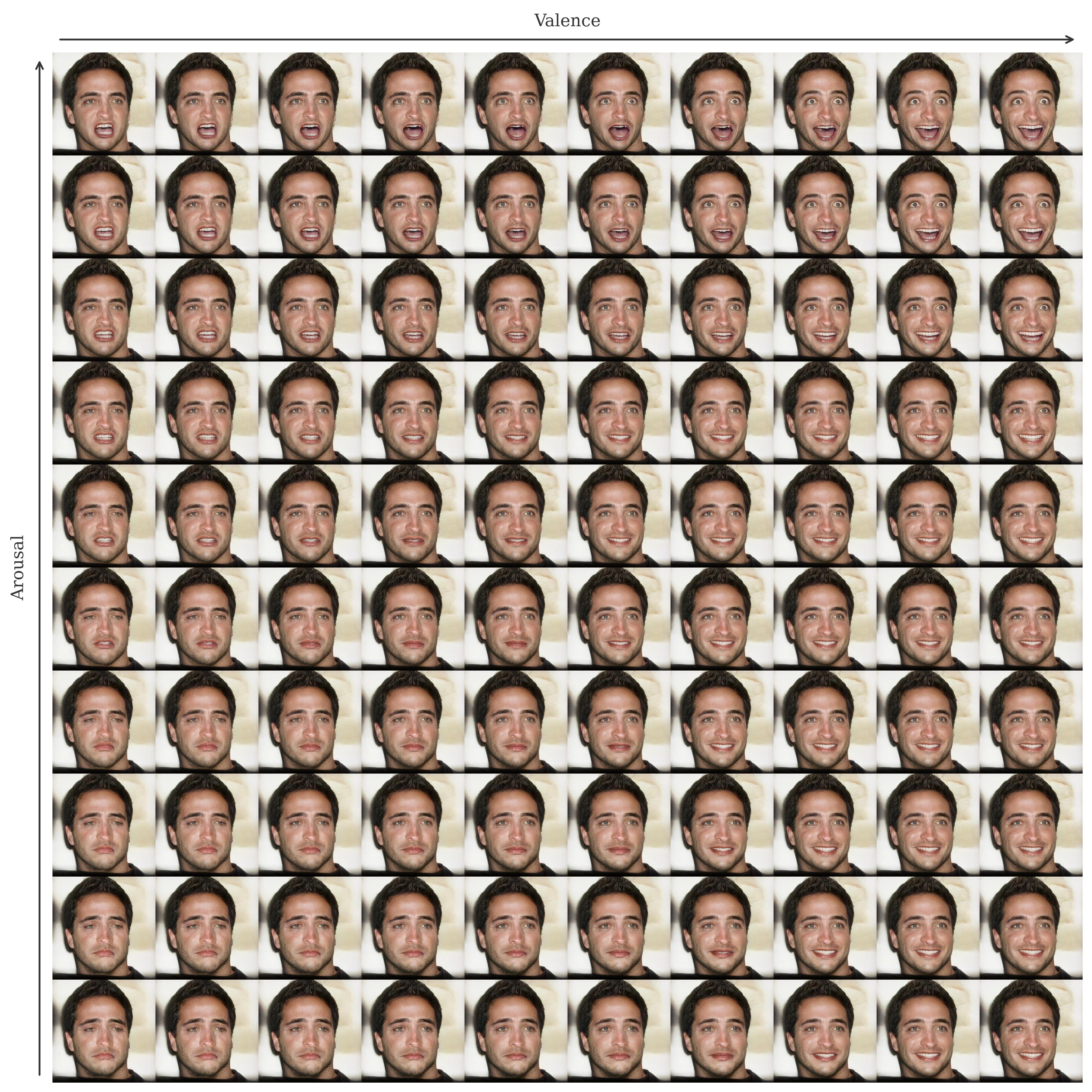}
  \caption{EmoStyle on CelebA. VA range [-1,1] with uniform step size for rows and columns.}
  \label{fig:grid_celeba_emostyle}
\end{figure}

\begin{figure}[t]
  \centering
  \includegraphics[width=\linewidth]{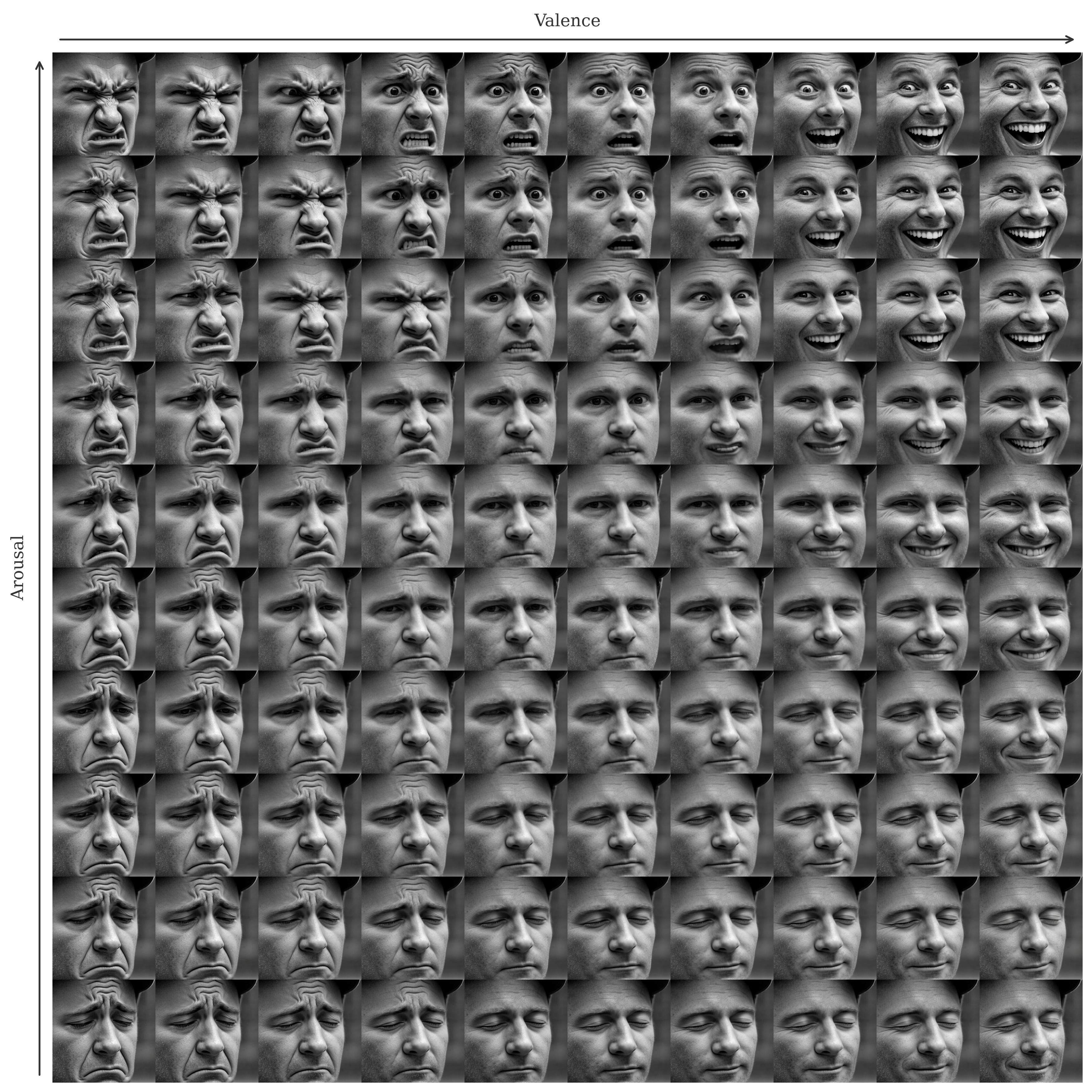}
  \caption{Our approach on AffectNet. VA range [-1,1] with uniform step size for rows and columns.}
  \label{fig:grid_affectnet_ours}
\end{figure}

\begin{figure}[t]
  \centering
  \includegraphics[width=\linewidth]{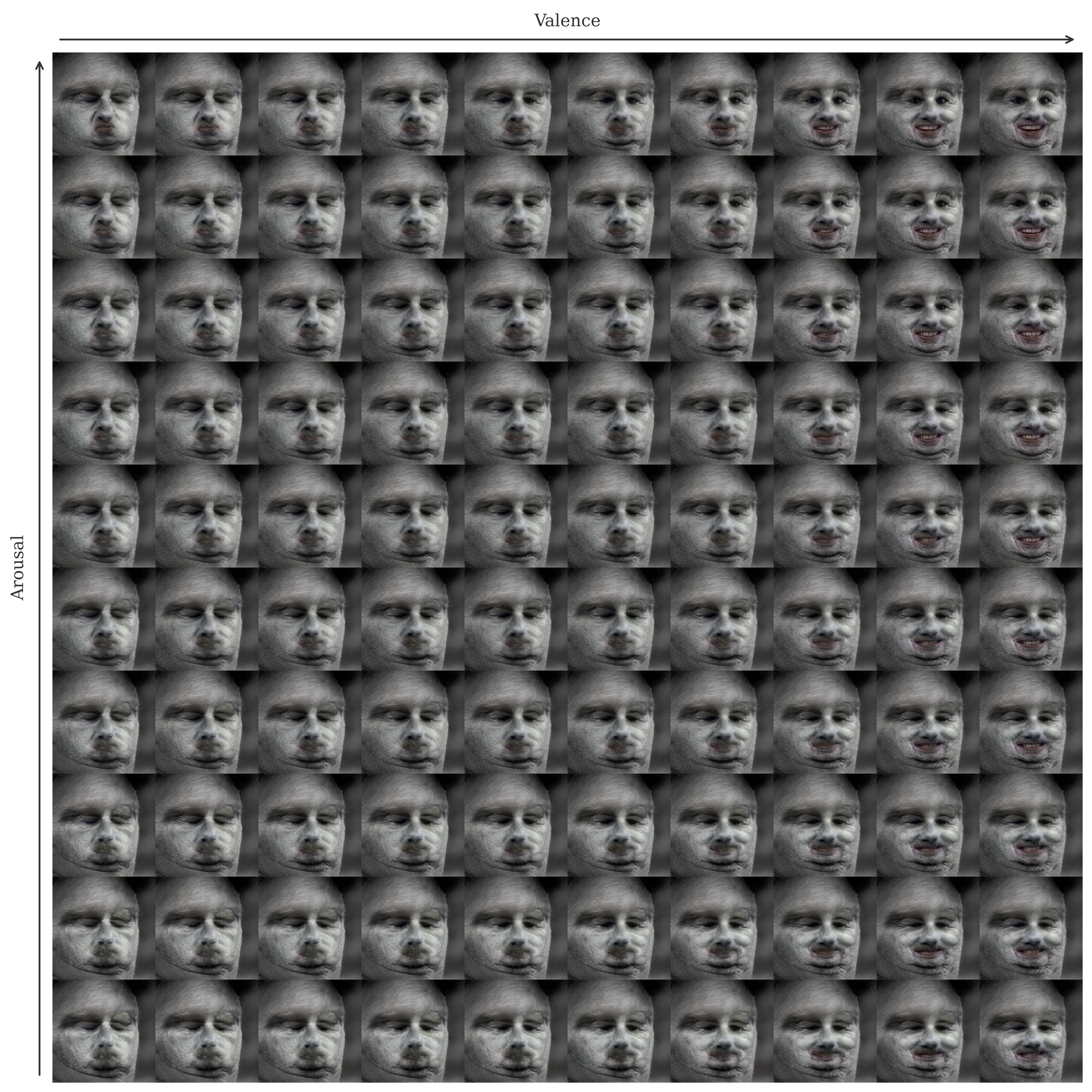}
  \caption{EmoStyle on AffectNet. VA range [-1,1] with uniform step size for rows and columns.}
  \label{fig:grid_affectnet_emostyle}
\end{figure}

We also provide a further qualitative comparison between our method and EmoStyle \citep{azari2024emostyle} on AffectNet \citep{mollahosseini2017affectnet}, CelebA \citep{CelebAMask-HQ}, and FFHQ \citep{karras2019style} in Figures \ref{fig:grid_affectnet_ours}, \ref{fig:grid_affectnet_emostyle}, \ref{fig:grid_celeba_ours}, \ref{fig:grid_celeba_emostyle}, \ref{fig:grid_ffhq_ours}, and \ref{fig:grid_ffhq_emostyle} for the full 10 by 10 grids used for VA focused metrics. These figures demonstrate more advantages of our method. Firstly, our approach is much better able to represent low arousal emotions. Secondly, our approach is better able to handle occlusion of the face convincingly. Finally, we find that EmoStyle does not deal with cropped faces well, which are common in the AffectNet dataset. Our approach on the other hand is robust to the cropping of the face.

\subsection{More Information on Metrics}
\label{sec:appendix_metrics}

This section provides further experimentation and discussion to further support our choice of metrics.

\begin{figure}
  \centering
  \includegraphics[height=6.2cm]{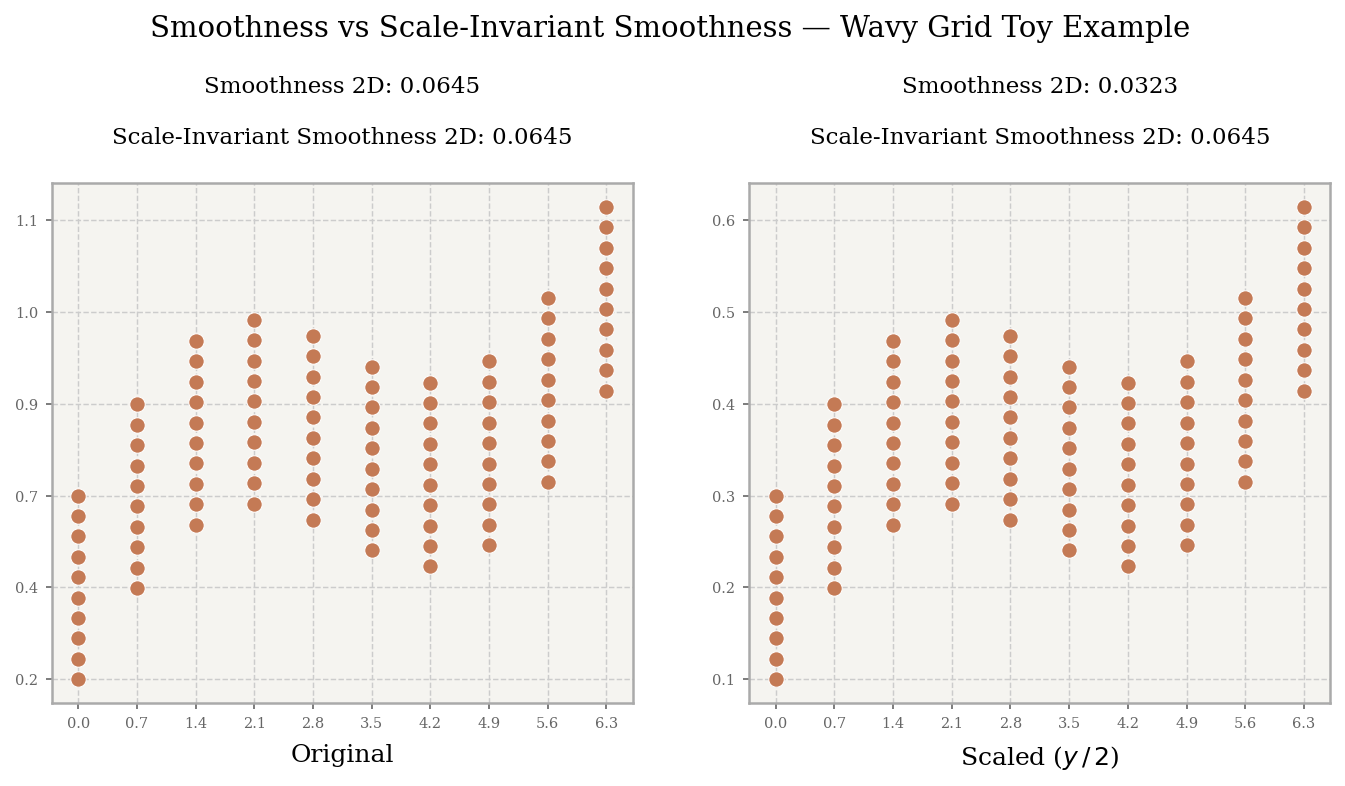}
\caption{Smoothness Metric Toy Example}
  \label{fig:smoothness_toy_example}
\end{figure}

\paragraph{Smoothness} To evaluate the smoothness of expression editing over the entire VA space we propose using a Laplacian to measure the curvature of VA predictions of a 10 by 10 grid of image with different uniformly spaced target VA targets.  Consider a function $z(x,y)$ which maps from a point on a square lattice $(x,y)$ to a valence arousal point $(v,a)$. From this we can estimate the curvature using a discrete Laplacian, approximated using finite differences, as
\begin{align}
\Delta z(x,y) = z(x+1,y)+z(x-1,y)+z(x,y+1)+z(x,y-1)-4z(x,y).
\end{align}
To measure the overall smoothness of our valence and arousal surface, we take the magnitude of these Laplacians
\begin{align}
S_{\mathrm{2D}} &= \frac{1}{(W-2)(H-2)}\sum_{i=2}^{H-1}\sum_{j=2}^{W-1}\sqrt{\Delta z^v(x,y)^2 + \Delta z^a(x,y)^2},
\end{align}
where $W$ and $H$ are the width and height of the grid (i.e. $W=10$ and $H=10$), and $\Delta z^v$ and $\Delta z^a$ are the valence and arousal components of $\Delta z$, respectively. However, if we reduce the range of one dimension, we will reduce the smoothness, even though the rate of change with respect to the total range remains unchanged. In practice, as seen in the toy example in Figure \ref{fig:smoothness_toy_example}, the same set as valence arousal points will results is different $S_{2D}$ values when scaled up. To account for the fact that different models will generated images with different ranges of valence and arousal predictions, we can scale the valence values by the range of valence values $r_v$, and the arousal values by the range of arousal values $r_a$. This allows to define a scale-invariant version of the $S_{2D}$ metric
\begin{align}
S^*_{\mathrm{2D}} &= \frac{1}{(W-2)(H-2)}\sum_{i=2}^{H-1}\sum_{j=2}^{W-1}\sqrt{\left(\frac{\Delta z^v(x,y)}{r_v}\right)^2 + \left(\frac{\Delta z^a(x,y)}{r_a}\right)^2}.
\end{align}
Now consider a scaled version of the same valence arousal points. We can scale the valence dimension by $\alpha$ and the arousal dimension by $\beta$, which means our estimate of curvature becomes

\begin{align}
\begin{pmatrix} \alpha \\ \beta \end{pmatrix} \cdot\Delta z_{i,j} = \begin{pmatrix} \alpha \\ \beta \end{pmatrix} \cdot (z_{i+1,j}+z_{i-1,j}+z_{i,j+1}+z_{i,j-1}-4z_{i,j}).
\end{align}
Since scaling the the dimensions also scales the minimum and maximum values by the same amount, the range is also scaled by the same amount. Our smoothness metric is then 

\begin{align}
S^*_{\mathrm{2D}} = \frac{1}{(W-2)(H-2)}\sum_{i=2}^{H-1}\sum_{j=2}^{W-1} \sqrt{\left(\frac{\alpha \Delta z^v(x,y)}{\alpha r_v}\right)^2+\left(\frac{\beta \Delta z^a(x,y)}{\beta r_a}\right)^2}.
\end{align}

The scaling factors therefore cancel out for both dimensions leaving us with our original $S^*_{\mathrm{2D}}$ metric. This ensures that our metric is scale invariant and does not unfairly punish more diverse sets of valence arousal points.

\begin{figure}[t]
  \centering
  \includegraphics[height=5.0cm]{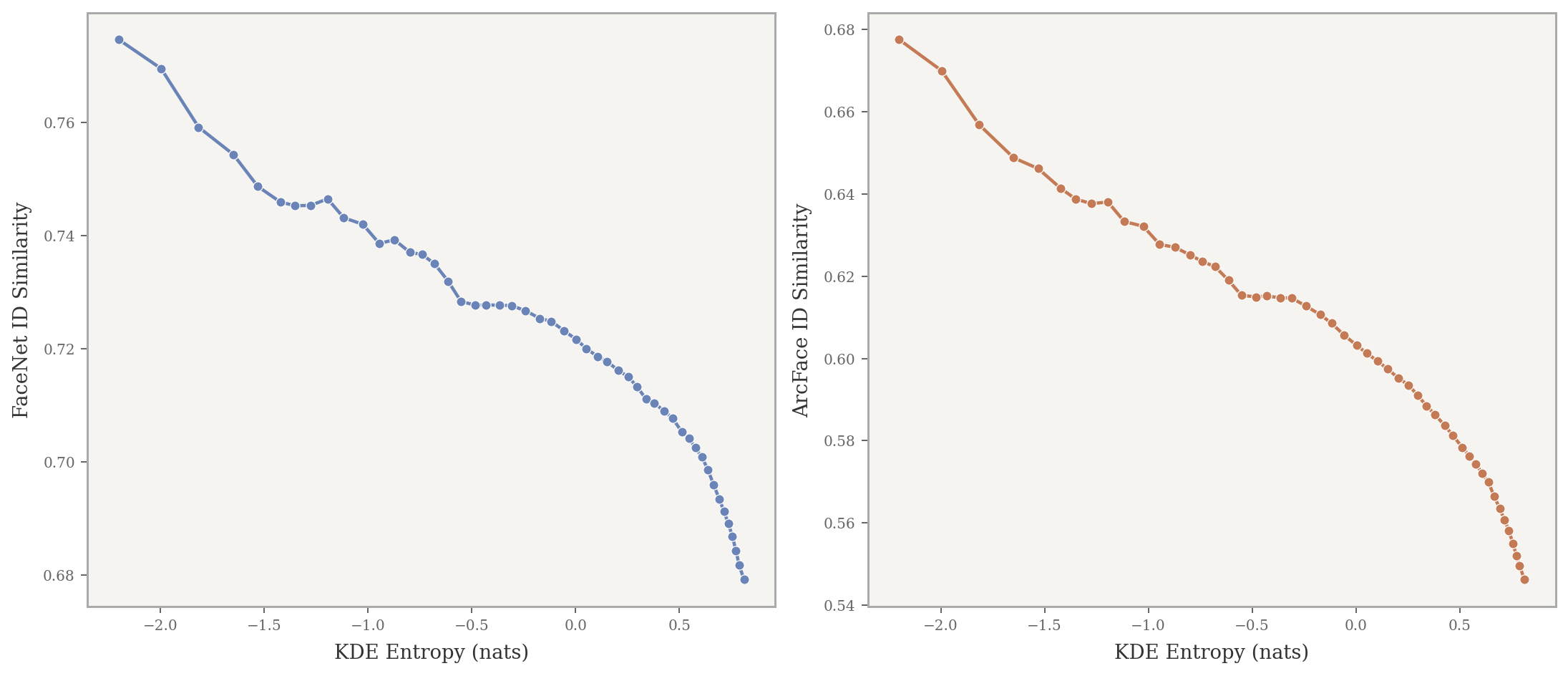}
  \caption{On synthetic images, diversity of facial expression trades-off against ID similarity as measured by FaceNet and ArcFace.}
  \label{fig:trade_off_synth}
\end{figure}

\paragraph{Divserity} Here we further verify the trade-off between facial expression diversity and ID similarity metrics. In Figure \ref{fig:trade_off_synth} we examine this trade-off on synthetic images using 3500 images generated from AffectNet validation images with random VA values with our full method, except with 14 inference steps and 7 anchor steps. We measure ID similarity using FaceNet and ArcFace between the source and generated images. We reduce the entropy by calculating the radius of each predicted VA values then gradually removing the highest radius points over the course of 50 steps. Once points have been removed, we recalculate the entropy of the remaining VA points and also take the average ArcFace and FaceNet similarity of the remaining points. Figure \ref{fig:trade_off_synth} shows that for the synthetic images, ID similarity clearly trades-off against facial expression diversity. As well as posing problems for evaluation, this also could also cause potential problems when optimizing a facial expression editing model using a facial recognition model.

\begin{figure}[htbp]
  \centering
  \begin{minipage}{0.24\linewidth}
    \centering
    \includegraphics[height=3.0cm]{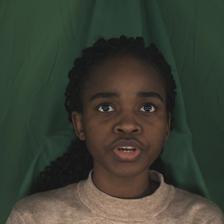}\\
    \small Neutral
  \end{minipage}
  \hfill
  \begin{minipage}{0.24\linewidth}
    \centering
    \includegraphics[height=3.0cm]{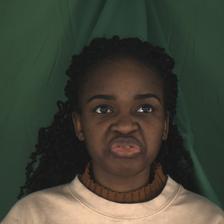}\\
    \small Disgust (Weak)
  \end{minipage}
  \hfill
  \begin{minipage}{0.24\linewidth}
    \centering
    \includegraphics[height=3.0cm]{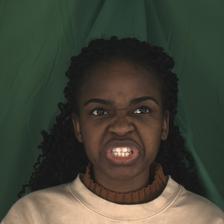}\\
    \small Disgust (Medium)
  \end{minipage}
  \hfill
  \begin{minipage}{0.24\linewidth}
    \centering
    \includegraphics[height=3.0cm]{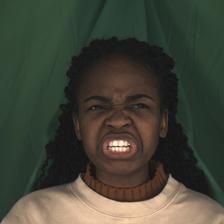}\\
    \small Disgust (Strong)
  \end{minipage}
  \caption{Comparison of facial expression intensity levels in the MEAD dataset, demonstrating the incremental progression from neutral to strong disgust.}
  \label{fig:mead_examples}
\end{figure}

\subsection{Morph Step}
\label{sec:morph_step}

\begin{figure}
  \centering
  \includegraphics[height=8.0cm]{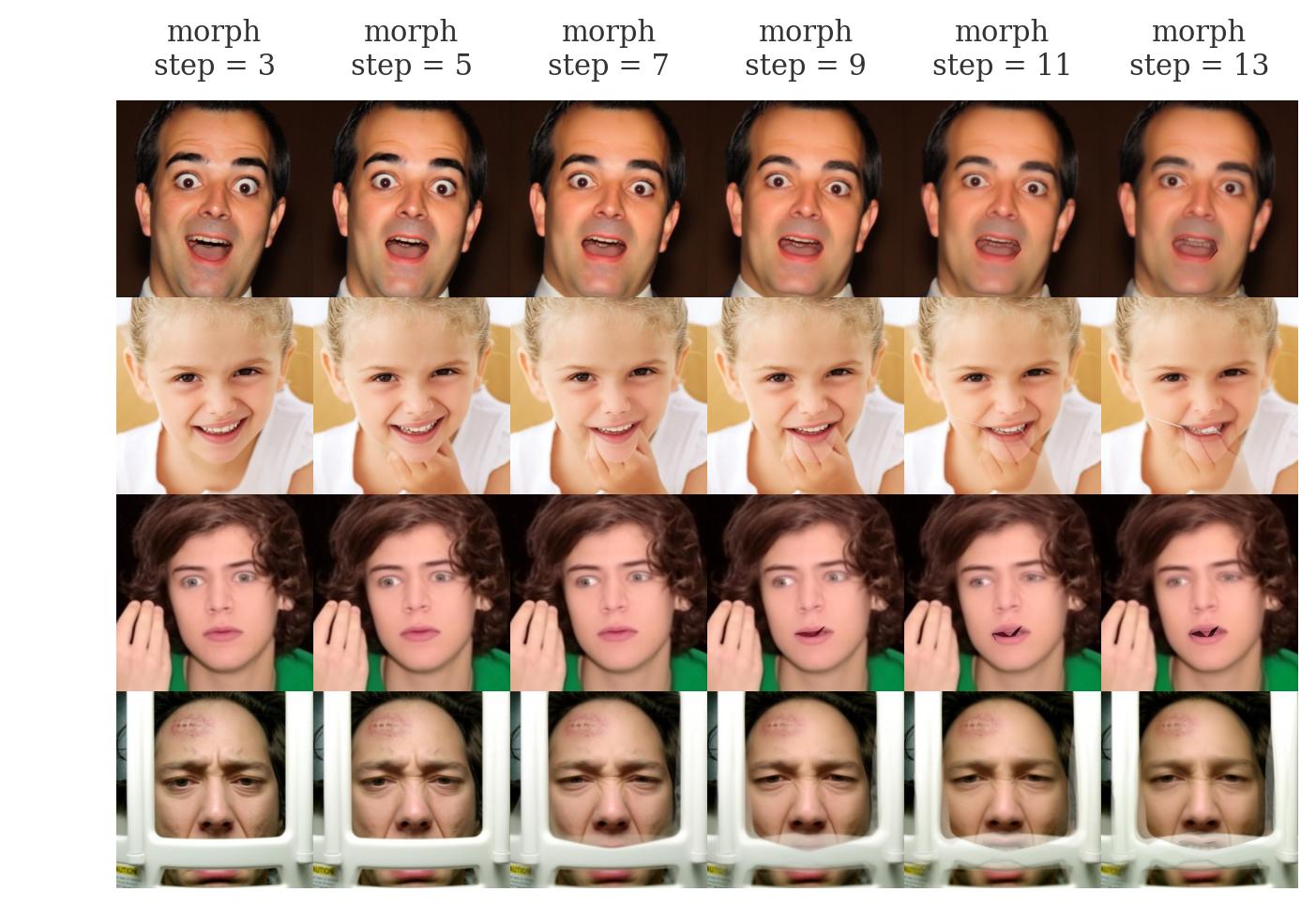}
\caption{Comparison of different morph step on AffectNet images. The morph step refers to the step at which we restart the diffusion process after face morphing is performed.}
  \label{fig:morph_step}
\end{figure}

We explore at which step is best to restart the diffusion process when using geometry-aware latent interpolation. In Figure \ref{fig:morph_step}, we edit faces from AffectNet, restarting the diffusion process at $t=\{3,5,7,9,11,13\}$ with a total of 15 inference steps. While the morphing step can introduce ghosting artifacts, the reverse diffusion process is capable of fixing this provided we give it enough steps. When $t=13$, ghosting artifacts are common. Working backwards, we see that by $t=5$ these artifact have been rectified. We therefore set the morph step to be a third of the way through the reverse diffusion process through out the paper.

\revision{\subsection{Compound Emotions}\label{sec:appendix_compound}}

\begin{figure}
  \centering
  \begin{subfigure}{0.3\linewidth}
  \centering
  \includegraphics[height=4.0cm]{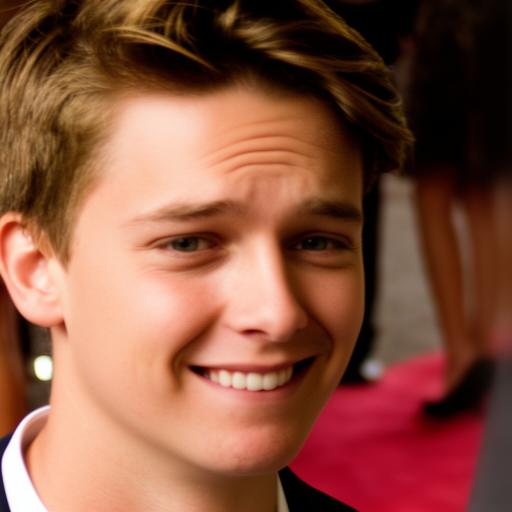}
    \caption{Happy-Sad.}
  \end{subfigure}
  \hfill
  \begin{subfigure}{0.3\linewidth}
  \centering
  \includegraphics[height=4.0cm]{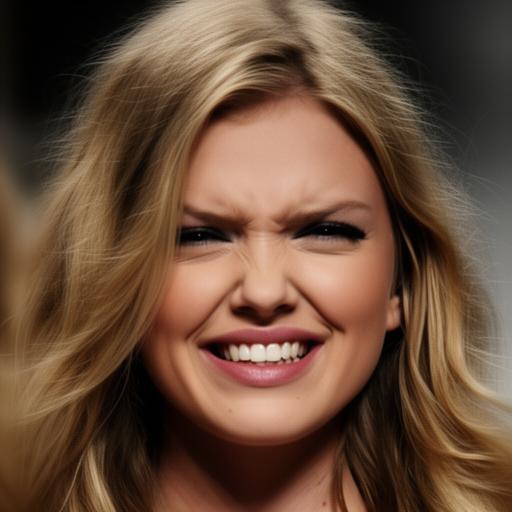}
    \caption{Happy-Disgust.}
  \end{subfigure}
  \hfill
  \begin{subfigure}{0.3\linewidth}
  \centering
  \includegraphics[height=4.0cm]{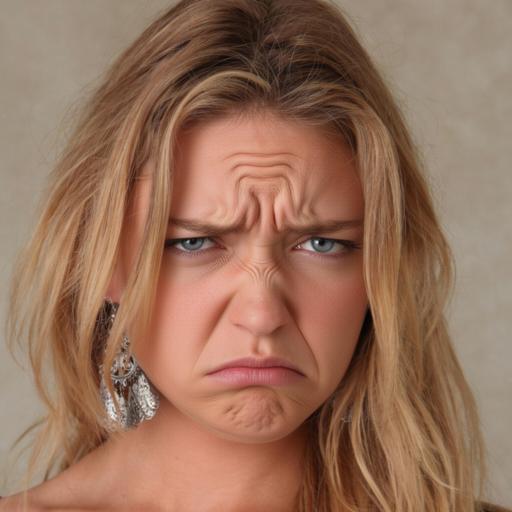}
    \caption{Anger-Sad.}
  \end{subfigure}
  \caption{\revision{Examples of compound emotions generated using two VA points with FLUX.2-dev used as a backbone.}}
  \label{fig:compound_emotions}
\end{figure}

\revision{As a straightforward extension, Barycentric Guidance can combine two distant VA targets by computing guidance for each and blending them with $\gamma \in [0, 1]$. This can allow for the generation of compound emotions that could not be easily represented using the VA model on its own. Mathematically, this naturally expands Equation \ref{eq:barycentric_editing} to $\gamma(w_1D_1 + w_2D_2 + w_3D_3) + (1-\gamma)(w_4D_4 + w_5D_5 + w_6D_6)$. Preliminary examples on happy–sad, sad–angry, and happy–disgust show mixed classifier activations consistent with the intended compound emotions. These examples use the exact VA location of the specific target emotions to derive the weights. CelebA images are used as the input images. Qualitative results for these compound emotions are shown in Figure~\ref{fig:compound_emotions}.}

\revision{\subsection{Controllability–Fidelity Trade-Off}\label{sec:appendix_trade_off}}

\revision{As discussed in the main body, stronger VA control introduces a controllability–fidelity trade-off. It is not only due to larger edit magnitude, although edit magnitude is an important factor. To better disentangle this, we conducted a matched edit-magnitude analysis on CelebA with the following steps: (1) Pair up EmoStyle edits against Our edits based on having the same or similar (< 0.01 distance) edit VA magnitudes. This essentially filters our images so that the distribution of VA edit strengths is the same for both subsets. (2) From these matched pairs, we further filter to strictly include only edits where the absolute VA magnitude is less than 0.2. (3) Recalculate Inception Score (IS) and ArcFace ID similarity for both subsets (yielding a subset of 465 images for each method; FID is omitted here as this sample size is too small for a stable calculation).}
 
\begin{table}
  \caption{\revision{Restricted edit strength magnitude experiment on EmoStyle and our approach. We compare 465 images with the same small edit strengths (less than 0.2 change in VA distance) the recalculate relevant metrics on this smaller subset. CelebA used for all source images.}}
  \label{tab:trade_off}
  \centering
  \small
  \begin{tabular}{lcc}
    \toprule
 Method & IS & ID Similarity \\ 
 \midrule
EmoStyle & 2.85 & 0.666 \\ 
Ours & 3.40 & 0.686 \\
\bottomrule
\end{tabular}
\end{table}

\revision{Result are shown in Table~\ref{tab:trade_off}. These results suggest that when edit magnitudes are restricted to comparable, low-intensity ranges, our approach outperforms EmoStyle in both image quality and identity preservation. This suggests that the reduced ID similarity observed in the unrestricted user study is a consequence of our method successfully reaching extreme, high-magnitude affective states that EmoStyle is incapable of generating, rather than an inherent flaw in our framework. }

\revision{Combining this with prior evidence from the main body of the paper allows us to be more precise when analysing the controllability-fidelity trade-off within our approach. Barycentric Guidance provides stronger human-perceived affective controllability, but this comes with a controllability–fidelity trade-off mainly driven by edit magnitude. It is also worth highlighting that our experiments on Qwen Image Edit obtain substantially improved fidelity (FID 10.67, IS 4.71, ID similarity 0.72, LPIPS 0.062) and the second best VA performance (RMSE 0.43, Entropy 0.81, VA std 0.55/0.49, $S^*_{2D}$ 0.28) compared to the FLUX.2 variant of our method. This suggest the underlying editing backbone also plays a significant role in this trade-off.}

\revision{\subsection{GPT-Image-2}\label{sec:gpt-image-2}}

\revision{As an additional point of comparison, we compare our approach with one of the leading closed source image generators (as of July 2026). Specifically, we evaluate GPT-Image-2's performance on continuous facial expression editing using AffectNet images and the same evaluation setup as in Table~\ref{tab:fee}. However, GPT-Image-2 accepts only text and image inputs. After initial prompt exploration, we found our best performing prompt to be ``Edit the provided face image to match target valence-arousal [<valence>, <arousal>]. Preserve the person's identity exactly: same face shape, eyes, nose, mouth structure, skin texture, age, hairstyle, facial hair, accessories, pose, camera angle, lighting, background, and image quality. Only change facial expression muscles enough to express the target affect. Valence is negative-to-positive emotion. Arousal is calm-to-activated energy. Return a photorealistic edited image of the same person. Valence and arousal range from -1.0 to 1.0.'' Here, <valence> and <arousal> are the target VA values.}

\begin{figure}[t]
  \centering
  \begin{subfigure}{0.485\linewidth}
  \centering
  \includegraphics[height=7.0cm]{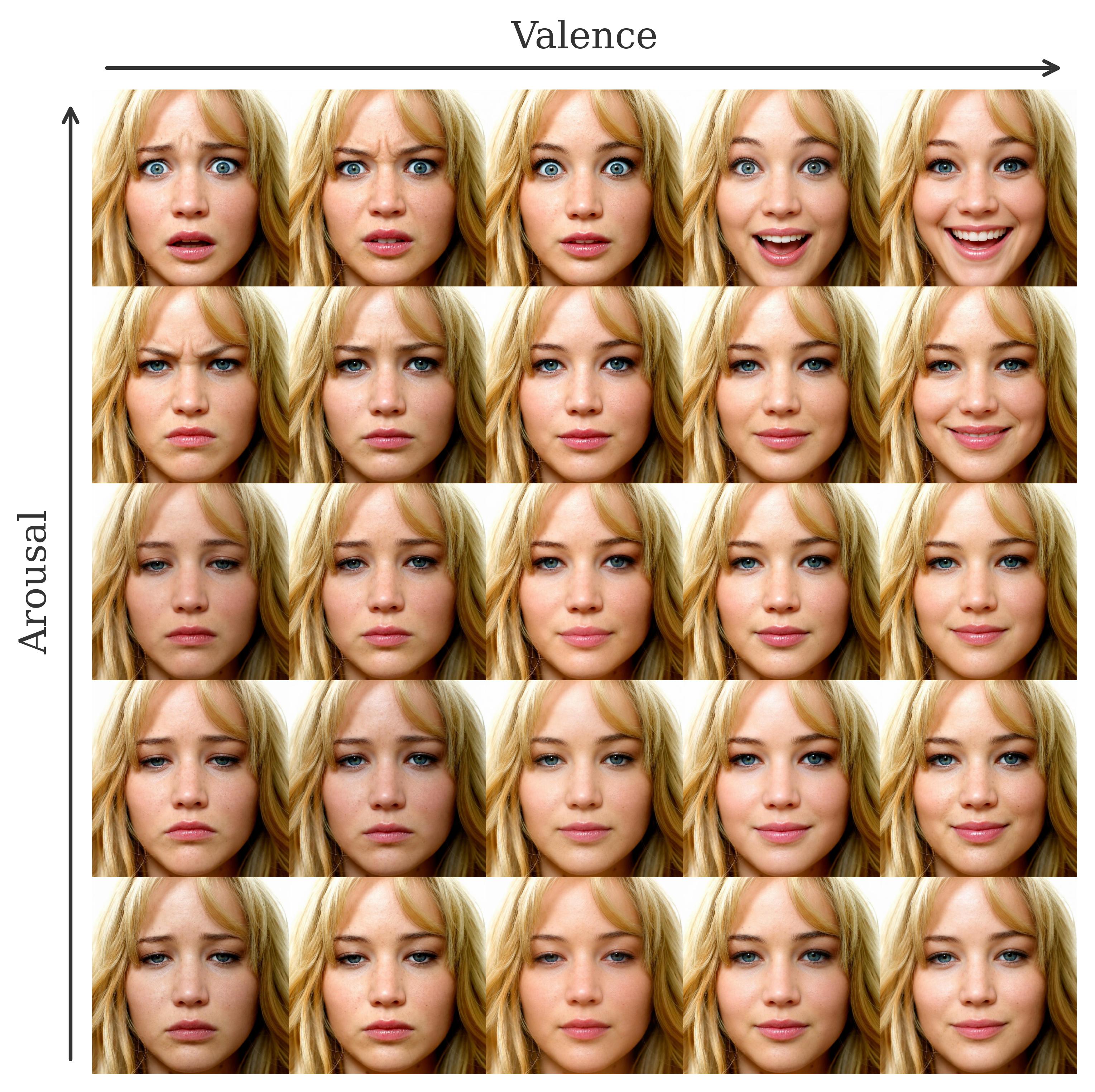}
    \caption{GPT-Image-2}
    \label{fig:gpt-comparison-a}
  \end{subfigure} \hfill
  \begin{subfigure}{0.485\linewidth}
  \centering
  \includegraphics[height=7.0cm]{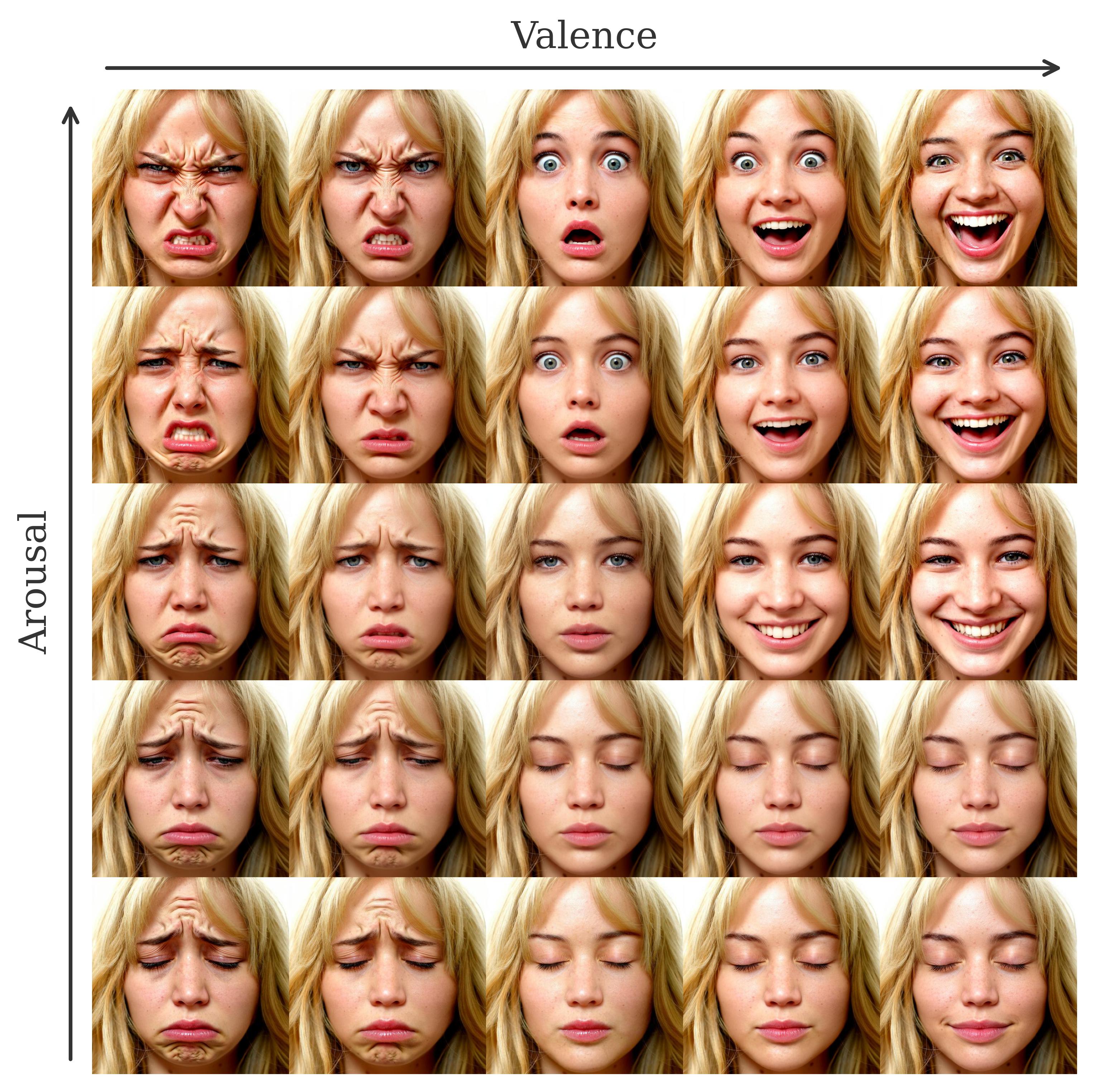}
    \caption{Ours (FLUX.2-dev)}
    \label{fig:gpt-comparison-b}
  \end{subfigure} 
   \\
  \caption{Qualitative comparison between our approach and GPT-Image-2 for AffectNet facial expression editing. VA range [-1,1] with uniform step size for rows and columns}
  \label{fig:gpt-comparison}
\end{figure}

\revision{Table~\ref{tab:fee_gpt} shows the performance of GPT-Image-2 compared with our approach, which uses FLUX.2-dev as the backbone. While GPT-Image-2 performs extremely well on image quality and identity preservation metrics, it lags behind on all VA controllability metrics, with our approach outperforming GPT-Image-2 in VA accuracy and diversity, as well as VA smoothness. A plausible explanation is that text provides a poor interface for numerical and continuous manipulation of emotions, whereas our latent-space-based approach allows for gradual and controlled changes in emotions. In Figure~\ref{fig:gpt-comparison} we present a visual comparison between our approach (with FLUX.2-dev as our backbone) and GPT-Image-2 on AffectNet facial expression editing. In line with the results in Table~\ref{tab:fee_gpt}, we see a noticeable increase in diversity while also better capturing extreme emotions.}

\begin{table}
  \caption{\revision{Facial Expression Editing comparison between GPT-Image-2 and our approach.}}
  \label{tab:fee_gpt}
  \centering
  \scriptsize
  \begin{tabular}{lcccc|cccc}
    \toprule
    Model & \multicolumn{4}{c|}{\textbf{VA Controllability}} & \multicolumn{4}{c}{Image Quality/ Preservation}  \\

\cmidrule(lr){2-5} \cmidrule(lr){6-9}

 & {RMSE}$\downarrow$ & {VA Entropy}$\uparrow$ & {VA Std}$\uparrow$ & {$S^{*}_{2D}$}$\downarrow$ & FID$\downarrow$ & IS$\uparrow$ & ID Sim.$\uparrow$ & LPIPS$\downarrow$
 \\
 \midrule
    \revision{GPT-Image-2} & \revision{0.530} & \revision{0.327} & \revision{0.52 / 0.35} & \revision{0.402} & \revision{\bf 9.38} & \revision{\bf 4.71} & \revision{\bf 0.84} & \revision{\bf 0.066} \\
    Ours & \bf 0.433 & \bf 0.916 & \bf 0.64 / 0.57 & \bf 0.303 & 11.02 & 4.42 & 0.55 & 0.096 \\
    \bottomrule
  \end{tabular}
\end{table}

\revision{\subsection{Relative Improvements}\label{sec:appendix_relative}}

\revision{Here we provide additional information regarding the relative percentage improvements presented in the abstract. Table~\ref{tab:relative_improvements} provides the calculation breakdown. Each percentage is computed against the strongest baseline for the corresponding dataset and metric.}

\begin{table}
  \caption{\revision{Relative improvements over the strongest baseline. Numbers are derived from the results in Table~\ref{tab:fee}.}}
  \label{tab:relative_improvements}
  \centering
  \small
  \begin{tabular}{lcccc}
    \toprule
    Dataset & Entropy $\uparrow$ & VA Std (V) $\uparrow$ & VA Std (A) $\uparrow$ & RMSE $\downarrow$ \\
    \midrule
    AffectNet & +48.7\% & +36.2\% & +50.0\% & +29.0\% \\
    FFHQ & +29.5\% & +28.0\% & +41.7\% & +26.4\% \\
    CelebA-HQ & +34.7\% & +35.4\% & +45.9\% & +25.1\% \\
    \bottomrule
  \end{tabular}
\end{table}

\revision{\subsection{Derivations}
\label{sec:derivations}}

\revision{In this section, we include the following derivation/proof sketch for Equations~\ref{eq:barycentric_editing} and \ref{eq:barycentric_generation}. These derivations are based on observations from \citet{song2020score} and \citet{ho2022classifier} that diffusion models approximate the score function and that Bayes can be applied to conditional generation, respectively. Equation \ref{eq:barycentric_editing} is given as: }

\revision{\begin{equation}
    D^{edit}_{bary}(x_t)= w_1D(x_t,y_1)+ w_2 D(x_t,y_2) + w_3 D(x_t,y_3) 
\end{equation}}

\revision{But can be rewritten as a summation like so: }

\revision{\begin{equation}
=\sum_{i=1}^3 w_i D(x_t,y_i)
\end{equation}}

\revision{\citet{song2020score} finds that diffusion models learn approximate the score function. For conditional diffusion we therefore have the approximation $D_{\theta}(x_t,y_i)\approx \nabla_{x_t}\log p_t(x_t|y_i)$, which when plugged into the sum gives us: }

\revision{\begin{equation}
\approx\sum_{i=1}^3 w_i \nabla_{x_t}\log p_t(x_t|y_i) 
\end{equation}}

\revision{Through Bayes rule (the denominator in Bayes rule can be ignored because we are taking the gradient with respect to $x_t$), we can rewrite this as: }

\revision{\begin{equation}
=\sum_{i=1}^3 w_i (\nabla_{x_t}\log p_t(y_i|x_t)+\nabla_{x_t}\log p_t(x_t))
\end{equation}}

\revision{However, since all weights sum to one the second score function can be taken outside of the summation to get the final expression:}

\revision{\begin{equation}
=\nabla_{x_t} \log p_t(x_t)  + \sum_{i=1}^3 w_i \nabla_{x_t} \log p_t(y_i \mid x_t)
\end{equation}}

\revision{For Equation~\ref{eq:barycentric_generation} a similar derivation can be done using the same logic. Specifically, using the score function approximation from \citet{song2020score} we get: }

\revision{\begin{equation}
D^{gen}_{bary}(x_t) = (w_{cfg}-s)D_{\theta}(x_t,y_{base}) + (1-w_{cfg})D_{\theta}(x_t) + s\left(\sum_{i=1}^3w_iD_{\theta}(x_t,y_i)\right)
\end{equation}}

\revision{\begin{equation}
\approx (w_{cfg}-s)\nabla_{x_t}\log p_t(x_t|y_{base}) + (1-w_{cfg})\nabla_{x_t}\log p_t(x_t) + s\left(\sum_{i=1}^3w_i\nabla_{x_t}\log p_t(x_t|y_{i})\right)
\end{equation}}

\revision{We can apply Bayes rule like with the prior derivation (i.e. $\nabla_{x_t}\log p_t(x_t|y)=\nabla_{x_t}\log p_t(y|x_t)+\nabla_{x_t}\log p_t(x_t)$) then group terms we get the final score function based perspective: }

\revision{\begin{equation}
\nabla_{x_t} \log p_t(x_t) + (w_{cfg}-s) \nabla_{x_t} \log p_t(y_{base} \mid x_t) + s\sum_{i=1}^3 w_i \nabla_{x_t} \log p_t(y_i \mid x_t)
\end{equation}}

\subsection{Experimental Details}
\label{sec:exp_details}

\begin{figure}
  \centering
  \begin{subfigure}{0.3\linewidth}
  \centering
  \includegraphics[height=4.5cm]{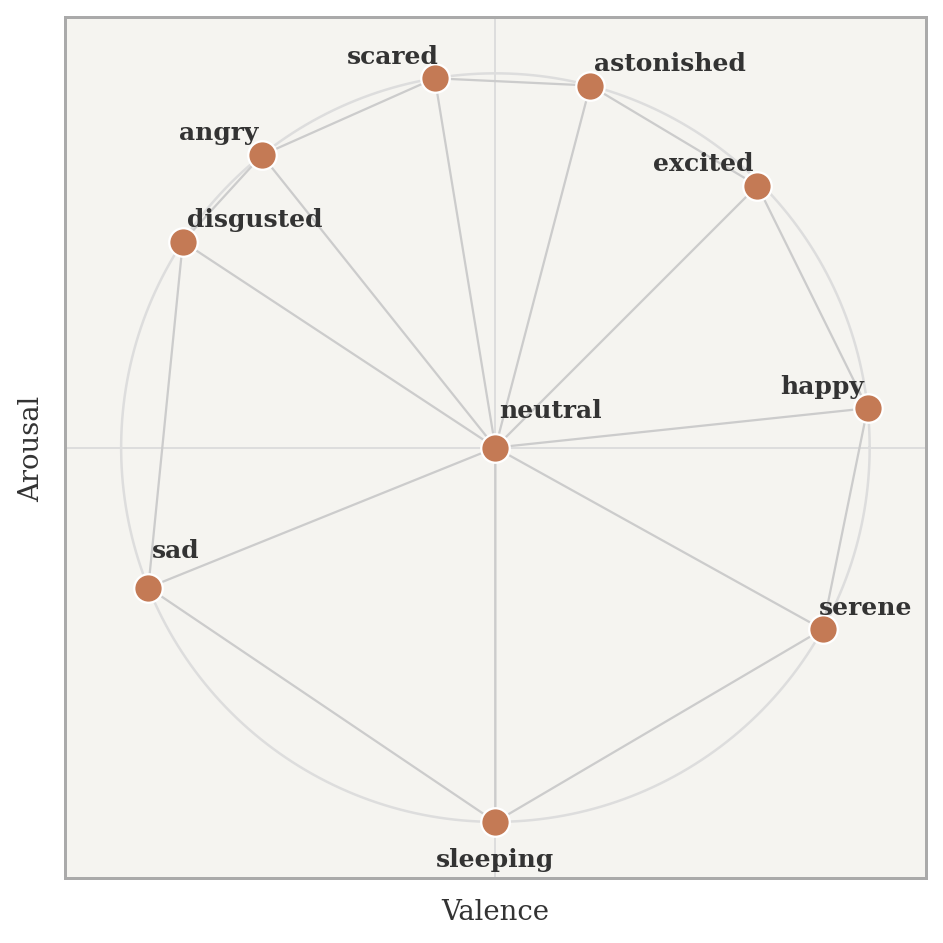}
    \caption{Mesh used for faces.}
    \label{fig:mesh-a}
  \end{subfigure}
  \hfill
  \begin{subfigure}{0.3\linewidth}
  \centering
  \includegraphics[height=4.5cm]{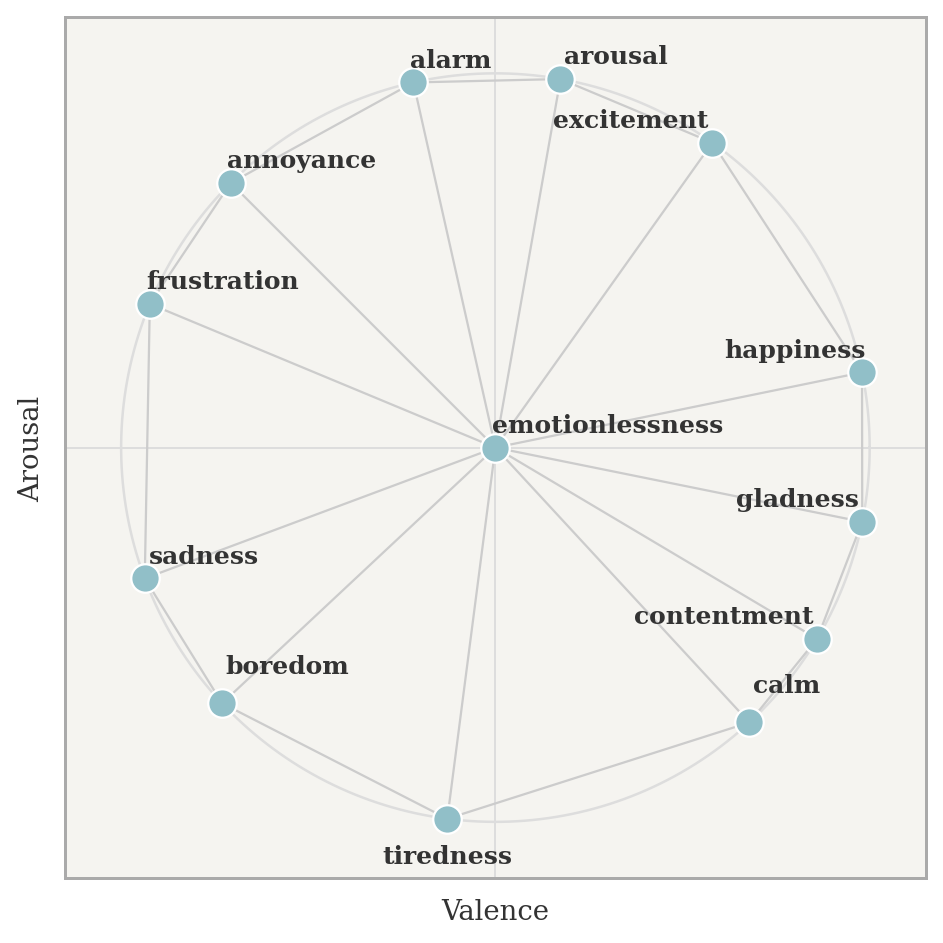}
    \caption{Mesh used for artworks.}
    \label{fig:mesh-b}
  \end{subfigure}
  \hfill
  \begin{subfigure}{0.3\linewidth}
  \centering
  \includegraphics[height=4.5cm]{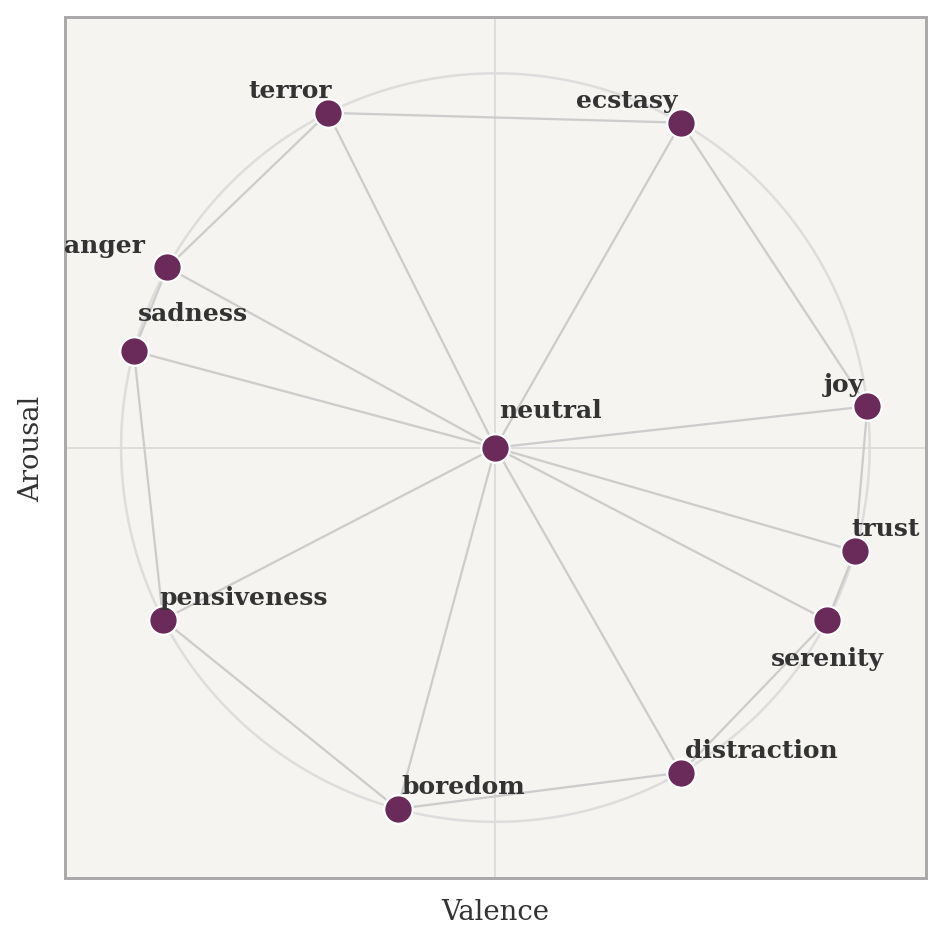}
    \caption{Mesh used for scenes.}
    \label{fig:mesh-c}
  \end{subfigure}
  \caption{Triangle meshes used for different domains. Nodes and their locations are derived from datasets for the specific domain, in this case, AffectNet, EmoArt, and FindingEmo.}
  \label{fig:meshes}
\end{figure}

We use a guidance scale of $w_{cfg}=4.0$ across all modalities, and for FLUX.2-dev we use the default value for distilled guidance of 4.0. 

\revision{\paragraph{Mesh Construction} In early experiments, we explored using a denser mesh with over 100 emotion words. However, we found that ambiguous anchors (e.g., ``bitter'') and rare words (e.g., ``droopy'') led to severe misalignments between the intended target and the generated expression. If we compare the predictive error (RMSE when performing facial expression editing on AffectNet, testing 100 edits per emotion) of basic emotions against rarer, denser mesh words, the drop in accuracy (shown in Table~\ref{tab:mesh_experiment}) is stark. }

\begin{table}
  \caption{\revision{The VA accuracy of different emotion words when performing facial expression editing on AffectNet with FLUX.2-dev. Overall, 100 different emotion words were considered. Neither Barycentric Guidance face morphing were used here, just our standard prompt template, where only the emotion word is altered.}}
  \label{tab:mesh_experiment}
  \centering
  \small
  \begin{tabular}{lc}
\toprule
& RMSE \\
\midrule
Happy & 0.054 \\
Sad & 0.149  \\
Angry & 0.306  \\ 
Bitter & 0.451  \\
Droopy & 0.825   \\
At Ease & 0.908  \\
\midrule
Basic Emotions Mean & 0.216 \\  
Other Emotions Mean & 0.501  \\
\bottomrule
\end{tabular}
\end{table}

\revision{The sensitivity to the mesh therefore primarily depends on the text encoder used and how well each discrete anchor will be rendered. This informed our choice of a compact mesh using common, visually grounded, and semantically unambiguous emotion anchors. A denser mesh is not automatically better; it helps only when the added anchors are reliably expressible by the editor. }

\begin{figure}
  \centering
  \begin{subfigure}{0.49\linewidth}
  \centering
  \includegraphics[width=\linewidth]{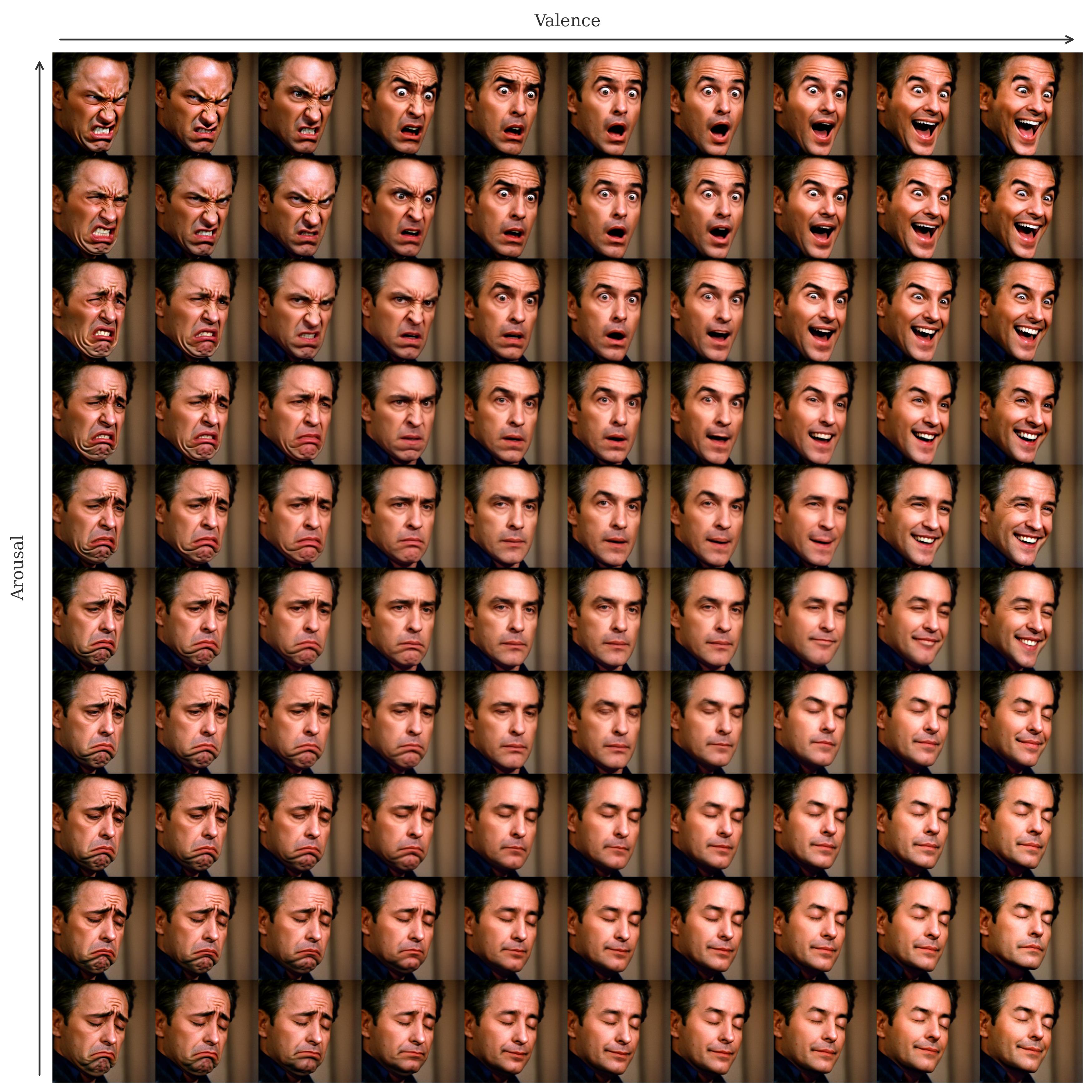}
    \caption{Original Mesh.}
    \label{fig:mesh_augment_a}
  \end{subfigure}
  \begin{subfigure}{0.49\linewidth}
  \centering
  \includegraphics[width=\linewidth]{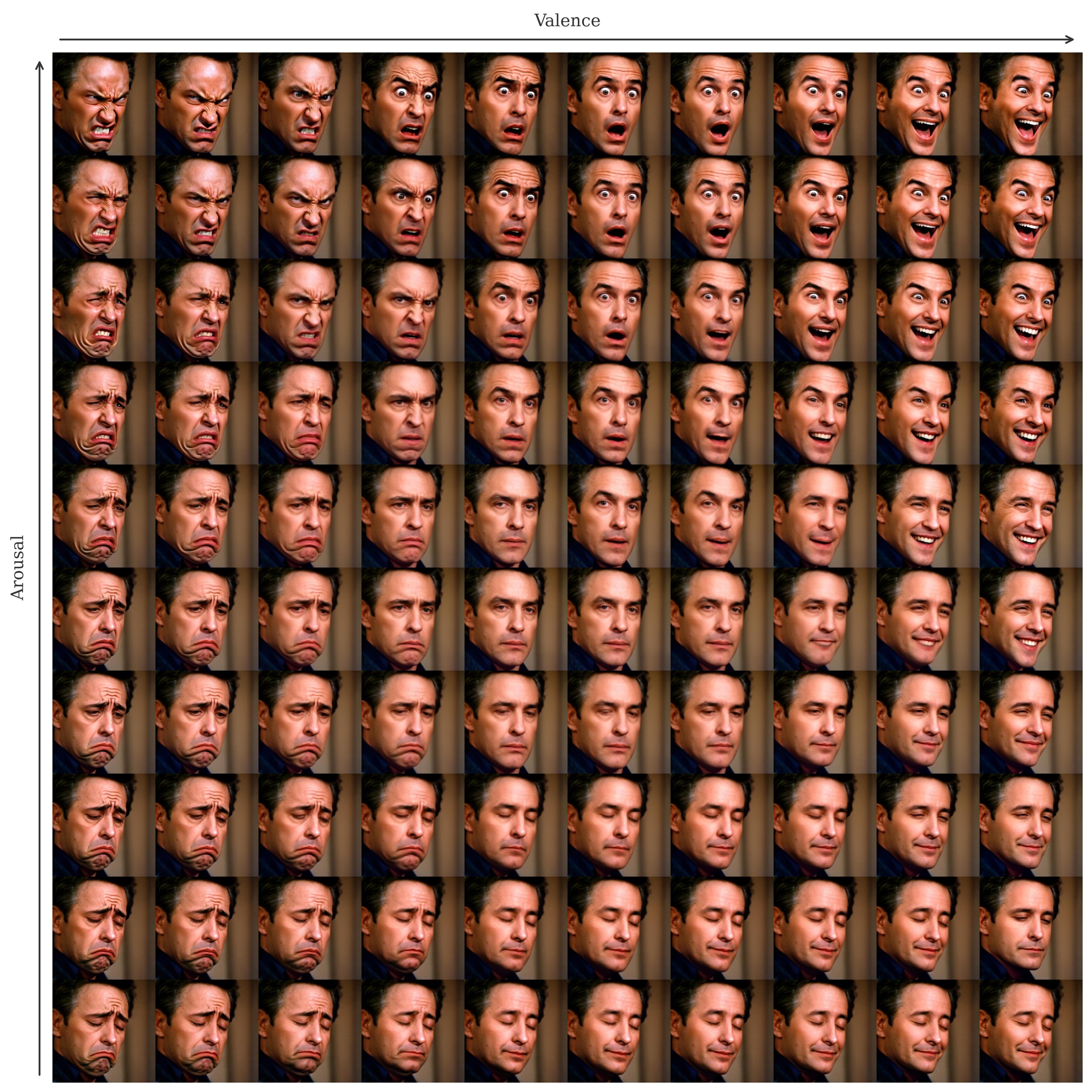}
    \caption{Additional Lower Right Anchor.}
    \label{fig:mesh_augment_b}
  \end{subfigure}
  
  \caption{\revision{10 by 10 grids of our approach (backbone FLUX.2-dev) being used to edit an image from AffectNet. Our standard mesh used for all other facial expression editing results is shown in (a), whereas (b) shows editing results on the same mesh except with an additional anchor, ``relaxed'', in the bottom right corner.}}
  \label{fig:mesh_augment}
\end{figure}

\revision{We also examined VA placement. Adding a new anchor ``relaxed'' in the lower-right region of the VA space. As shown in Figure~\ref{fig:mesh_augment}, this improves the gradual low-arousal transitions, producing more natural sleepy/relaxed expressions, such as progressive eye closure, compared with the current sparser lower-arousal design. This suggest that results may be improved by further refinement of the mesh, which we leave as a promising direction for future work.}

\revision{Triangle meshes used for editing and generation throughout the paper are shown Figure \ref{fig:meshes}.}

\paragraph{Compute Costs} We run our core experiments on NVIDIA H200 GPUs, which takes on average 0.64 seconds per inference step when using FLUX.2-dev. For normal generation, the number of function evaluations (NFEs) corresponds to the number of steps (i.e. $30$ steps = $30$ NFEs). For our full method, in the experiments we have at the moment, we use $30$ steps to generate each anchor, then we restart the diffusion process at step $10$ out of $30$. Barycentric guidance and CFG for the restarted process uses 4 function evaluations per step. In total therefore, our method uses:  $30\times3+20\times4=180$ NFEs. If no morphing is used we would have $30\times4=120$ NFEs. Therefore, generation with prompt interpolation takes roughly 19.2 seconds per image, whereas our full approach takes around 76.8 seconds without morphing and 115.2 seconds with morphing.

\paragraph{Facial Expression Editing} For GANmut, we trained on AffectNet using the default configuration provided in the source repo. We train on uncropped rather than cropped faces. This allows us to compare background preservation between methods. GANmut utilizes polar instead of cartesian coordinates; for diversity metrics, we therefore sampled ten evenly spaced angles across ten different radii. For EmoStyle we use the default configuration in the original repo and run both phase 1 and phase 2 of the generation process. The prompt template used for our method is ``Change this persons facial expression to be <emotion> while maintaining the same facial features, composition, lighting, and camera angle''. We use 30 inference steps for our approach for both the anchors (discrete emotions) and the main image generation. The LPIPS metric used to measure the non-identity, non-affect content preservation is calculated by first using MediaPipe \citep{lugaresi2019mediapipe} to predict face landmarks. These landmarks are then used to create a binary mask, blocking out the face region only. This is done both the source and edited image. Finally, we calculate LPIPS between the face masked source image and face masked edited image.

\paragraph{Continuous Editing Comparison} The classifier used for measuring smoothness for our 1D editing comparison was trained on AffectNet. Specifically, we use a VGG model \citep{simonyan2014very} finetuned on AffectNet data for prediction of the 7 basic emotions, undersampling the overrepresented emotions in order to balance the dataset. All approach use the default parameters provided in the official implementations. 14 inference steps are used for all approaches.

\paragraph{EmoArt} For the EmoArt dataset we finetune a DINOv3 ConvNeXt-Tiny \citep{simeoni2025dinov3} model to predict the emotion and the art style of the image, ultimately achieving 70.8\% accuracy. The classifier predicts dicrete emotion, however, the EmoArt dataset is designed to represent emotion spread evenly over the VA model. We therefore recovered the VA value via a probability-weighted sum over the EmoArt VA points for each emotion. We use the prompt template ``Change this artwork to evoke <emotion>, maintaining the art style as <style>, keep the content of the art the same'', where ``<style>'' refers to the art style of the original image. Our approach uses 14 inference steps.

\paragraph{FindingEmo} For FindingEmo, we follow \citep{dang2025emoticrafter} and train a linear probe on top of a frozen CLIP model (vit\_base\_patch32\_clip\_224) to predict the valence and arousal of an image. FindingEmo contains valence values in the range of $[-3.0, 3.0]$ and arousal values in the range of $[0.0, 6.0]$. We map both dimensions to a range of $[-1.0, 1.0]$ to allow for better comparison with other modalities. For this dataset we use the prompt template ``Change the people and environment to feel more <emotion>, it should remain as a realistic photograph''. Base prompts are taken from EmotiCrafter's official code base \citep{dang2025emoticrafter}. Our approach uses 14 inference steps.

\section{Text Encoders}
\label{sec:appendix_text_encoders}

\begin{figure}
  \centering
  \begin{subfigure}{0.3\linewidth}
  \centering
  \includegraphics[height=4.5cm]{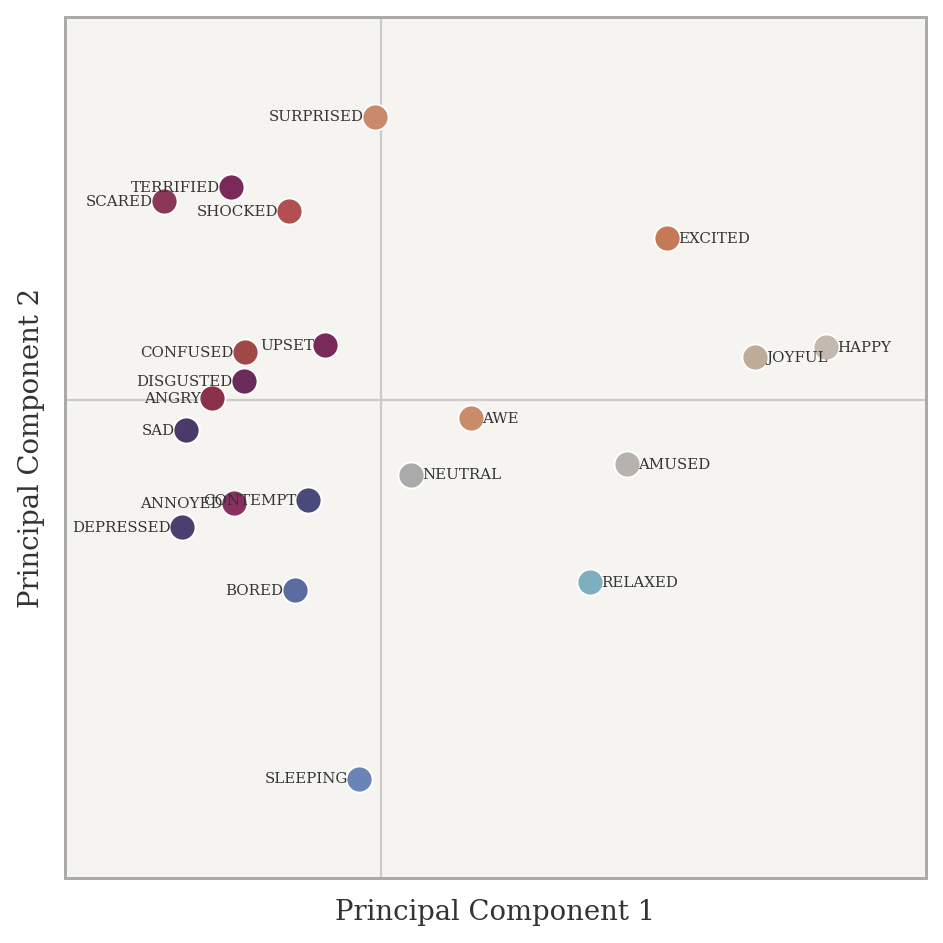}
    \caption{{\bf Prompt Template:} "a picture of a <emotion> man".}
    \label{fig:text-pca-a}
  \end{subfigure}
  \hfill
  \begin{subfigure}{0.3\linewidth}
  \centering
  \includegraphics[height=4.5cm]{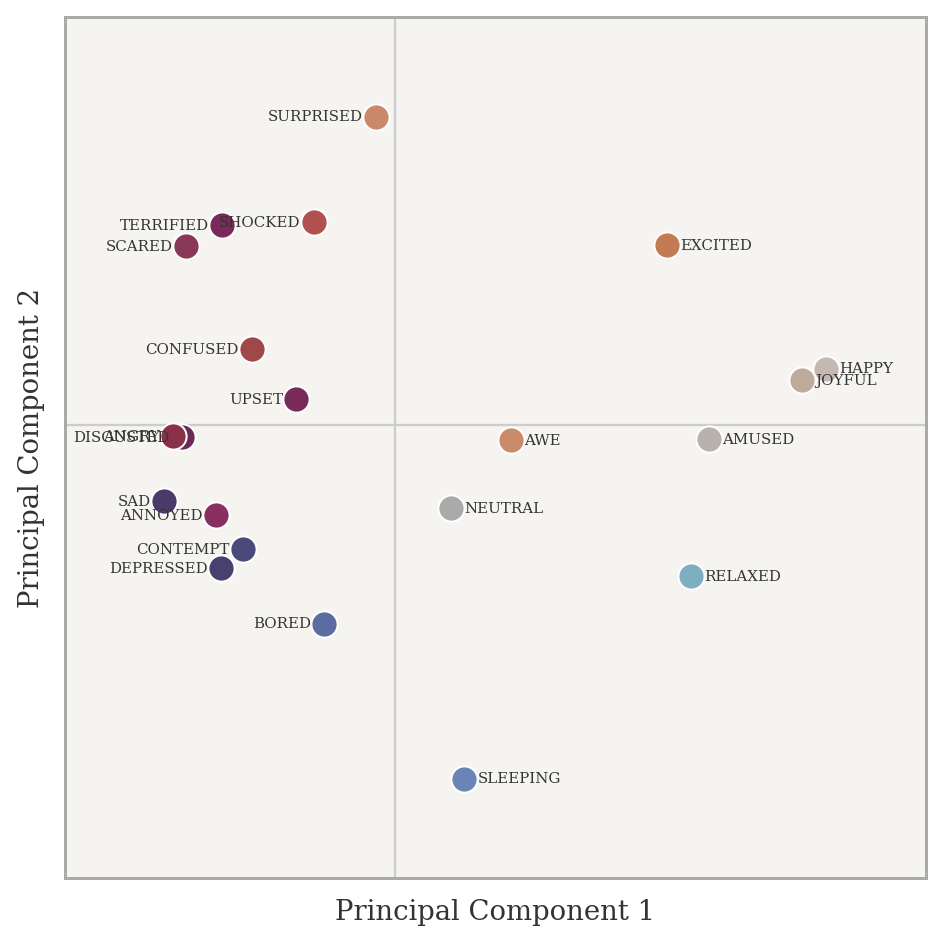}
    \caption{{\bf Prompt Template:} "an image of a <emotion> woman".}
    \label{fig:text-pca-b}
  \end{subfigure}
  \hfill
  \begin{subfigure}{0.3\linewidth}
  \centering
  \includegraphics[height=4.5cm]{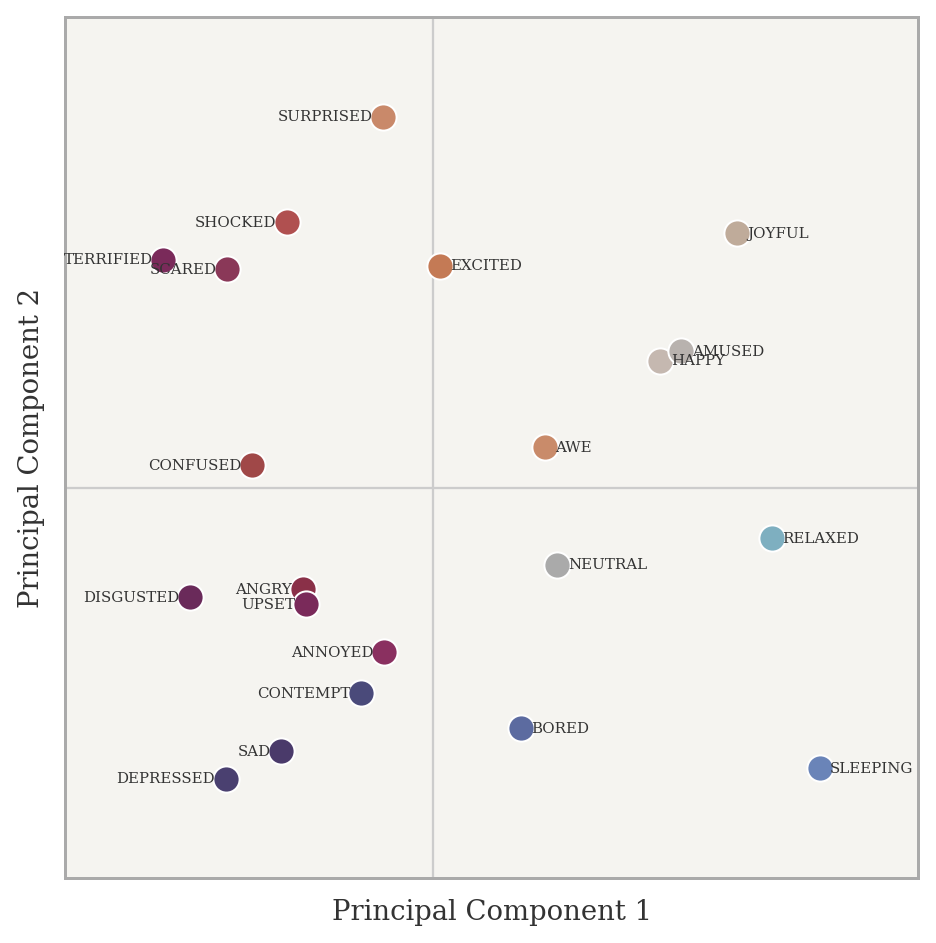}
    \caption{{\bf Prompt Template:} "<emotion>".}
    \label{fig:text-pca-c}
  \end{subfigure}
  \caption{PCA performed on different prompt templates on the CLIP ViT-L/14 model.}
  \label{fig:pca_prompt_templates}
\end{figure}

\begin{figure}
  \centering
  \begin{subfigure}{0.3\linewidth}
  \centering
  \includegraphics[height=4.5cm]{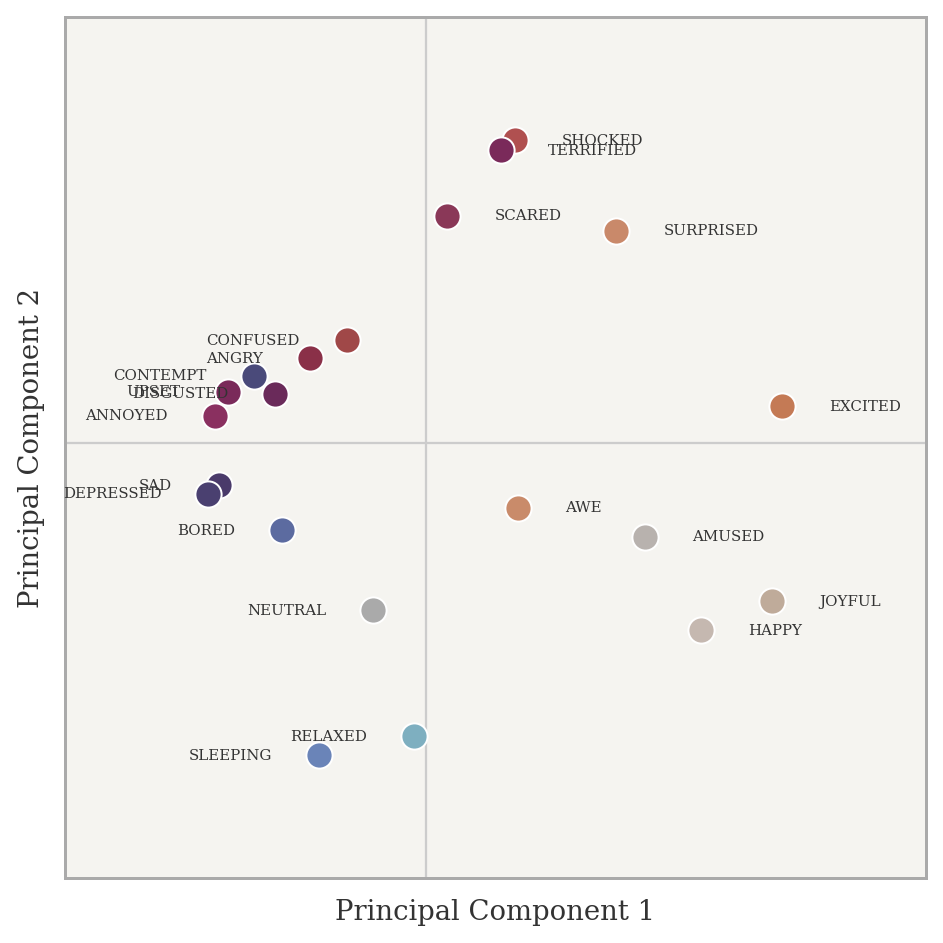}
    \caption{OpenCLIP ViT-bigG/14.}
    \label{fig:text-pca-d}
  \end{subfigure}
  \hfill
  \begin{subfigure}{0.3\linewidth}
  \centering
  \includegraphics[height=4.5cm]{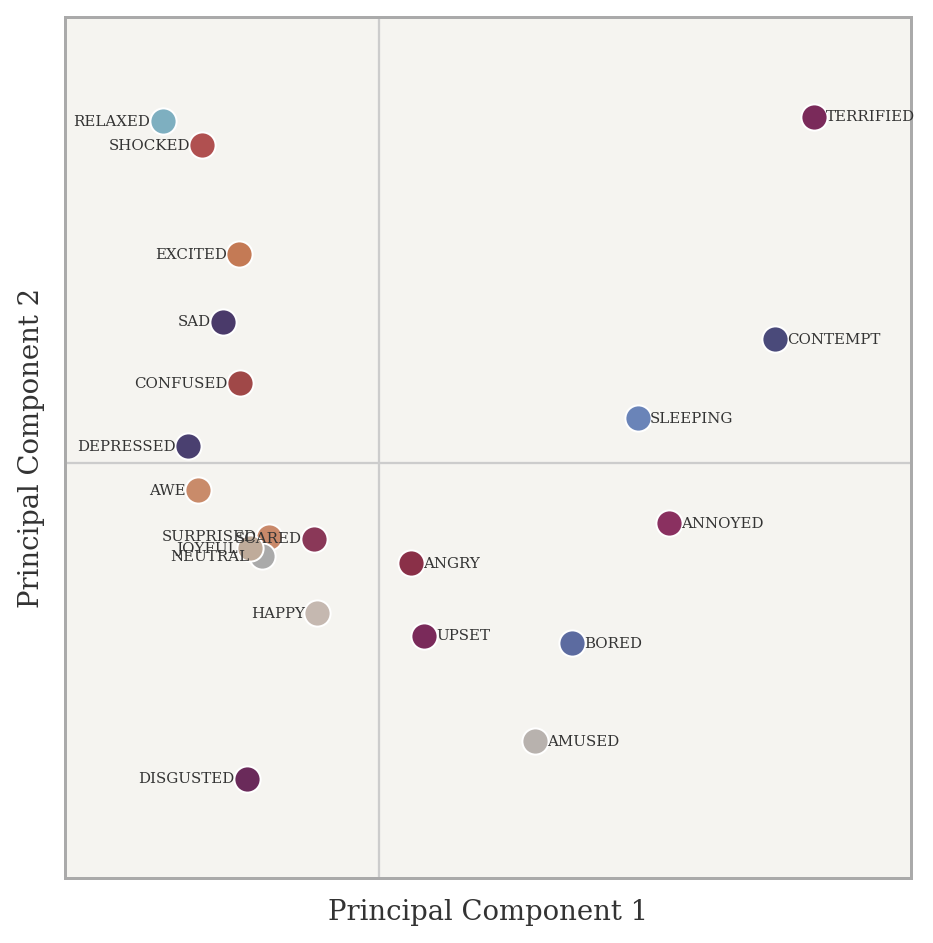}
    \caption{T5 v1.1 XXL.}
    \label{fig:text-pca-e}
  \end{subfigure}
  \hfill
  \begin{subfigure}{0.3\linewidth}
  \centering
  \includegraphics[height=4.5cm]{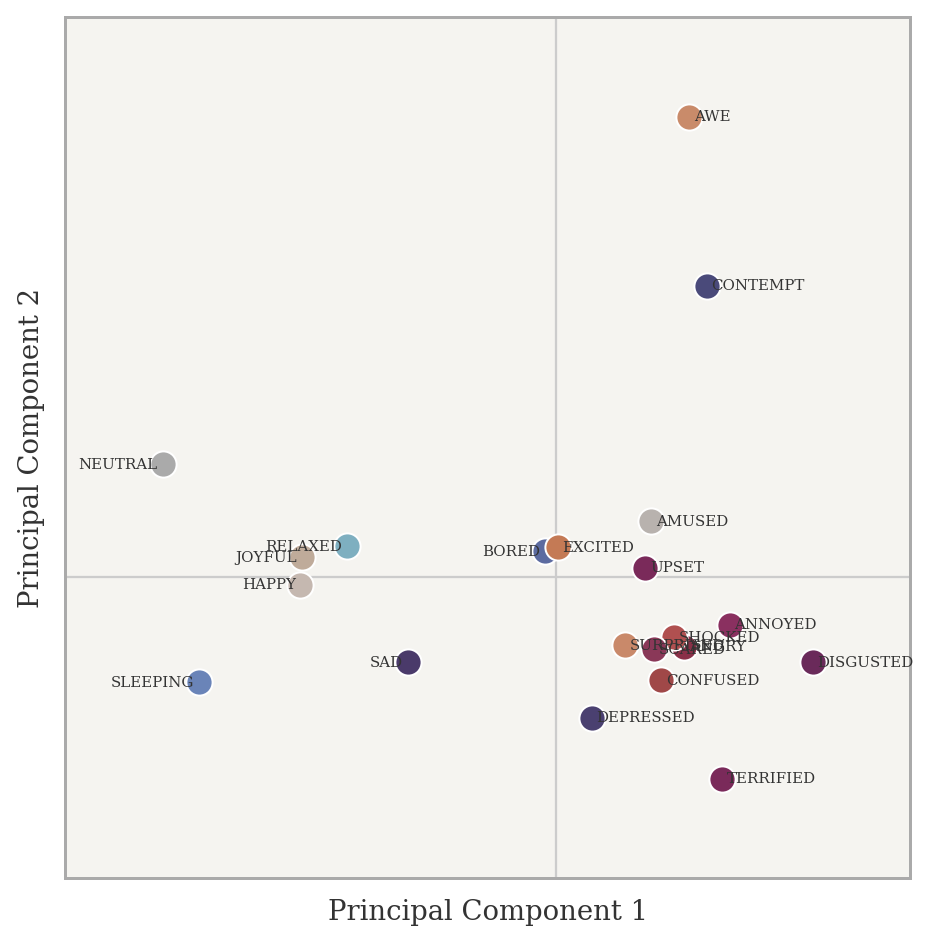}
    \caption{Qwen2.5-VL.}
    \label{fig:text-pca-f}
  \end{subfigure}
  \caption{PCA performed text embeddings from different text encoders. Prompt template used was ``a portrait photo of a <emotion> face''.}
  \label{fig:pca_models}
\end{figure}

In this section, we further explore the results of performing PCA on different text encoder's embeddings. In the main body, we provide an example of performing PCA on CLIP embeddings. We used the CLIP ViT-L/14 model with the prompt template ``a portrait photo of a <emotion> face'', where ``<emotion>'' is replaced with an emotion word. This process recovers a good approximation of the VA model on the principal components. Sometimes principal components need to be rotated or mirrored, but importantly, the ordering of points reliably follows the ordering of points in the VA model. To show that this is not dependent on the prompt template, we replicate this result in Figure \ref{fig:pca_prompt_templates} using three different prompts. All prompt templates considered recover a good approximation of the VA model on their principal components. 

In Figure \ref{fig:pca_models}, we repeat this experiment on different text encoders with the prompt template ``a portrait photo of a <emotion> face''. For OpenCLIP ViT-bigG/14, as shown in Figure \ref{fig:text-pca-d}, we once again recover the VA model. However, for T5 v1.1 XXL and Qwen2\_5\_VLForConditionalGeneration (shown in Figures \ref{fig:text-pca-e} \& \ref{fig:text-pca-f}) the process largely fails, with opposite emotions often appearing side by side. 

While tradition text-to-image models like Stable Diffusion uses CLIP encoders to condition on text, more modern image editing models do not. For example, FLUX.1-Kontext uses T5 as its primary text encoder. FLUX.2 models use QwenVL and MistralVL text encoders, neither of which provide VA-like subspaces. Finally, Qwen Image Edit also used QwenVL as a text encoder, thus motivating the guidance based approach which we take in this paper.

\section{Additional User Study Details}
\label{sec:additional_user_study}

Our user study was  conducted uses the Qualtrics platform which allows for anonymised survey's to be conducted. Consent and participant information is provided at the start of the online survey.  After this, a user must tick a consent tickbox to continue to the survey.  If they tick the ``I do not consent'' tickbox, they will be taken to the end of the survey. Following this there is a short tutorial explaining the survey and the valence-arousal model to users. It includes the following text alongside accompanying visuals:
\begin{enumerate}
    \item ``Emotions can be described by two dimensions: Valence (x-axis, unpleasant → pleasant) and Arousal (y-axis, calm → activated). Every emotion maps to a unique point within the unit circle. Emotions are also continuous under the Valence–Arousal Model, for example in between the astonished and happy, we have excited.''
    \item ``Each point on the VA circle corresponds to a specific emotion. Valence (x-axis) separates positive emotions on the right from negative ones on the left. Here the happy face sits exactly on the Happy point, and the sad face exactly on Sad.''
    \item ``Arousal (y-axis) runs from calm at the bottom to activated at the top, the sleepy face sits exactly on Sleepy. Emotions are also continuous: the fearful expression sits between Astonished and Angry, illustrating that every point in the circle is a distinct emotional blend.''
    \item ``Intensity corresponds to distance from the centre. All three faces point toward disgust, but move progressively further out; weak, medium, then strong.''
    \item `` Each question shows a source face (blue border) alongside a target emotion marked as a orange cross on the valence-arousal diagram. The blue dot shows the source face's emotion. Below the divider are two edits (orange border), each produced by a different image-editing method from the same source. You will be asked which edits looks better to you. This is a subjective measure, so if you are not sure, you can also select that they are equal.''
\end{enumerate}

\begin{figure}
  \centering
  \begin{subfigure}{0.49\linewidth}
  \centering
  \includegraphics[height=9.0cm]{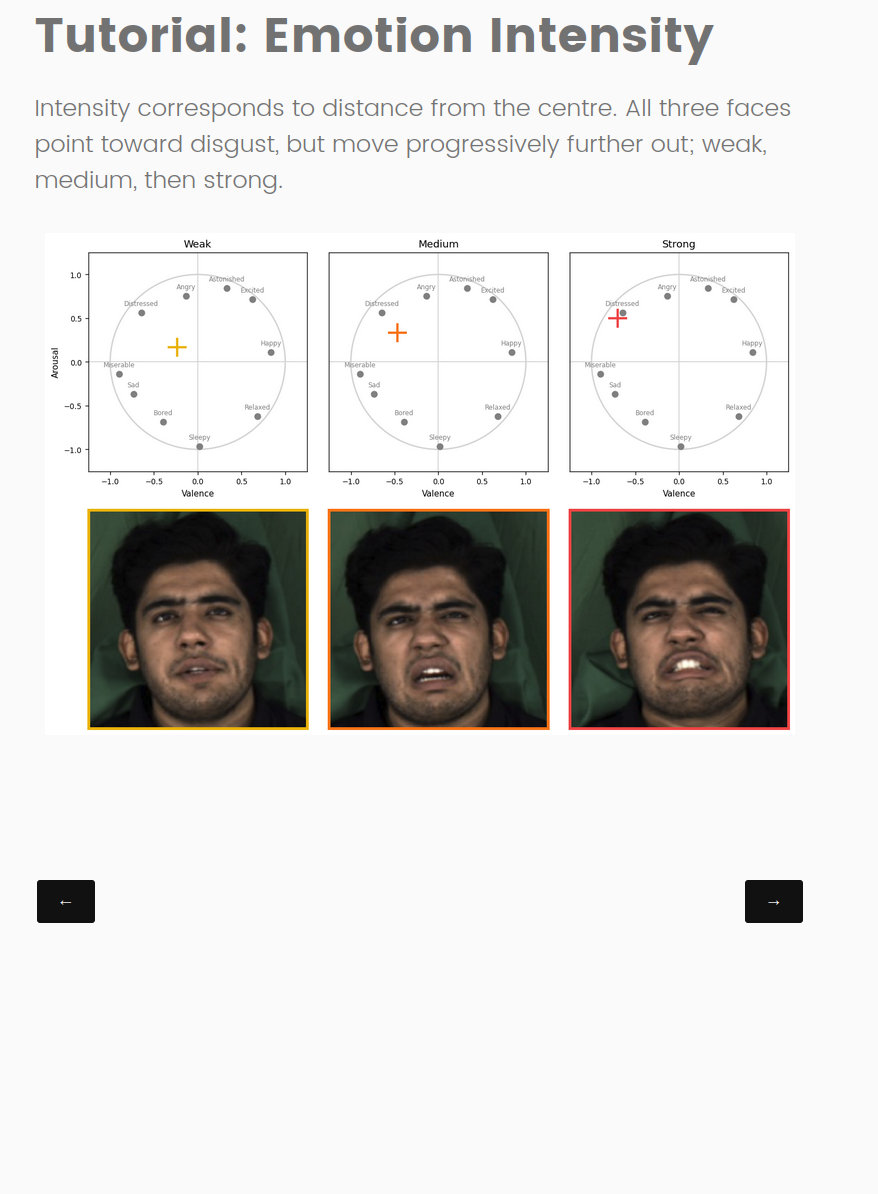}
    \caption{Screenshot from our a user studies tutorial.}
    \label{fig:user_study_screenshots-a}
  \end{subfigure}
  \begin{subfigure}{0.49\linewidth}
  \centering
  \includegraphics[height=9.0cm]{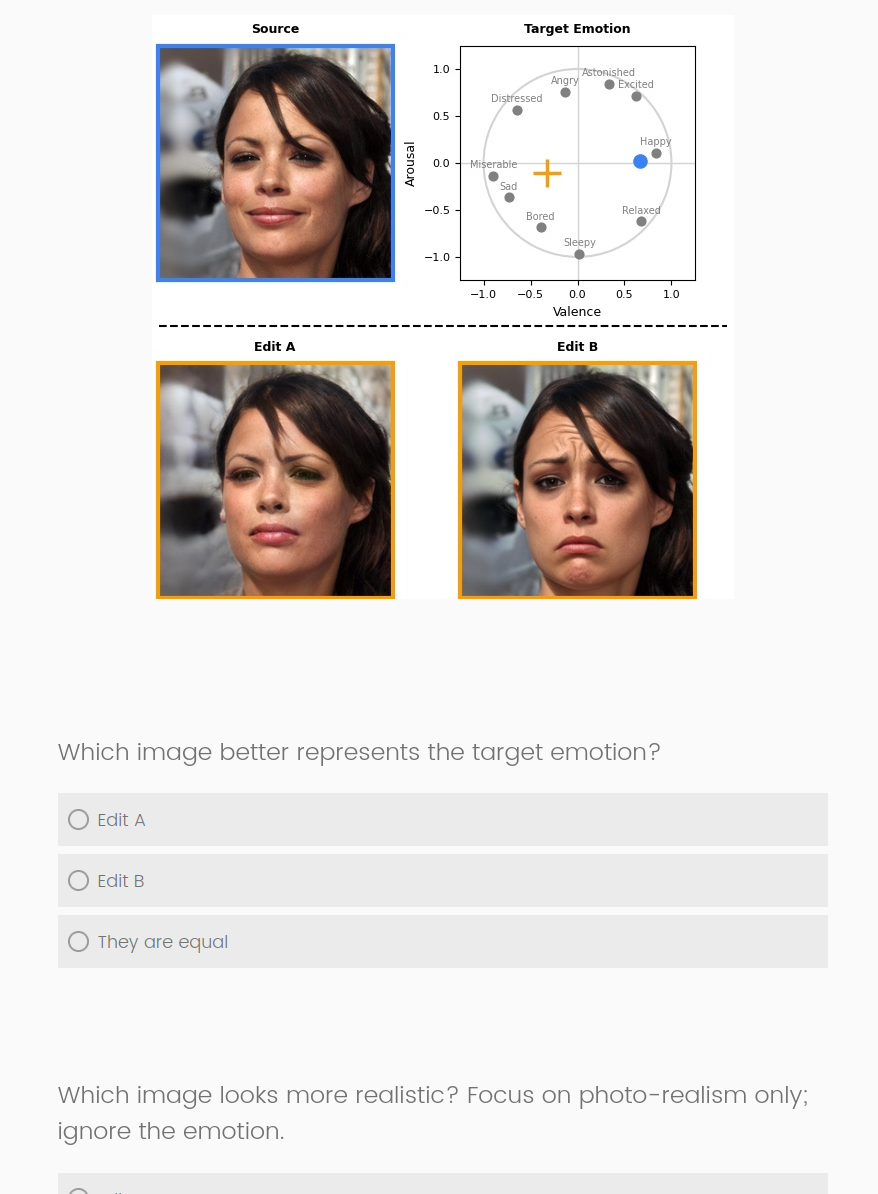}
    \caption{Screenshot of a question from our a user study.}
    \label{fig:user_study_screenshots-b}
  \end{subfigure}
  
  \caption{User study screenshots.}
  \label{fig:user_study_screenshots}
\end{figure}

\begin{figure}[h!]
  \centering
  \begin{subfigure}{0.485\linewidth}
  \centering
  \includegraphics[height=6.0cm]{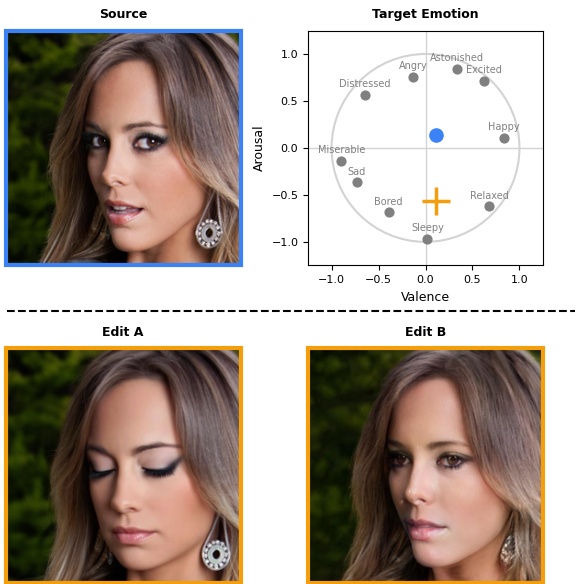}
    \caption{Example 1}
    \label{fig:user_study_example-a}
  \end{subfigure} \hfill
  \begin{subfigure}{0.485\linewidth}
  \centering
  \includegraphics[height=6.0cm]{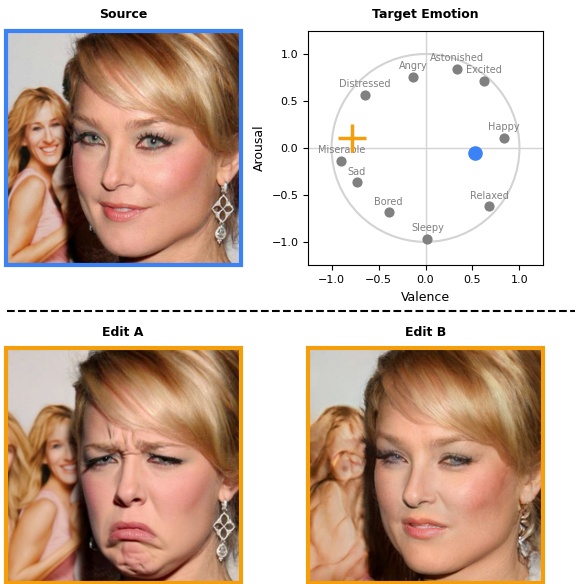}
    \caption{Example 2}
    \label{fig:user_study_example-b}
  \end{subfigure} 
   \\
  \begin{subfigure}{0.485\linewidth}
  \centering
  \includegraphics[height=6.0cm]{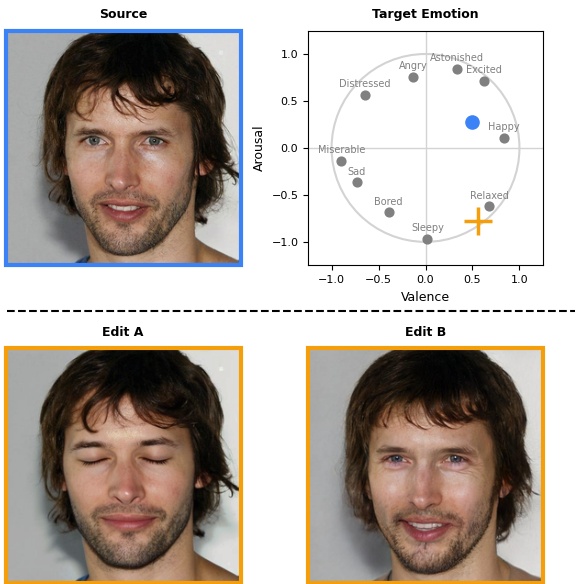}
    \caption{Example 3}
    \label{fig:user_study_example-c}
  \end{subfigure} \hfill
  \begin{subfigure}{0.485\linewidth}
  \centering
  \includegraphics[height=6.0cm]{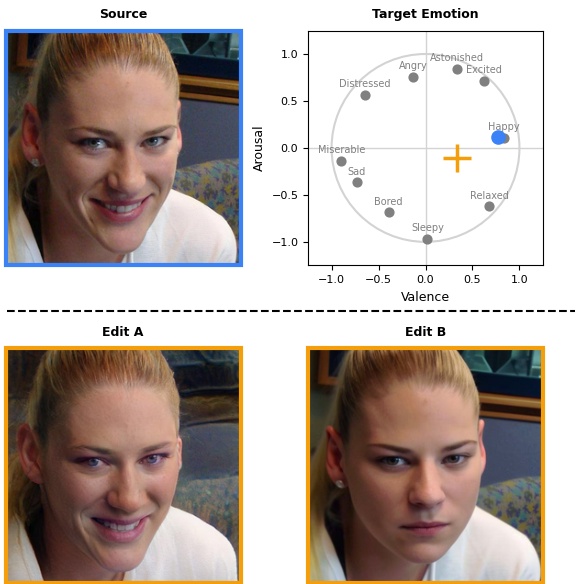}
    \caption{Example 4}
    \label{fig:user_study_example-d}
  \end{subfigure}
  \caption{Examples of image comparisons from our user study. Our edits are Edit A, Edit A, Edit A, and Edit B in (a), (b), (c), and (d), respectively.}
  \label{fig:user_study_example}
\end{figure}

\begin{figure}[t]
  \centering
  \includegraphics[height=4.0cm]{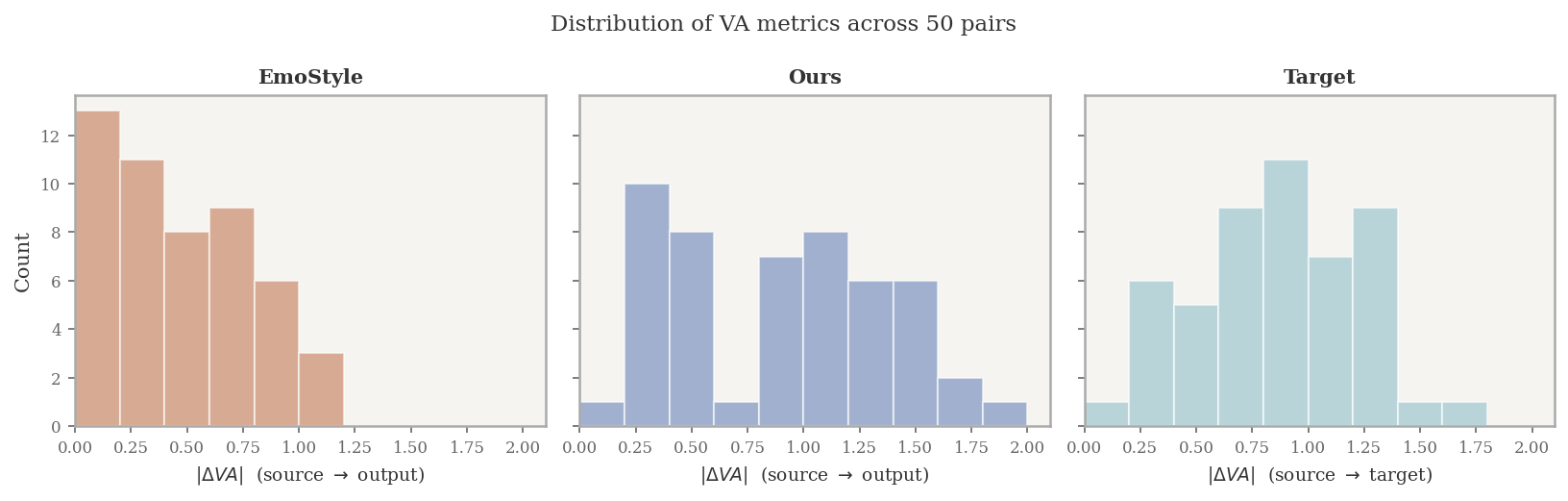}
  \caption{Magnitudes of the valence-arousal (VA) edits shown in our user study, as measured by EmotiEffLib, for EmoStyle (left) and ours (middle). The plot of the right shows the magnitude of the VA edit, as measured by the distance between the source images VA values (measured by EmotiEffLib) and the target VA value that both editing models are aiming at.}
  \label{fig:user_study_VA_changes}
\end{figure}

Following the tutorial we then show user 20 sets of images, randomly selected from a total of 50 sets, and shown in a random order. From this we ask four questions:
\begin{enumerate}
    \item ``Which image better represents the target emotion?''
    \item ``Which image looks more realistic? Focus on photo-realism only; ignore the emotion.''
    \item ``Which image better preserves the identity of the original image?''
    \item ``Which image better preserves other visual details (excluding identity and emotion), such as background, pose, clothing/accessories, and colour/style?''
\end{enumerate}
For all questions, users must select one of the following options: ``Edit A'', ``Edit B'', or ``They Are Equal''. Screenshots from the user study are shown in Figure \ref{fig:user_study_screenshots}. We also provide additional images from four more questions in Figure \ref{fig:user_study_example}. Finally, we analyse the magnitude of VA changes between the source and edited images used in the survey in Figure \ref{fig:user_study_VA_changes}. VA is predicted using EmotiEffLib \citep{savchenko2023facial}. From this we can see that EmoStyle is much more limited in the strength of VA edits it can make, and that the size of VA edits made by our model are more closely aligned with the size we are aiming for with our target VA values.

We obtain ethics approval from our Institutional Review Board (IRB) to conduct this user study. The consent form contained this section regarding possible risks for the participants: ``No risks are envisaged for this study.  Some paired image may only contain subtle differences so please take care to avoid eye stain when considering the images.''.

\section{Broader Impact}
\label{sec:broader_impact}

This work introduces a training-free framework for continuous emotional image editing, enabling nuanced creative expression and psychological research across diverse visual domains. While these advancements improve affective controllability, they could be misused to generate deceptive content or deepfakes by manipulating the emotional states of subjects in images. Finally, because our method relies on the internal representations of large foundation models, users should be aware that the range and intensity of generated emotions are ultimately constrained by the patterns present in the models' original training data.

\newpage

\newpage

\newpage

\newpage

\newpage

\newpage


\clearpage

\end{document}